\documentclass{article}

 \usepackage[preprint]{neurips_2026}

\usepackage[utf8]{inputenc} 
\usepackage[T1]{fontenc}    
\usepackage{hyperref}       
\usepackage{url}            
\usepackage{booktabs}       
\usepackage{amsfonts}       
\usepackage{nicefrac}       
\usepackage{microtype}      
\usepackage{xcolor}         
\usepackage{comment}

\usepackage{hyperref}
\usepackage{multicol}

\usepackage{amsmath}
\usepackage{amssymb}
\usepackage{mathtools}
\usepackage{amsthm}
\usepackage[textsize=tiny]{todonotes}
\usepackage{enumitem}
\usepackage{algorithm}
\usepackage{algorithmic}
\usepackage[capitalize,noabbrev]{cleveref}
\usepackage{subcaption}
\usepackage{wrapfig}
\usepackage[rightcaption]{sidecap}

\newcommand{\namedcomment}[2]{%
  \textcolor{red}{[#1: #2]}%
}

\newcommand{\travis}[1]{\namedcomment{Travis}{#1}}

\title{Unsupervised Continual Learning with Growing Self-Organizing Maps and Synthetic Replay}

\author{
  Pujan Thapa \\
  Department of Computer Science \\
  Rochester Institute of Technology \\
  Rochester, NY, USA \\
  \texttt{pt6757@rit.edu}
  \And
  Alexander Ororbia \\
  Department of Computer Science \\
  Rochester Institute of Technology \\
  Rochester, NY, USA \\
  \texttt{ago@cs.rit.edu}
  \And
  Travis Desell \\
  Department of Computer Science \\
  Rochester Institute of Technology \\
  Rochester, NY, USA \\
  \texttt{tjdvse@rit.edu}
}

\begin{document}

\maketitle

\begin{abstract}
This work presents a generative continual learning framework based on growing self-organizing maps (GSOMs) that are augmented with learned distributional statistics as well as encoder–decoder models for class-incremental learning. The proposed approach enables exemplar-free replay using distributional statistical memory, which eliminates the need to store raw data. Each GSOM unit maintains its own mean, variance, and covariance estimates, which are subsequently used to generate synthetic samples for replay; in encoder–decoder configurations, these samples are then decoded back into the input space (via ancestral sampling) for subsequent training. Our method is fully unsupervised, as it does not rely on explicit task boundaries or class labels during training. Results across multiple benchmarks show that the proposed approach achieves performance competitive even with supervised state-of-the-art memory-based methods while consistently outperforming memory-free approaches. In several settings, our framework matches or exceeds existing baselines, particularly in challenging single-class incremental scenarios. We also provide baseline results for single-class incremental TinyImageNet and MiniImageNet, offering a useful reference for future work. 
This work highlights the effectiveness of an unsupervised, adaptive, topology-driven neural form of statistical replay as a scalable, flexible approach to continual learning.
\end{abstract}

\section{Introduction}
\label{sec:intro}

Computational intelligent systems operating in real-world environments are often exposed to continuous streams of data, where the underlying data distribution may change over time. In these settings, adaptive systems must extract and incorporate new information as it arrives while preserving previously acquired knowledge. This ability, commonly referred to as continual learning (CL) or lifelong learning~\citep{thrun1998lifelong}, is essential for building systems that remain functional and reliable under non-stationary data streams.  A key challenge in continual learning is catastrophic forgetting \cite{french1999catastrophic}, where artificial neural networks (ANNs) tend to overwrite previously-acquired knowledge when trained on new data. This problem is closely tied to the stability–plasticity dilemma~\cite{abraham}, which governs the trade-off between learning new information and preserving prior knowledge (encoded as internal representations). Among the categories of CL, class incremental learning (CIL) is one of the most challenging, as the model must continuously discriminate among all previously seen classes without access to task identification information~\citep{van2022three}.



In general, most CL benchmarks assume clearly defined task boundaries, where new tasks introduce several classes simultaneously, e.g., as in Split-MNIST or Split-CIFAR~\citep{survey1, survey2_ptm, survey3}. However, such assumptions rarely hold in real-world data streams, where task boundaries are often unknown or ill-defined. Most existing approaches address forgetting through one of three strategies:
(1) regularization,
(2) architectural expansion, or
(3) replay.
Regularization-based methods, such as elastic weight consolidation (EWC)~\citep{schwarz2018progresscompressscalable, cf_1}, constrain parameter updates to preserve important knowledge. Expansion-based approaches, including progressive neural networks~\citep{rusu2022progressiveneuralnetworks}, increase model capacity to accommodate new information. Replay-based methods, such as DER++~\citep{buzzega2020darkexperiencegeneralcontinual} and MEMO~\citep{zhou2023model603exemplarsmemoryefficient}, mitigate forgetting by revisiting past samples (or approximations of them). While effective, these approaches rely on assumptions that are difficult to satisfy in real-world settings; these include known task boundaries, access to labeled data, or the ability to store and manage growing memory buffers.

As a result, recent work has shifted to task-free CL (TFCL), where data is presented as a continuous stream without explicit task boundaries or identities ~\citep{Aljundi_2019_CVPR, gunasekara2023survey, ororbia2021continual}. Early work by Aljundi et al.\citep{Aljundi_2019_CVPR} proposed one of the first TFCL formulations, extending importance-weight regularization to online settings by detecting stable learning regimes from the loss surface and updating parameter importance only when the model reached plateau regions. Memory-based task-free methods, e.g., GMED \citep{jin2021gradientbasededitingmemoryexamples} and ODDL \citep{ye2022task}, demonstrated that replay can be strengthened not only by selecting past samples but also by actively ``editing'' stored examples to make them more useful for future replay. Other more recent memory-based schemes in this direction include those based on neural Dirichlet mixture processes \citep{lee2020neural} as well as those that continuously organize and maintain samples/prototypes per class~\citep{delange2021continualprototypeevolutionlearning}.
Although effective, memory-based methods underscore a broader reliance on stored samples and replay heuristics, raising concerns regarding memory growth, privacy, and long-term scalability.
More recently, memory-free task-free approaches have been explored that avoid explicit replay buffers and instead rely on implicit knowledge retention and adaptive optimization strategies \citep{michel2025offlineonlinememoryfreetaskfree}. Nevertheless, such methods often struggle to maintain stable representations over long data streams due to the absence of explicit mechanisms to preserve past information.

Most CL research has focused on deep ANNs~\cite{10599804}; however, self-organizing maps (SOMs)~\cite{kohonen}, which come from a broader class of competitive neural models \cite{ororbia2021continual}, offer a topology-preserving, unsupervised alternative with inherently local updates that can reduce interference between past and new knowledge. Prior SOM-based CL approaches, such as SOMLP~\cite{bashivan2019}, DendSOM~\cite{pinitas2021}, and c-SOM~\cite{hiteshcSOM}, demonstrate this potential yet are limited by fixed capacity or lack an adaptive, scalable replay mechanisms. To address these limitations for task-free settings, we propose a \emph{growing self-organizing map (GSOM) framework} for task-agnostic continual learning in the context of a class-sequential data stream. Unlike conventional approaches, GSOM provides an adaptive, topology-preserving structure that expands incrementally as new patterns arrive ~\citep{Alahakoon}, organizing its latent space into localized regions, which reduces interference (or neural cross-talk \cite{french1999catastrophic,ororbia2021continual}). Notably, our model ``grows'' new portions of its topology via a criterion we develop based on statistical ``surprisal'', which is used to determine if incoming samples are poorly explained (side-stepping any need for predefined error thresholds). The GSOM's combination of localized representation and probabilistic growth enables it to stably, continually adapt to data complexity without relying on task boundaries or explicitly-stored patterns. Furthermore, we integrate our GSOM with encoder-decoder structures (specifically convolutional VAEs~\citep{Kingma_2019} and pretrained visual encoders such as  CLIP~\citep{radford2021learningtransferablevisualmodels}) in order to handle complex data patterns, e.g., natural images.

In effect, the main contributions of this work are {\it i)} a GSOM-centric CL framework that combines generative modeling with topology-preserving clustering to enable interpretable, unsupervised learning from data streams;  {\it ii)} a surprisal-driven growth mechanism for the GSOM that dynamically expands the internal map when current representations become insufficient to explain incoming samples (meaning the topology expands as needed, without task boundary triggers); and, {\it iii)} a generative replay scheme that stores compact summary statistics (mean, variance, and covariance) per GSOM unit, enabling rehearsal without buffers or data storage. We evaluate three variants of our model framework on MNIST, CIFAR-10/100, TinyImageNet, and MiniImageNet, under task-boundary-free protocols, and show improved or competitive performance with prior CL methodology.

\section{Methodology}
\label{sec:methodology}

Self-organizing maps (SOMs), also known as Kohonen maps~\citep{kohonen}, are a form unsupervised learning that entails projecting high-dimensional data onto a  two-dimensional grid while preserving topological relationships. Each unit in the grid represents a prototype vector and, during training, input samples are mapped to their best-matching unit (BMU). The BMU, and its neighboring units, are then updated to move towards the input sample, with the magnitude of the update applied decreasing as a function of distance from the BMU. After training, each input can be associated with its BMU, which serves as a representative anchor within the SOM's learned topology. For labeled datasets, BMUs can also accumulate class distributions, supporting interpretability as well as weakly-supervised clustering.


\subsection{Growing Self-Organizing Maps}
\label{sec:growing_soms}

Conventional SOMs require their topological grid size to be pre-defined, however growing SOMs (GSOMs)~\citep{fritzke1994growing,Alahakoon} allow the map structure to expand when the existing representation becomes insufficient to capture local data variance (Figure ~\ref{fig:gsom_growth}, see Appendix). Growth is triggered when the accumulated error of a unit exceeds a predefined threshold derived from a spread factor parameter. When this occurs, new units are inserted at the map boundary so as to represent previously under-modeled regions of the data space. Through this mechanism, the GSOM allocates representational capacity where it is most needed. Dense, complex regions of the data distribution naturally receive more units, whereas simpler regions remain sparsely represented. Thus, GSOM learns a topology that more closely reflects the dataset's underlying structure without requiring a predetermined grid size.

In our approach, instead of relying solely on accumulated quantization error to trigger growth, we introduce a surprisal-driven growth mechanism that measures how poorly the current map representation ``explains'' an incoming sample. When surprisal of an observation exceeds a dynamically model-maintained threshold, our GSOM expands locally to accommodate the new information, allowing the topology to adapt continuously as new data arrives. Growth is triggered at a specific unit (referred to as the \textit{growth node}) when its accumulated error or surprisal exceeds a predefined threshold. Instead of expanding globally, the map grows locally by inserting new units in the immediate $4$-neighborhood of the growth node, i.e., at coordinates $(x \pm 1, y)$ and $(x, y \pm 1)$. If a neighboring position is not occupied, a new unit is created at that location. The weights of newly-added units are initialized using the growth-triggering input data and surrounding existing units. This localized expansion allows the GSOM to incrementally refine regions where the current representation is insufficient while still preserving well-structured areas, balancing plasticity and stability.


Building on the above mechanism, we develop a task-agnostic class-incremental CL framework in which GSOM neurons act as generative memory units storing per-unit distributional summaries (mean, variance, and covariance) of previously observed data. The full training procedure is formalised in Algorithm~\ref{alg:unified_algo} (see Appendix). At each batch $t$, the model processes $\mathcal{T}_t = \mathcal{B}_t \cup \mathcal{R}$, where $\mathcal{B}_t$ is the incoming batch and $\mathcal{R}$ contains replayed samples from the current GSOM units. If a global encoder is used, it is updated on $\mathcal{T}_t$ and the GSOM receives latent codes $\mathcal{Z}_t$; otherwise the GSOM operates directly on $\mathcal{T}_t$. After each batch, $K$ synthetic samples are drawn from each active BMU's bias-corrected statistics and decoded to form $\mathcal{R}$ for the next step. The GSOM is then trained with batch $t$ for $I_{total}$ iterations, with growth only being allowed for the first $I_{grow}$ iterations.

Our framework naturally extends to different data modalities; 
for high-dimensional visual datasets (e.g., CIFAR-10/100, TinyImageNet and MiniImageNet), our GSOM is applied over a learned latent space obtained from models such as variational autoencoders or pre-trained encoders. Importantly, the training process remains fully unsupervised: class labels are never used for representation learning or the GSOM updates and are only employed post-hoc for evaluation (via BMU-based labeling).

\subsection{Tracking Neural Unit Distributional Statistics}
\label{subsec:running_stats}

Each GSOM neural unit maintains running estimates of the distribution of samples that are mapped to it. These statistics serve two purposes in the proposed framework:
(i) enable generative replay, and
(ii) drive our probabilistic surprisal criterion used to trigger our GSOM's topology growth. Given a GSOM unit at position $(i,j)$ and a momentum factor $\alpha$, an input sample $x$ mapped to the best matching unit (BMU) updates the unit's core properties as follows:
\textit{(i)}, a running \textbf{mean vector} (or exponential moving average): $\mu_{ij} \leftarrow (1 - \alpha)\mu_{ij} + \alpha x$;
\textit{(ii)}, a running \textbf{variance vector} $\sigma^2_{ij}$, which captures per-dimension variability, calculated by: $\sigma^2_{ij} \leftarrow (1 - \alpha)\sigma^2_{ij} + \alpha (\mu_{ij} - x)^2$; and,
\textit{(iii)}, a running \textbf{covariance matrix} $\Sigma_{ij}$, which models inter-feature relationships, calculated via: $\Sigma_{ij} \leftarrow (1 - \alpha)\Sigma_{ij} + \alpha (x - \mu_{ij})(x - \mu_{ij})^\top$.
Here, $x$ denotes either a raw input pattern (when the GSOM operates directly on data) or a latent representation produced by an encoder model at step $t$. These statistics characterize the local distribution of pattern vectors assigned to each GSOM unit. For a given BMU, neighboring units are updated according to the standard GSOM neighborhood adaptation rule to preserve topological smoothness. However, distributional statistics (mean, variance, and covariance) are updated only for the BMU associated with the input sample to maintain a localized and statistically consistent estimate of the underlying data distribution. We also further apply bias-correction similar to the Adam optimizer~\citep{kingma2017adammethodstochasticoptimization} to these statistics to alleviate bias during initialization (fully described in Appendix~\ref{app:bias_correction}).

\subsection{Gaussian Surprisal for BMU-Based GSOM.}
\label{sec:gaussian_surprise}

Given our model's per-unit Gaussian statistics, each GSOM neuron $b$ at position $(i,j)$ in the map models the distribution of the patterns assigned to it as a multivariate Gaussian
$z \sim \mathcal{N}(\mu_b, \Sigma_b)$,
where $z \in \mathbb{R}^d$ is the data (or latent representation of it), $\mu_b \in \mathbb{R}^d$ is the running mean of unit $b$, and $\Sigma_b \in \mathbb{R}^{d\times d}$ is the corresponding covariance matrix. The resulting likelihood of observing $z$ under the BMU distribution is:
\begin{equation}
p(z \mid \mu_b,\Sigma_b)
=
\frac{1}{(2\pi)^{d/2}|\Sigma_b|^{1/2}}
\exp\!\left(
-\frac{1}{2}(z-\mu_b)^T\Sigma_b^{-1}(z-\mu_b)
\right).
\end{equation}
Consequently, we define the surprisal of sample $z$ under BMU $b$ as the negative log-likelihood $S_b(z) = -\log p(z \mid \mu_b,\Sigma_b)$. Substituting the Gaussian density, taking the negative logarithm, and applying standard log identities yields the following:
\begin{align}
S_b(z)
&=  \frac{1}{2}(z-\mu_b)^T \Sigma_b^{-1}(z-\mu_b)
+ \frac{1}{2}\log |\Sigma_b|
+ \frac{d}{2}\log(2\pi).
\end{align}
The full derivation is provided in the Appendix~\ref{app:nll_derivation}. For topology growth decisions, additive constants in the negative log-likelihood can be omitted because they do not affect inequality-based threshold comparisons. The term $\frac{d}{2}\log(2\pi)$ is constant, while the log-determinant term $\frac{1}{2}\log|\Sigma_b|$ reflects the global volume of the Gaussian region rather than the local deviation of $z$ from the BMU center. Since the goal of our GSOM's growth is to detect when a sample lies far from the existing representation of a unit, our criterion can be based solely on the quadratic term.
This leads to our GSOM's normalized Mahalanobis surprisal measure:
\begin{equation}
S_b(z)
=
\frac{1}{d}(z-\mu_b)^T\Sigma_b^{-1}(z-\mu_b),
\end{equation}
which provides a dimension-independent measure of statistical deviation. When the surprisal of an incoming sample exceeds a threshold (defined later), the GSOM expands locally around the BMU to accommodate the new region of the data distribution.

This formulation provides a statistically grounded alternative to the traditional quantization error used in GSOM growth mechanisms. Instead of measuring only the distance between a sample and the BMU prototype, the surprisal measure also considers the spread and shape of the data assigned to that unit. As a result, our model is better equipped to detect when an incoming sample does not fit an existing representation. When such cases occur, the GSOM expands locally to create new units, allowing its topology to adapt to unseen regions of the data distribution in a streaming fashion.


\noindent\textbf{Growth decision rule: }
Building on the surprisal measure $S_b(z)$ defined before, growth decisions are governed by both distributional as well as structural criteria.
Our model's surprisal threshold adapts online to the recent history
$\mathcal{H} = \{S(\cdot)\}_{t-1000:t}$:
\begin{equation}
  T_S =
  \begin{cases}
    \operatorname{Percentile}_{92}(\mathcal{H}) & \text{if } |\mathcal{H}| > 50, \\
    1.7 & \text{otherwise,}
  \end{cases}
  \label{eq:threshold}
\end{equation}
where the fallback value (the second case of the piece-wise function) provides a conservative cold-start threshold before sufficient history has been accumulated.
Three constraints stabilize our GSOM's expansion:
\textit{(i)} a minimum hit count $h_b \geq h_{\min}$ that ensures reliable per-unit statistics before growth is permitted,
\textit{(ii)} a cool-down $\Delta t_b > t_{\text{cool}}$ which prevents repeated growth from the same unit in consecutive iterations, and
\textit{(iii)} a linearly decaying growth probability  (given the $i$th iteration of the batch),
$p_{\text{grow}}(i) = \max\!\left(0.2,\; 1 - 0.5\,\frac{i}{I_{\text{grow}}}\right)$,
which reduces the GSOM's expansion pressure as training progresses through the growth phase of $I_{\text{grow}}$ batches.
The full growth condition at unit $b$ for sample $z$ at iteration $i$ can then be defined as:

\begin{equation}
\begin{aligned}
\mathcal{G}_b(z) = \mathbf{1}\big[ \;
& S_b(z) > T_S \land E_b > T_E \land h_b \geq h_{\min} \\
& \land\; \Delta t_b > t_{\text{cool}}
\land u < p_{\text{grow}}(i)
\land \exists (x',y') \in \mathcal{N}_4(b) :
(x',y') \notin \mathcal{A}
\;\big]
\end{aligned}
\label{eq:growth_cond}
\end{equation}
where $u \sim \mathcal{U}(0,1)$.
Here $h_b$ denotes the accumulated hit count of unit $b$, $\Delta t_b$ the number of iterations since $b$ last triggered growth, and $T_E$ is the quantization error threshold derived from the spread factor $T_E = -d\log(\text{SF})$~\citep{Alahakoon}. $E_b$ denotes the accumulated quantization error of unit $b$, updated whenever $b$ is selected as the BMU, and $\mathcal{N}_4(b)$ denotes the $4$-neighbourhood of unit $b$
(i.e., adjacent grid locations) while $\mathcal{A}$ is the set of currently active units. Values for $h_{\min}$, $t_{\text{cool}}$, and $G_{\max}$ are provided in Table~\ref{tab:unified_hyperparams} in the Appendix. 

When no free position exists in $\mathcal{N}_4(b)$, i.e., an interior unit demands growth, growth cannot be applied directly. In this case, the accumulated error is redistributed to neighbouring units following the standard GSOM update rule~\citep{Alahakoon}. This allows growth pressure to propagate toward nearby regions where expansion is feasible. A detailed formulation of the error $E_b$ update and propagation process is provided in Appendix~\ref{appendix:gsom_error}. 

\noindent\textbf{Growth expansion control: } To control model expansion, GSOM growth is restricted to the early stages of training each batch, where the model is trained for a total of $I$ iterations but growth is  only permitted during the first $I_{\text{grow}}$ iterations. During each growth batch, the number of newly added units is further limited to, at most, $G_{\text{max}}$ per iteration. After the growth phase, the GSOM structure is fixed and only weight updates are performed for remaining iterations. This strategy prevents uncontrolled expansion while allowing the model to adapt to the data distribution stably and gradually.

\subsection{Task-Free Continual Learning with GSOM and Synthetic Replay}

Unlike task-based CL, our model is never provided with task identitiers, class labels, or boundary signals. Samples arrive in a continuous stream (of mini-batches) and all components, including the GSOM and the integrated VAE or pretrained encoders, are updated solely based on the input stream and GSOM-replayed samples, all without distinguishing between real and synthetic data and without access to an explicit task boundary, sample classes, or transition signals. This places the framework squarely in a task-free unsupervised CL regime. Unlike classical buffer-based replay methods, which store explicit samples, our  approach relies solely on distributional summaries for replay. Notably, even without explicit data storage, our model is still capable of generating meaningful and diverse samples that reflect the learned data distribution.

As described in Section~\ref{sec:growing_soms}, when a new mini-batch of samples arrives, synthetic samples are generated from the GSOM by sampling from the stored distributions of its neural units. These replayed samples are then combined with the mini-batch to mitigate forgetting. As each GSOM unit maintains distributional statistics over its assigned samples, synthetic samples are generated as:
\setlength{\abovedisplayskip}{0pt}
\setlength{\belowdisplayskip}{0pt}
\begin{align}
\tilde{x} &\sim \mathcal{N}(\hat{\mu}_{ij}, \hat{\sigma}^2_{ij})
\quad \text{(MNIST)} \nonumber \\
\tilde{z} &\sim \mathcal{N}(\hat{\mu}_{ij}, \hat{\Sigma}_{ij})
\quad \text{(CIFAR-10 / CIFAR-100 / TinyImageNet/ MiniImageNet)} \nonumber
\end{align}
For low-dimensional datasets, e.g., MNIST, synthetic sample generation using SOM-based statistics is generally effective, since the feature space ($28 \times 28$ grayscale pixels) is sufficiently less for independent mean–variance modeling. In this setting, per-dimension variance captures the data distribution reasonably well, enabling stable sampling and efficient replay.

However, this approach does not scale to higher-dimensional natural image datasets. In pixel space, mean–variance sampling produces low-quality samples and degrades downstream performance. Moreover, modeling full covariance in the original space is computationally prohibitive; for instance, a single BMU for a CIFAR10 image would require storing a $3072 \times 3072$ covariance matrix for RGB images, which is infeasible in practice. To address these limitations, we perform replay in a lower-dimensional latent space. High-dimensional inputs are first mapped to lower-dimensional embeddings via an encoder, which allows the GSOM to operate on compact representations. This significantly reduces both storage and computational requirements while enabling full covariance modeling. For this covariance-based sampling, eigenvalue regularization is applied to ensure numerical stability and to maintain a valid positive semi-definite covariance matrix (see Algorithm \ref{app:covar-synth-alg} in Appendix~\ref{app:experimental_settings}).


We study four configurations of our framework:
\textit{(i)} a GSOM operating directly on input space,
\textit{(ii)} a GSOM combined with a VAE where replay occurs in latent space and samples are decoded (i.e., ancestral sampled) into input space,
\textit{(iii)} a GSOM operating on pretrained embedding spaces such as those obtained from ResNet-18 and CLIP, where a trained decoder reconstructs images from our sampled latents (see Appendix~\ref{app:encdec_arch}), and \textit{(iv)} a modular form of replay where each GSOM unit is paired with its own local encoder-decoder (see Appendix ~\ref{app:per_bmu_enc_dec}). Our methodology is usefully agnostic to the choice of encoder-decoder architecture and for the later configurations, the encoder produces latent codes (of dimensionality $d=128$ or $d=512$) making full covariance estimation for each BMU tractiable.

\begin{wrapfigure}{r}{0.35\textwidth}
    \centering
    \vspace{-1.8cm}
    \includegraphics[width=\linewidth]{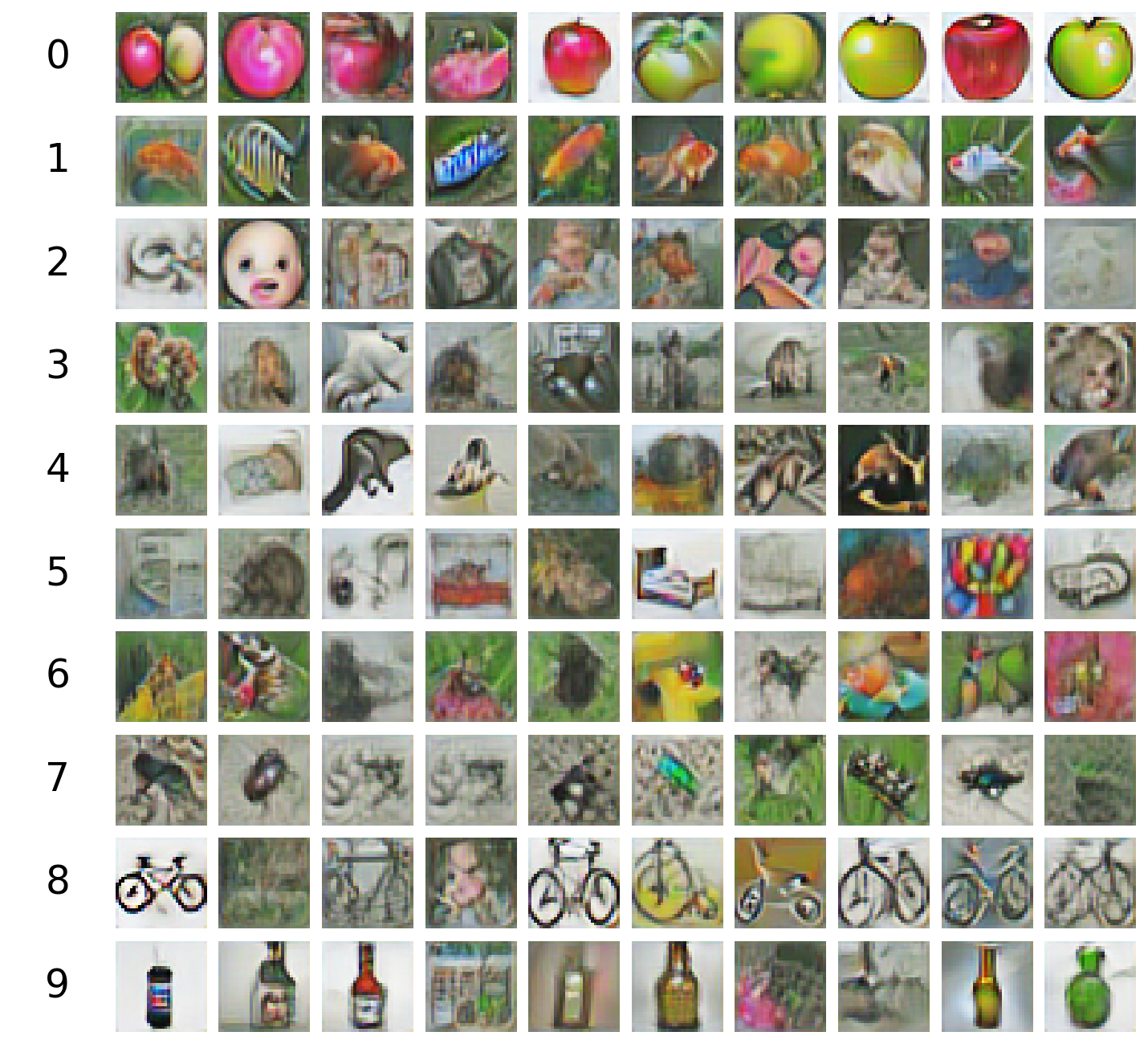}
    \vspace{-0.3cm}
    \caption{
    CIFAR-100 samples for the first $10$ classes generated by sampling from a GSOM integrated with ResNet-18 embeddings. Each row corresponds to one class ($0$–$9$).
    }
    \label{fig:som_synthetic_cifar100}
    \vspace{-0.4cm}
\end{wrapfigure}

In all cases, the GSOM serves as a structured memory that stores only statistical summaries (mean, variance, and covariance) for each neural unit. This enables exemplar-free replay and the storage scales with the number of GSOM units as opposed to the size of the dataset. The GSOM's topology-preserving structure further supports interpretability; Figure~\ref{fig:som_synthetic_cifar100} shows synthetic CIFAR-100 samples decoded from BMU statistics and Appendix~\ref{app:generative_model} provides further visualisations on other datasets.
By decoding unit representations or sampled latent vectors, it is also possible to inspect the evolution of learned features over time (see Appendix~\ref{app:som_representations}).
Well-separated and coherent clusters on the GSOM grid indicate stable learning, whereas fragmentation may reflect forgetting or underfitting. This makes the framework not only generative but also inherently interpretable.

\noindent

\section{Results}
\label{sec:results}

Our approach is evaluated on standard CL benchmarks, including MNIST, CIFAR-10, CIFAR-100, TinyImageNet, and MiniImageNet. Experiments were conducted under a streaming setting where data from different classes was presented sequentially as well as under a split protocol in which the dataset was divided into disjoint class groups. To account for variability due to data ordering, results are averaged over $20$ independent runs with different random class sequences. Performance is reported as the mean accuracy along with the standard deviation across runs. Hyperparameters were selected based on extensive ablation studies (see Appendix \ref{app:experimental_settings}). All experiments were conducted on a desktop workstation equipped with an Intel Core i7-14700F CPU, 32 GB RAM, and an NVIDIA GeForce RTX 4060 Ti GPU with 16GB VRAM, running Windows 11.


\paragraph{Label Usage and Evaluation Protocol.}
Note that all GSOM and encoder-decoder models were trained in an unsupervised manner without using class labels, task labels, or task identifiers -- data was organized into class-wise or split-wise sequences only for evaluation purposes against existing work. After training, each unit is assigned the majority class of its mapped training samples; test samples inherit their BMU's label. Therefore, class-wise representations, task-wise accuracy matrices, forgetting metrics, and confusion matrix analyses (see Appendices~\ref{app:c_matrix}, ~\ref{app:cl_metrices} and~\ref{app:som_representations}) are post-hoc evaluation tools and were not used to guide model training or replay.

\subsection{Standard Split Multi Class Incremental Learning}

\textbf{Split Benchmarks without pretraining:} Table~\ref{tab:split_non_ptm} presents results on Split-MNIST, Split CIFAR-10, and Split CIFAR-100 without the use of pretrained representations. We compare our VAE-integrated GSOM (GSOM+VAE) framework with standard CL baselines, including regularization methods (EWC~\citep{cf_1}, SI~\citep{zenke2017continuallearningsynapticintelligence}, LwF~\citep{8107520}), replay-based methods (GEM~\citep{lopezpaz2022gradientepisodicmemorycontinual}, iCaRL~\citep{rebuffi2017icarlincrementalclassifierrepresentation}, GSS~\citep{aljundi2019gradientbasedsampleselection}), and probabilistic approaches, e.g., CN-DPM~\citep{lee2020neural}. As expected, regularization-based methods perform poorly across all datasets, with accuracies close to fine-tuning, indicating severe catastrophic forgetting. Replay-based methods -- GEM, iCaRL, and GSS -- provide improved retention, with iCaRL achieving $72.55$\% on Split-MNIST and GSS reaching $49.22$\% on Split CIFAR-10. CN-DPM achieves strong performance on Split-MNIST ($93.81$\%) but does not generalize consistently across more complex datasets. In comparison, our GSOM-based approach achieves competitive performance without relying on explicit task identity or exemplar storage. In particular, the global VAE-GSOM model attains $50.11$\% on Split CIFAR-10 and $13.81$\% on Split CIFAR-100, outperforming most baselines in these settings. On Split-MNIST, the global VAE-GSOM model achieves over $92.11$\% accuracy, approaching the best-performing methods.

\begin{table*}[t]
\centering
\scriptsize
\caption{Classification accuracy for the class-incremental learning experiments on Split-MNIST, Split CIFAR-10, and Split CIFAR-100 without any pre-trained models. Our results are reported as mean $\pm$ standard deviation over $20$ runs, whereas the compared results are over $5$ runs. $^\dagger$ denotes task-free setup and $^\ddagger$ denotes unsupervised training.
Methods without symbols are task-aware and supervised.}
\label{tab:split_non_ptm}
\begin{tabular}{lccc}
\toprule
\textbf{Method} & \textbf{Split-MNIST} & \textbf{Split-CIFAR-10} & \textbf{Split-CIFAR-100} \\
\hline

iid-offline & $95.82 \pm 0.33$ & $80.54 \pm 0.63$ & $48.09 \pm 0.90$ \\
Fine-Tune   & $19.68 \pm 0.02$ & $19.19 \pm 0.06$ & $8.32 \pm 0.23$ \\

EWC & $19.92 \pm 0.35$ & $16.18 \pm 1.37$ & $4.41 \pm 0.37$ \\
SI  & $19.76 \pm 0.01$ & $17.27 \pm 0.87$ & $5.87 \pm 0.21$ \\
LwF & $20.54 \pm 0.64$ & $18.53 \pm 0.12$ & $6.93 \pm 0.32$ \\

GEM    & $48.57 \pm 5.26$ & $25.54 \pm 0.19$ & $6.18 \pm 0.20$ \\
iCaRL  & $72.55 \pm 0.45$ & $35.88 \pm 1.43$ & $\mathbf{15.76 \pm 0.15}$ \\
GSS    & $54.14 \pm 4.68$ & $49.22 \pm 1.71$ & $11.33 \pm 0.40$ \\
CN-DPM$^\dagger$ & $\mathbf{93.81 \pm 0.07}$ & $46.98 \pm 0.62$ & -- \\
\hline
\multicolumn{4}{l}{\textbf{Ours (GSOM + VAE)}} \\
\hline
GSOM only$^{\dagger\ddagger}$      & $91.22 \pm 1.23$ & -- & -- \\
Global VAE-GSOM$^{\dagger\ddagger}$  & $92.11 \pm 0.98$ & $\mathbf{50.11 \pm 0.89}$ & $13.81 \pm 1.05$ \\
VAE-per-BMU-GSOM$^{\dagger\ddagger}$ & $91.01 \pm 0.12$ & $47.12 \pm 1.23$ & $12.49 \pm 2.01$ \\
\hline
\end{tabular}
\vspace{-0.5cm}
\end{table*}


Without pretrained representations, both SOM-only and VAE-based variants struggle to scale to high-dimensional data such as TinyImageNet and MiniImageNet (see Table~\ref{tab:tiny_mini_imagenet}). In particular, VAE-based GSOM models achieve only around $6\%$ accuracy on TinyImageNet and approximately $4\%$ on MiniImageNet, consistent with their CIFAR-100 performance. These results highlight the difficulty of learning meaningful generative representations from limited per-class data when training encoder--decoder models from scratch, motivating the use of stronger pretrained feature representations.

\textbf{Split Benchmarks with pretrained models:} Table~\ref{tab:split_ptm} reports results on Split-MNIST, Split CIFAR-10, Split CIFAR-100, and Split MiniImageNet under a task-free evaluation setting. We compare against a broad range of baselines, including naive fine-tuning; replay-based methods such as GEM ~\citep{lopezpaz2022gradientepisodicmemorycontinual}, iCARL~\citep{rebuffi2017icarlincrementalclassifierrepresentation}, reservoir sampling, MIR, and GSS~\citep{aljundi2019gradientbasedsampleselection}; and expansion-based, probabilistic approaches such as CN-DPM~\citep{lee2020neural}, CURL~\citep{curl}, CoPE, CoPE-CE ~\citep{delange2021continualprototypeevolutionlearning}, and Dynamic OCM~\citep{dynaOCM}. We also include memory-based variants such as A-GEM~\citep{agem}, ER, ER with data augmentation (ER\_a) ~\citep{er}, and ER with gradient-based memory editing (GMED)~\citep{jin2021gradientbasededitingmemoryexamples}, along with the iid-online upper bound. All methods use a common ResNet-18 backbone; ensuring that performance differences reflect the learning strategy rather than the underlying architecture. As expected, fine-tuning performs poorly due to catastrophic forgetting. Replay-based methods provide moderate improvements but remain limited on more complex datasets, even with more advanced sampling or memory strategies. Expansion-based and prototype-based approaches (CN-DPM, CoPE, and Dynamic OCM) achieve stronger performance, but still show limitations in high-dimensional settings.

\begin{table}[!t]
\centering
\scriptsize
\caption{Performance comparison on Split-MNIST, Split-CIFAR10, and Split-CIFAR100.
All methods use a pretrained ResNet-18 encoder and are evaluated over $20$ independent runs.
We additionally evaluate our method using a pretrained CLIP encoder, also over $20$ independent runs. Results marked with * and $\dagger$ are quoted from \citep{delange2021continualprototypeevolutionlearning}
and \citep{jin2021gradientbasededitingmemoryexamples}, respectively. $^\S$ denotes unsupervised training; methods without $^\S$ are supervised. Global denotes a frozen shared encoder, Global FT denotes a globally fine-tuned shared encoder, and BMU Specific denotes BMU-wise encoder adaptation.
}
\label{tab:split_ptm}

\begin{tabular}{lcccc}
\toprule
\textbf{Method} & \textbf{Split-MNIST} & \textbf{Split-CIFAR10} & \textbf{Split-CIFAR100}  &\textbf{Split MiniImagenet}\\
\hline
iid-online  & 85.99 $\pm$ 0.03 & 62.23 $\pm$ 1.5 & 18.13 $\pm$ 0.80 & 17.53 $\pm$ 1.6 \\
finetune*    & 19.75 $\pm$ 0.05 & 18.55 $\pm$ 0.34 & 3.53 $\pm$ 0.04 & 2.84 $\pm$ 0.4\\
GEM*         & 93.25 $\pm$ 0.36 & 24.13 $\pm$ 2.46 & 11.12 $\pm$ 2.48 & - \\
iCARL*       & 83.95 $\pm$ 0.21 & 37.32 $\pm$ 2.66 & 10.80 $\pm$ 0.37 & - \\
CURL* $^{\S}$        & 92.59 $\pm$ 0.66 & -- & -- & -- \\
CN-DPM      & 93.23 $\pm$ 0.09 & 45.21 $\pm$ 0.18 & 20.10 $\pm$ 0.12 &  27.07 $\pm$ 2.3\\
reservoir*   & 92.16 $\pm$ 0.75 & 42.48 $\pm$ 3.04 & 19.57 $\pm$ 1.79 & -- \\
MIR*         & 93.20 $\pm$ 0.36 & 42.80 $\pm$ 2.22 & 20.00 $\pm$ 0.57 & 25.21 $\pm$ 2.2 \\
GSS*         & 92.47 $\pm$ 0.92 & 38.45 $\pm$ 1.41 & 13.10 $\pm$ 0.94 & -- \\
CoPE*  & 93.94 $\pm$ 0.20 & 48.92 $\pm$ 1.32 & 21.62 $\pm$ 0.69 & -- \\
CoPE-CE*     & 91.77 $\pm$ 0.87 & 39.73 $\pm$ 2.26 & 18.33 $\pm$ 1.52 & -- \\

AGEM$^\dagger$               & 29.02 $\pm$ 5.3 & 18.49 $\pm$ 0.6 & 2.40 $\pm$ 0.2 & 2.09 $\pm$ 0.3 \\
ER + GMED $^\dagger$         & 82.67 $\pm$ 1.9 & 34.84 $\pm$ 2.2 & 27.27 $\pm$ 1.8 & 27.27 $\pm$ 1.8 \\
ER\_a  $^\dagger$   & 80.14 $\pm$ 3.2 & 46.29 $\pm$ 2.7 & 30.77 $\pm$ 2.2 & \\
Dynamic OCM & 95.67 $\pm$ 0.22 & 51.27 $\pm$ 1.47 & 29.87 $\pm$ 0.69 & 28.03 $\pm$ 2.1 \\
\hline
\multicolumn{5}{l}{\textbf{Ours (GSOM + ResNet-18)}} \\
\hline

Global$^{\S}$ & 96.12 $\pm$ 0.3 & 58.00 $\pm$ 1.31 & \textbf{35.67 $\pm$ 1.96} & \textbf{32.67 $\pm$ 1.23} \\
Global FT$^{\S}$ & \textbf{97.21 $\pm$ 0.45} & \textbf{58.1 $\pm$ 1.42} & 35.33 $\pm$ 1.28 & 33.60 $\pm$ 1.45 \\
BMU Specific$^{\S}$ & 94.34 $\pm$ 1.1 & 57.11 $\pm$ 1.56 & 34.89 $\pm$ 1.72 & 32.5 $\pm$ 1.06 \\
\hline
\multicolumn{5}{l}{\textbf{Ours (GSOM + CLIP)}} \\
\hline

Global$^{\S}$ & 91.11 $\pm$ 1.42 & 75.22 $\pm$ 1.18 & 62.31 $\pm$ 1.67 & 52.78 $\pm$ 1.23 \\
Global FT$^{\S}$ & 92.23 $\pm$ 1.36 & 78.84 $\pm$ 1.52 & 65.01 $\pm$ 1.41 & 52.33 $\pm$ 1.45 \\
BMU Specific$^{\S}$ & 90.01 $\pm$ 1.58 & 81.92 $\pm$ 1.27 & 64.22 $\pm$ 1.49 & 51.56 $\pm$ 1.72 \\
\hline
\end{tabular}
\vspace{-0.5cm}
\end{table}

In comparison, our GSOM-based methods consistently achieve superior or competitive performance across all benchmarks. A  significant aspect of our results is the learning
setting in which they are achieved.
All compared baselines -- except CURL -- are trained with full class supervision, receiving ground-truth labels at every update step throughout the data stream, whereas our method is unsupervised.
GSOM + ResNet-18 (Global FT) achieves the best results on Split-MNIST ($97.21$\%) and Split CIFAR-10 ($58.1$\%), while also outperforming all baselines on Split CIFAR-100 ($35.67$\%) and Split MiniImageNet ($33.60$\%). Notably, the gains are more pronounced on higher-complexity datasets such as CIFAR-100 and MiniImageNet, where catastrophic forgetting is more severe. Compared to strong baselines such as Dynamic OCM, CoPE, and ER-based methods, GSOM demonstrates improved stability, suggesting that its topology-preserving structure and distribution-driven replay better capture evolving data distributions.  When combined with CLIP embeddings, GSOM further improves performance, achieving up to $81.92$\% on split CIFAR-10 and $65.01$\% on split CIFAR-100. The improved performance with CLIP highlights the importance of high-quality pretrained embeddings, which provide better class separability and more stable representations under non-i.i.d. streams. This, in turn, enables GSOM to construct more reliable local distributions and to generate more effective replay samples. Consistent improvements in both ResNet-18 and CLIP settings indicate that GSOM benefits from stronger feature spaces while maintaining robustness through its distribution-driven task-free learning mechanism.

Table~\ref{tab:tiny_mini_imagenet} reports our GSOM performance on the Split-TinyImageNet benchmark using pretrained representations. With a ResNet-18 backbone, GSOM achieves stable performance in the range of $32$–$33$\%, indicating its ability to maintain consistent learning across multiple incremental tasks. When combined with CLIP embeddings, performance improves significantly, reaching up to $60.89$\%. This substantial gain highlights the impact of stronger pretrained feature representations on large-scale continual learning tasks. In particular, CLIP provides more discriminative and stable embeddings, which help to improve class separability and enable GSOM to construct more reliable local distributions for replay.
Overall, our empirical results demonstrate that although GSOM performs competitively with standard pretrained backbones, its effectiveness scales notably with higher-quality representations, reinforcing the importance of feature space quality in CL.

\subsection{Single Class Incremental Learning}
\label{subsec:scil}

 Table~\ref{tab:cil_pretrained_and_single} compares our GSOM-based methodology with several well-established CL approaches under the single-class incremental (SCI) setting, where each class is introduced sequentially ($10$, $100$, and $200$ tasks for MNIST, CIFAR-10, CIFAR-100, and TinyImageNet, respectively). While these baselines were originally designed for task-based class-incremental learning, we adapt them to the SCI protocol by treating each class arrival as an incremental update step. Importantly, these methods still rely on boundary-driven updates and are therefore not inherently task-free. In contrast, our GSOM-based approach operates in a fully task-free manner, where both representation learning and replay are driven continuously by the evolving data distribution without requiring explicit task or class boundaries. This distinction is especially important in the SCI setting, where the absence of task boundaries leads to more severe distribution shifts and increased forgetting.

\begin{table}[!t]
\centering
\scriptsize
\setlength{\tabcolsep}{3pt}   
\renewcommand{\arraystretch}{0.9}
\caption{Final classification accuracy for single class incremental learning. All baseline methods in this table are task-aware and supervised; they rely on explicit class arrival signals to trigger incremental updates, as opposed to the proposed GSOM framework which is task-free and unsupervised.}
\label{tab:cil_pretrained_and_single}
\begin{tabular}{lp{1.4cm}p{1.4cm}p{1.57cm}p{1.4cm}p{1.57cm}}
\toprule
\textbf{Method}
  & \multicolumn{3}{c}{\textbf{w/o pretraining}}
  & \multicolumn{2}{c}{\textbf{w/ pretraining}} \\
\cmidrule(lr){2-4} \cmidrule(lr){5-6}
  & \textbf{MNIST} & \textbf{CIFAR10} & \textbf{CIFAR100}
  & \textbf{CIFAR10} & \textbf{CIFAR100} \\
\hline
EWC       & 9.91  & 10.01 & 1.03  & 10.21 & 2.93 \\
LwF       & 19.96 & 10.05 & 2.13  & 19.39 & 6.25 \\
IMM       & 29.16 & 10.25 & 1.21  & 51.22 & 12.58 \\
PGMA      & 71.36 & 20.08 & 1.86  & 56.22 & 12.37 \\
RPSNet    & 40.29 & 16.31 & 1.96  & 55.54 & 4.13 \\
OWM       & 94.46 & 19.63 & 3.67  & \textbf{83.03} & 63.26 \\
DisCOIL   & \textbf{96.69} & 44.54 & --    & --    & -- \\
\midrule
\multicolumn{6}{l}{\textbf{Ours (GSOM-based)}} \\
\hline
GSOM only & 92.12 $\pm$ 0.84 & -- & -- & -- & --  \\
VAE (FT global) & 91.11 $\pm$ 1.26 & \textbf{51.12 $\pm$ 0.57} & \textbf{11.22 $\pm$ 1.43} & -- & -- \\
VAE (FT BMU specific) & 90.55 $\pm$ 0.91 & 47.72 $\pm$ 1.68 & 11.91 $\pm$ 0.36 & -- & -- \\
ResNet-18 (global) & -- & -- & -- & 58.21 $\pm$ 1.12 & 36.11 $\pm$ 0.49  \\
ResNet-18 (FT global) & -- & -- & -- & 59.01 $\pm$ 1.75 & 35.91 $\pm$ 0.28  \\
ResNet-18 (FT BMU specific) & -- & -- & -- & 58.65 $\pm$ 0.63 & 35.93 $\pm$ 1.34 \\
CLIP (global) & -- & -- & -- & 76.11 $\pm$ 0.72 & \textbf{63.87 $\pm$ 1.19} \\
CLIP (FT global) & -- & -- & -- & 79.21 $\pm$ 1.51 & 63.83 $\pm$ 0.41 \\
CLIP (BMU specific) & -- & -- & -- & 79.03 $\pm$ 0.95 & 63.12 $\pm$ 1.07 \\
\bottomrule
\end{tabular}
\vspace{-0.5cm}
\end{table}

\begin{table}[!t]
\centering
\scriptsize
\caption{GSOM-based method performance on Split-TinyImageNet and single-class incremental learning (SCIL) TinyImageNet and MiniImageNet.}
\label{tab:tiny_mini_imagenet}
\begin{tabular}{lccc}
\toprule
\textbf{Method} & \textbf{Split-TinyImageNet}  & \textbf{TinyImageNet (SCIL)} & \textbf{MiniImageNet (SCIL)}\\
\hline
VAE (Global FT) & 6.22 $\pm$ 0.67 & 6.12 $\pm$ 0.58 &  6.28 $\pm$ 1.12\\
VAE (BMU Specific) & 6.33 $\pm$ 0.88 & 6.01 $\pm$ 0.73 & 5.94 $\pm$ 1.36\\
GSOM + ResNet-18 (global) & 32.12 $\pm$ 0.78 & 30.67 $\pm$ 1.23 & 24.21 $\pm$ 0.92\\
GSOM + ResNet-18 (global FT) & 32.78 $\pm$ 1.56 & 31.23 $\pm$ 0.15 & 25.04 $\pm$ 0.71\\
GSOM + ResNet-18 (BMU Specific) & 32.59 $\pm$ 1.89 & 31.61 $\pm$ 1.51 & 24.63 $\pm$ 0.84\\
GSOM + CLIP (global) & 57.98 $\pm$ 1.11 & 40.45 $\pm$ 1.02 & 33.18 $\pm$ 0.58\\
GSOM + CLIP (global FT) & 58.25 $\pm$ 1.02 & \textbf{42.61 $\pm$ 0.91} & 34.02 $\pm$ 1.47\\
GSOM + CLIP (BMU Specific) & \textbf{60.89} $\pm$ 0.72 & 40.91 $\pm$ 1.14 & \textbf{34.57 $\pm$ 1.52} \\

\bottomrule
\end{tabular}
\vspace{-0.5cm}
\end{table}

As shown in Table~\ref{tab:cil_pretrained_and_single}, traditional regularization-based methods (EWC ~\citep{cf_1} and LwF ~\citep{8107520}) exhibit significant performance degradation across datasets, particularly on CIFAR-10 and CIFAR-100, indicating their inability to retain knowledge under extreme class-sequential streams. More advanced methods, e.g.,  PGMA~\citep{pgma}, RPSNet~\citep{rpsnet}, and OWM ~\citep{owm}, achieve moderate improvements but still struggle in high-dimensional settings. In contrast, our GSOM-based methods consistently achieve strong performance, particularly when combined with generative replay. The VAE-based GSOM variants outperform most baselines in the non-pretrained setting, achieving $51.12$\% on CIFAR-10 and $11.22$\% on CIFAR-100, demonstrating the effectiveness of distribution-driven replay without relying on stored exemplars or task identifiers. When combined with pretrained representations, the gains are further amplified. Using CLIP embeddings, GSOM achieves up to $79.21$\% on CIFAR-10 and $63.87$\% on CIFAR-100, significantly outperforming most baseline approaches and the best performing rehearsal-based method, i.e., DisCOIL ~\citep{discoil}. These results highlight that structured, topology-preserving replay combined with strong feature representations enables robust CL even under highly non-i.i.d. single-class streams.


\textbf{Benchmark: Single Class Incremental Learning (SCIL) TinyImageNet ($200$ sequential classes) and MiniImageNet ($100$ sequential classes):} To our knowledge, prior work does not report results under this protocol, as most studies adopt multi-class task splits. SCIL presents a significantly more challenging setting due to the absence of intra-task diversity, as the sequential batch of data consists of only a single class at a time. As shown in Table~\ref{tab:tiny_mini_imagenet}, while VAE-based variants trained from scratch perform poorly ($6$\%), pretrained features substantially improve results, with GSOM + CLIP achieving $42.61$\% on TinyImageNet and $34.57$\% on MiniImageNet. These performance improvements are achieved without explicit task boundaries, suggesting that GSOM is better suited for realistic task-free continual learning scenarios when such information is unavailable.

\section{Conclusion and Future Work}
\label{sec:conclusion}

In this work, we developed a GSOM-based framework for task-free continual learning (CL) that combines surprisal-driven structural growth with generative replay. The GSOM model dynamically adapts its topology to evolving data distributions while its units maintain distributional statistics used to generate synthetic samples, eliminating the need to store raw data. By integrating encoder–decoder architectures, 
our framework scales from low- to high-dimensional settings. Experimental results on MNIST, CIFAR-10, CIFAR-100, TinyImageNet, and MiniImageNet demonstrate that the proposed approach mitigates catastrophic forgetting and remains competitive with existing CL schemes. 

Despite the strengths shown by our proposed method, several limitations remain. The per-BMU encoder–decoder variant, while conceptually preferable for localized specialization, suffers from limited data availability per unit, leading to unstable training and reduced generative quality compared to global models. 
Additionally, our reliance on Gaussian-based distributional assumptions may not fully capture complex, multimodal data distributions, which can affect the fidelity of synthetic replay. In some cases, overlapping distributions across units may also reduce discriminative performance.

Future work will focus on improving both scalability and representation quality. Parallelizing GSOM updates is a natural direction for improving computational efficiency, particularly in large-scale settings. We will explore parameter-efficient per-BMU fine-tuning, where most encoder–decoder parameters are shared and only lightweight components (e.g., final layers or adapter modules) are BMU-specific. Extending the framework to a purely online learning regime, where each sample is processed in a single pass, is another key step toward real-world deployment. Additionally, more expressive generative models and non-Gaussian or mixture-based sampling strategies could improve our GSOM replay. In sum, this study demonstrates that combining topology-preserving structures with probabilistic growth and generative replay provides a promising direction for scalable, task-agnostic continual learning.

\bibliography{references}
\bibliographystyle{unsrt}

\newpage


\appendix

\begin{center}
    {\bf \LARGE Appendix}
\end{center}

\begin{figure}[!th]
    \centering
    \includegraphics[width=0.6\linewidth]{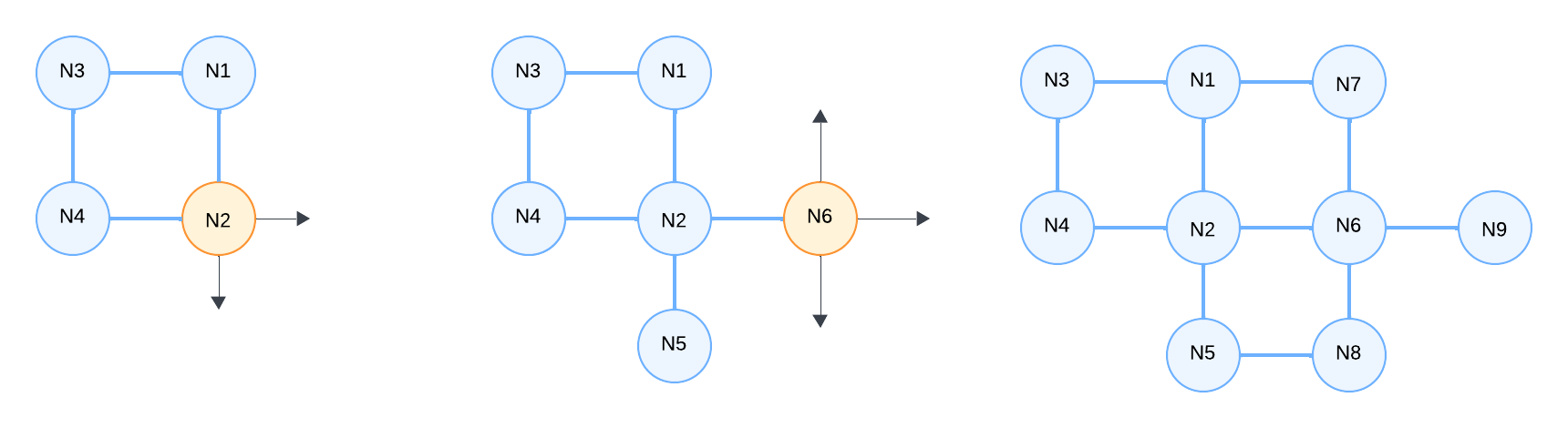}
    \caption{The proposed GSOM growth mechanism. When a unit (highlighted in yellow) exceeds the growth threshold, new units are added in its immediate $4$-neighborhood, i.e., at $(x, y \pm 1)$ and $(x \pm 1, y)$ if those positions are not already occupied.}
    \label{fig:gsom_growth}
    \vspace{-0.5cm}
\end{figure}

\section{Negative Log-Likelihood of a Multivariate Gaussian}
\label{app:nll_derivation}
The surprisal (negative log-likelihood) of a sample $z$ under a Gaussian
distribution $\mathcal{N}(\mu_b, \Sigma_b)$ is obtained by taking the
negative logarithm of the probability density function and applying
standard logarithmic identities.

\begin{align}
I_b(z)
&= - \log \left[
\frac{1}{(2\pi)^{d/2} |\Sigma_b|^{1/2}}
\exp\!\left(
-\frac{1}{2}(z-\mu_b)^T \Sigma_b^{-1}(z-\mu_b)
\right)
\right] \\
&= - \log \left( \frac{1}{(2\pi)^{d/2} |\Sigma_b|^{1/2}} \right)
- \log \left(
\exp\!\left(
-\frac{1}{2}(z-\mu_b)^T \Sigma_b^{-1}(z-\mu_b)
\right)
\right) \\
&= \log \left( (2\pi)^{d/2} |\Sigma_b|^{1/2} \right)
+ \frac{1}{2}(z-\mu_b)^T \Sigma_b^{-1}(z-\mu_b) \\
&= \frac{1}{2}(z-\mu_b)^T \Sigma_b^{-1}(z-\mu_b)
+ \frac{1}{2}\log |\Sigma_b|
+ \frac{d}{2}\log(2\pi).
\end{align}

\begin{algorithm}[H]
\scriptsize
\caption{Unified Algorithm for batch-wise Continual learning with the GSOM and optional encoder/decoder models.}
\label{alg:unified_algo}
\textbf{Input}: Stream of batches $\{\mathcal{B}_t\}_{t=1}^{T}$; flags \texttt{GLOBAL}; replay samples per BMU $K$ \\
\textbf{Output}: GSOM, (optional) global encoder/decoder
\begin{multicols}{2}
\begin{algorithmic}[1]
\STATE Initialize GSOM; initialize replay set $\mathcal{R}\leftarrow \emptyset$
\STATE \textbf{if} \texttt{GLOBAL} \textbf{then} initialize global encoder/decoder

\FOR{$t=1$ to $T$}
  \STATE $\mathcal{T}_t \leftarrow \mathcal{B}_t \cup \mathcal{R}$
  \vspace{0.25em}

  \STATE \textbf{// Encode features for GSOM update}
  \IF{\texttt{GLOBAL}}
    \STATE Train or update global encoder/decoder on $\mathcal{T}_t$
    \STATE $\mathcal{Z}_t \leftarrow \mathrm{Enc}_{\mathrm{global}}(\mathcal{T}_t)$
    \STATE Update GSOM with $\mathcal{Z}_t$ (including neighborhood and growth updates)
  \ELSE
    \STATE Update GSOM with $\mathcal{T}_t$ (including neighborhood and growth updates)
  \ENDIF
  \vspace{0.25em}

  \STATE $\mathcal{R}_{t} \leftarrow \emptyset$ \textbf{// Build replay for subsequent batches}
  \FOR{each active BMU $(i,j)$}
    \STATE Obtain BMU statistics $(\mu_{ij}, \Sigma_{ij})$
    \STATE $(\hat{\mu}_{ij}, \hat{\sigma}^2_{ij}, \hat{\Sigma}_{ij}) \leftarrow$
    \STATE \textsc{BiasCorrect}$(\mu_{ij}, \sigma^2_{ij}, \Sigma_{ij}, \beta, t_{ij}, \lambda)$
    \FOR{$k=1$ to $K$}
      \STATE $\tilde{z}\sim\mathcal{N}(\hat{\mu}_{ij}, \hat{\Sigma}_{ij})$
      \STATE $\tilde{x}\leftarrow
      \begin{cases}
      \mathrm{Dec}_{g}(\tilde{z}), & \texttt{GLOBAL}\\
      \tilde{z}, & \text{GSOM-only}
      \end{cases}$
      \STATE Append $\tilde{x}$ to $\mathcal{R}_{t}$
    \ENDFOR
  \ENDFOR
  \STATE Set $\mathcal{R}\leftarrow \mathcal{R}_{t}$
\ENDFOR

\STATE \textbf{return} GSOM, (global encoder/decoder if used)
\end{algorithmic}
\end{multicols}
\vspace{-0.375cm}
\end{algorithm}

\section{Running Distributional Statistic Bias Correction}
\label{app:bias_correction}
A practical issue with exponential moving averages, as computed by the GSOM units, is their bias towards the initial value encountered during early training. For instance, when updating the running mean via $\mu_{ij,t} = \alpha x_t + (1-\alpha)\mu_{ij,t-1},$
with initial condition $\mu_{ij,0}=0$, the estimate $\mu_{ij,t}$ tends to underestimate the true mean because it implicitly incorporates a zero-initialization. This issue is particularly pronounced in continuous learning settings, where some units may initially receive only a small number of samples. To mitigate this effect, we apply a bias correction similar to that used in the well-established Adam optimizer~\citep{kingma2017adammethodstochasticoptimization}. The corrected estimates for each unit are then as follows: 
\(
\begin{aligned}
\hat{\mu}_{ij,t} &= \frac{\mu_{ij,t}}{1 - \beta_\mu^t}, \quad
\hat{\sigma}^2_{ij,t} = \frac{\sigma^2_{ij,t}}{1 - \beta_\sigma^t}, \quad
\hat{\Sigma}_{ij,t} = \frac{\Sigma_{ij,t}}{1 - \beta_\Sigma^t}
\end{aligned}
\)
where $t$ denotes the number of BMU updates received by the unit to date and $\beta_\mu, \beta_\sigma, \beta_\Sigma \in [0,1)$ are exponential decay rates for the corresponding statistics. This correction ensures that the estimated parameters remain unbiased during early training, providing stable distribution estimates and well-conditioned covariances for both generative replay and surprisal evaluation. Unlike unbiased online estimators, such as Welford's algorithm \cite{welford1962note}, exponential updates are used because the distribution of samples mapped to a GSOM unit may evolve as the topology grows and change as a result of the data stream.

\section{GSOM Error Accumulation and Redistribution}
\label{appendix:gsom_error}

For each unit $b$, the accumulated quantization error is updated whenever
the unit is selected as the Best Matching Unit (BMU):
\begin{equation}
E_b \leftarrow E_b + \|z - w_b\|_2,
\end{equation}
where $z$ is the input sample and $w_b$ is the weight vector of unit $b$.
This formulation follows the standard GSOM definition, where the error
accumulates as the Euclidean distance between the input and the BMU weight.

When growth is triggered at a boundary unit, the accumulated error at that
unit is reset:
\begin{equation}
E_b \leftarrow 0.
\end{equation}

When an interior unit exceeds the growth threshold but no free neighbouring position exists in its four-neighbourhood $\mathcal{N}_4(b)$, direct expansion is not possible. In this case, the accumulated error is redistributed to its neighbours following the standard GSOM rule ~\citep{Alahakoon}:
\begin{equation}
E_b^{\mathrm{old}} = E_b,
\end{equation}
\begin{equation}
E_b \leftarrow \frac{1}{2} E_b^{\mathrm{old}}, \quad
E_n \leftarrow E_n + \frac{1}{8} E_b^{\mathrm{old}}
\quad \forall\, n \in \mathcal{N}_4(b),
\end{equation}
where $\mathcal{N}_4(b)$ denotes the set of the four adjacent grid neighbours
of unit $b$.

This redistribution conserves the total accumulated error: the BMU retains half of its error, while the remaining half is distributed equally among its four neighbours. This mechanism allows growth pressure to propagate from interior regions toward nearby boundary units where expansion is feasible.

\section{Encoder-Decoder per BMU setup}
\label{app:per_bmu_enc_dec}

While a global encoder–decoder uses single model memory only and benefits from training on the full dataset, it must generalize across diverse and often underrepresented regions of the latent space. In contrast, the per-BMU approach trains localized models using samples assigned to each unit, enabling region-specific specialization at the cost of additional memory (Algorithm ~\ref{alg:per_bmu_algo}). After mapping inputs to their corresponding BMUs, the associated samples are used to train local models, whose decoders focus on reconstructing data from specific regions of the latent space. During replay, latent samples drawn from GSOM statistics are decoded using the corresponding local model rather than a shared decoder. The motivation for this design is to align the generative capacity with the SOM topology, allowing each decoder to specialize in a local region of the latent space and improving the quality of reconstruction. However, this approach introduces additional memory overhead due to maintaining a separate model per active unit, and may become less scalable as the number of units increases. The results presented in the main paper for this variant are intended as an initial exploration of its potential with further discussion on scalability and potential improvements provided in the Discussion and Future Work section.

\begin{algorithm}[tb]
\scriptsize
\caption{Unified Algorithm for Batch-Wise Continual Learning with GSOM with Encoder/Decoder Models per GSOM unit }
\label{alg:per_bmu_algo}
\textbf{Input}: Stream of batches $\{\mathcal{B}_t\}_{t=1}^{T}$; replay samples per BMU $K$ \\
\textbf{Output}: GSOM, global encoder/decoder, per-BMU encoder/decoder models $\mathcal{V}$
\begin{multicols}{2}
\begin{algorithmic}[1]
\STATE Initialize GSOM; initialize replay set $\mathcal{R}\leftarrow \emptyset$
\STATE Initialize global encoder/decoder $(\mathrm{Enc}_g,\mathrm{Dec}_g)$
\STATE Initialize per-BMU model dictionary $\mathcal{V}\leftarrow \emptyset$

\FOR{$t=1$ to $T$}
  \STATE $\mathcal{T}_t \leftarrow \mathcal{B}_t \cup \mathcal{R}$
  \vspace{0.25em}

    \STATE Train or update global encoder/decoder on $\mathcal{T}_t$
    \STATE $\mathcal{Z}_t \leftarrow \mathrm{Enc}_{\mathrm{global}}(\mathcal{T}_t)$
    \STATE Update GSOM with $\mathcal{Z}_t$ (including neighborhood and growth updates)

  \STATE \textbf{// Train per-BMU local models on assigned subsets}
    \STATE Assign each $x\in\mathcal{T}_t$ to BMU $(i,j)$ using its current representation 
    \STATE \hspace{1em} ($\mathcal{Z}_t$ if global encoder is used, else raw $\mathcal{T}_t$)
    \FOR{each BMU $(i,j)$ with assigned set $\mathcal{S}_{ij}\subseteq\mathcal{T}_t$}
      \STATE Train or update local model $\mathcal{M}_{ij}$ on $\mathcal{S}_{ij}$
      \STATE Store/refresh $\mathcal{V}[(i,j)] \leftarrow \mathcal{M}_{ij}$
      \STATE Re-encode $\mathcal{S}_{ij}$ with the encoder of $\mathcal{M}_{ij}$ to refine local GSOM updates
    \ENDFOR
  \vspace{0.25em}

  \STATE \textbf{// Build replay for subsequent batches}
  \STATE $\mathcal{R}_{t} \leftarrow \emptyset$
  \FOR{each BMU $(i,j)$ such that $(i,j)\in\mathcal{V}$}
    \STATE Obtain BMU statistics $(\mu_{ij},\sigma^2_{ij},\Sigma_{ij})$
    \STATE $(\hat{\mu}_{ij},\hat{\sigma}^{2}_{ij},\hat{\Sigma}_{ij}) \leftarrow$
    \STATE \hspace{1em}\textsc{BiasCorrect}$(\mu_{ij},\sigma^2_{ij},\Sigma_{ij},\beta,t_{ij},\lambda)$

    \FOR{$k=1$ to $K$}
      \STATE $\tilde{z}\sim\mathcal{N}(\hat{\mu}_{ij},\hat{\Sigma}_{ij})$
      \STATE $\tilde{x}\leftarrow \mathrm{Dec}_{ij}(\tilde{z})$
      \STATE Append $\tilde{x}$ to $\mathcal{R}_{t}$
    \ENDFOR
  \ENDFOR
  \STATE Set $\mathcal{R}\leftarrow \mathcal{R}_{t}$
\ENDFOR

\STATE \textbf{return} GSOM, global $(\mathrm{Enc}_g,\mathrm{Dec}_g)$, per-BMU models $\mathcal{V}$
\end{algorithmic}
\end{multicols}
\end{algorithm}

\section{Experimental Settings}
\label{app:experimental_settings}

Table~\ref{tab:unified_hyperparams} summarizes the hyperparameter configurations used across all GSOM-based experiments. Unless otherwise specified, GSOM units were initialized using random samples drawn from the input data or latent representations of the first batch of data observed during training, consistent with the class-incremental learning setup. Alternative initialization strategies, including uniform sampling over $[0,1]$ and initialization using the global mean of the data, were also evaluated but resulted in inferior performance. For each GSOM unit, the associated statistics were initialized as follows: the mean vector was initialized to zero, the variance vector to ones, and the covariance matrix to zero. We additionally explored diagonal covariance initialization and initialization using latent statistics from a pretrained VAE; however, these alternatives did not provide consistent improvements. During training, exponential moving averages were maintained for the mean, variance, and covariance of each unit. The Best Matching Units (BMUs) were identified using Euclidean distance between the input (or latent) vector and GSOM unit weights. While other distance metrics, including cosine similarity, Manhattan (L1) distance, and Mahalanobis distance, were explored, Euclidean distance as a measure for the BMUs consistently provided the most stable and accurate results across datasets. The following hyperparameter settings were explored for the bias correction:

\begin{table}[H]
\footnotesize
\centering
\caption{Bias-correction hyperparameters with best-performing values and additional settings tested.}
\label{tab:beta_values}
\begin{tabular}{lccc}
\toprule
\textbf{Parameter} & \textbf{Symbol} & \textbf{Best value} & \textbf{Other values tested} \\
\midrule
Mean               & $\beta_\mu$     & $0.99$              & $0.98$, $0.95$ \\
Variance / Cov.    & $\beta_\sigma, \beta_\Sigma$ & $0.95$ & $0.91$, $0.99$, $0.90$ \\
\bottomrule
\end{tabular}
\end{table}

Unlike fixed SOMs, GSOM dynamically expands its topology during training. To control model growth, expansion was constrained using two hyperparameters: the number of growth iterations ($I_{\text{grow}}$) and the maximum number of units added per iteration ($G_{\text{max}}$). Growth was only permitted during the initial $I_{\text{grow}}$ iterations of each batch, after which the map structure was fixed for the remaining training iterations. This design prevents uncontrolled expansion while allowing the model to adapt to new data.

We observed that increasing growth flexibility (i.e., larger $I_{\text{grow}}$ and $G_{\text{max}}$) improves representational capacity and accuracy but leads to higher computational and memory costs. Conversely, more restrictive growth produces compact maps with faster training at the expense of reduced performance. For all reported experiments, we adopt conservative growth settings (e.g., $I_{\text{grow}} = 5$, $G_{\text{max}} = 4$) to balance accuracy and efficiency. Growth decisions were driven by a Mahalanobis-based surprisal criterion computed using the running statistics of each unit. The growth threshold was determined adaptively to ensure stable, data-driven expansion. The threshold was set to the 92nd percentile of the most recent values after sufficient observations had been collected, allowing growth sensitivity to adjust to the evolving data distribution.

\begin{algorithm}
\scriptsize
\caption{Synthetic Sample Generation using Mean and Covariance (bias-corrected, Appendix~\ref{app:bias_correction})}
\begin{algorithmic}[1]
\REQUIRE Trained SOM, BMU coordinates $(i,j)$, number of samples $n$, regularization constant $\epsilon$
\STATE $\mu \leftarrow \text{GSOM}.\text{running\_mean}[i][j]$
\STATE $\Sigma \leftarrow \text{GSOM}.\text{running\_cov}[i][j]$
\STATE $\Sigma \leftarrow \Sigma + \epsilon I$ \hfill // Add diagonal regularization
\STATE Compute eigen-decomposition: $Q \Lambda Q^\top = \Sigma$
\STATE $\Lambda \leftarrow \max(\Lambda, \epsilon)$ \hfill // Clamp eigenvalues
\STATE $\Sigma_{\text{pd}} \leftarrow Q \cdot \text{diag}(\Lambda) \cdot Q^\top$
\STATE Sample $\tilde{z} \sim \mathcal{N}(\mu, \Sigma_{\text{pd}})$
\RETURN $n$ samples $\{\tilde{z}_1, ..., \tilde{z}_n\}$
\end{algorithmic}
\label{app:covar-synth-alg}
\end{algorithm}

For generative configurations, synthetic samples were generated using the estimated mean and covariance of each GSOM unit, following the procedure described in Algorithm~\ref{app:covar-synth-alg}. VAE hyperparameters, including latent dimensionality, learning rate, and batch size, were selected based on reconstruction quality and training stability. In pretrained encoder-based setups (e.g., CLIP ViT-B/32 and ResNet-18), embeddings were kept frozen by default to preserve learned representations, with optional fine-tuning of the final layers. These configurations provided stable performance across datasets while maintaining reasonable computational cost.

\begin{table*}[t]
\centering
\scriptsize
\caption{Unified hyperparameter configuration across all experiments. Shared GSOM parameters are consistent unless otherwise specified.}
\label{tab:unified_hyperparams}
\begin{tabular}{lll}
\toprule
\textbf{Component} & \textbf{Parameter} & \textbf{Value} \\
\midrule

\multicolumn{3}{l}{\textbf{Datasets}} \\
\midrule
& MNIST & $28 \times 28$ (grayscale) \\
& CIFAR-10 / CIFAR-100 & $32 \times 32 \times 3$ \\
& TinyImageNet / MiniImageNet & $64 \times 64 \times 3$ / $84 \times 84 \times 3$ \\

\midrule
\multicolumn{3}{l}{\textbf{Shared GSOM Parameters}} \\
\midrule
& Initial Map Size & $2 \times 2$ (MNIST), $5 \times 5$ (CIFAR10)\\
& & $7 \times 7$ (CIFAR100/Mini/TinyImagenet)  \\
& Max Nodes & 2000 \\
& SOM Iterations per Batch/Class & 30 (MNIST), 100 (Others) \\
& Learning Rate & 0.4--0.5 \\
& Sigma (Neighborhood Width) & 0.9--0.96 \\
& Neighborhood Function & Gaussian \\
& Distance Metric & Euclidean \\
& Batch Size & 32 (MNIST), 64--128 (others) \\

\midrule
\multicolumn{3}{l}{\textbf{Growth Parameters}} \\
\midrule
& Spread Factor & 0.7--0.8 \\
& Growth Iterations ($E_{\text{grow}}$) & 5 (MNIST), 9 (ResNet/VAE), 10 (CLIP) \\
& Max Growth per Iterations ($G_{\text{max}}$) & 2 (MNIST), 4 (ResNet/VAE), 14 (CLIP) \\
& Min BMU Hits & 5--7 \\
& Growth Cooldown & 40 iterations \\
& Mahalanobis Threshold & 92--93 percentile (adaptive) \\
& Growth Probability & Linearly decayed (1.0 $\rightarrow$ 0.5) \\
& Error Decay & 0.85--0.88 \\

\midrule
\multicolumn{3}{l}{\textbf{Replay and Statistics}} \\
\midrule
& Sampling Method & Gaussian (per-BMU mean/covariance) \\
& Covariance Regularization & $\lambda = 1\times10^{-4}$ \\
& Replay Samples per BMU & $K = 1$ \\

\midrule
\multicolumn{3}{l}{\textbf{VAE Configuration}} \\
\midrule
& Latent Dimension & 32, 64, 128 \\
& Iterations & 200 \\
& Batch Size & 128, 256 \\
& Learning Rate & $1e{-4}$ to $1e{-5}$ \\
& Optimizer & Adam \\
& Architecture & ResNet-style VAE \\
& Loss & Reconstruction + KL \\

\midrule
\multicolumn{3}{l}{\textbf{ResNet-18 Configuration}} \\
\midrule
& Encoder & ResNet-18 (ImageNet pretrained) \\
& Embedding Dimension & 512 \\
& Input Resolution & $224 \times 224$ \\
& Fine-tuning & Frozen / last block FT \\
& Decoder & Conv / transposed conv \\
& Training & End-to-end with replay \\

\midrule
\multicolumn{3}{l}{\textbf{CLIP Configuration}} \\
\midrule
& Encoder & CLIP ViT-B/32 (OpenAI) \\
& Embedding Dimension & 512 \\
& Input Resolution & $224 \times 224$ (upsampled) \\
& Fine-tuning & Frozen / last block FT \\
& Decoder & Transposed conv (attention-based) \\
& Feature Handling & Pooled embedding \\

\bottomrule
\end{tabular}
\end{table*}

\section{Ablation Studies}
\label{app:abalation_study}

\subsection{GSOM versus Fixed-Capcity SOM}

The results in Table~\ref{tab:fixed_vs_gsom_side} compare fixed SOM and GSOM under bias-corrected settings across datasets and configurations. While the fixed SOM generally achieves higher accuracy—particularly in the CLIP-based setting—this is expected due to differences in the learning paradigm. The fixed SOM operates in a task-aware manner, with replay performed at the end of each class, thereby effectively leveraging known task boundaries. In contrast, GSOM is fully task-free and performs replay continuously at the batch level without access to class information, making the learning problem inherently more challenging due to the absence of explicit distribution shift signals.

In addition to accuracy, the table highlights differences in representational capacity. The fixed SOM uses a constant $40 \times 40$ topology (1600 units) across all datasets, whereas GSOM dynamically adapts its size based on data complexity. As observed, GSOM produces significantly more compact maps for simpler datasets (e.g., 357–376 units for MNIST), while scaling to larger sizes for more complex datasets (e.g., 956–1392 units for VAE-based settings and up to 1508 units for CLIP-based settings). 

Overall, these results highlight a trade-off between performance and flexibility. While fixed SOM benefits from task-aware replay and a fixed high-capacity topology, GSOM offers a fully task-agnostic alternative with adaptive capacity, achieving competitive performance while efficiently allocating representational resources based on the data.

\begin{table}[t]
\centering
\scriptsize
\caption{Comparison of fixed SOM and GSOM (with bias) across datasets. Results are reported as accuracy (\%), with the number of best-matching units (BMUs) after final training shown in parentheses. The fixed SOM uses a $40 \times 40$ grid (1600 units) for all datasets.}
\label{tab:fixed_vs_gsom_side}
\begin{tabular}{llcccc}
\toprule
\textbf{Method} 
& \textbf{Model} 
& \textbf{Split-MNIST} 
& \textbf{Split-CIFAR-10} 
& \textbf{Split-CIFAR-100} 
& \textbf{Split-TinyImageNet} \\
\midrule

\multicolumn{6}{l}{\textbf{SOM-only}} \\
\midrule
Fixed & SOM-only 
& 92.11 $\pm$ 1.18 (1600)
& -- 
& -- 
& -- \\
GSOM & GSOM-only 
& 91.22 $\pm$ 1.23 (357)
& -- 
& -- 
& -- \\

\midrule
\multicolumn{6}{l}{\textbf{VAE (Global FT)}} \\
\midrule
Fixed & SOM + VAE 
& 91.23 $\pm$ 1.21 (1600)
& 53.01 $\pm$ 1.18 (1600)
& 14.63 $\pm$ 0.33 (1600)
& 12.74 $\pm$ 0.84 (1600) \\
GSOM & GSOM + VAE 
& 92.11 $\pm$ 0.98 (376)
& 50.11 $\pm$ 0.89 (956)
& 13.81 $\pm$ 1.05 (1167)
& 6.22 $\pm$ 0.67 (1241) \\

\midrule
\multicolumn{6}{l}{\textbf{CLIP (Global FT)}} \\
\midrule
Fixed & SOM + CLIP 
& -- 
& 81.24 (1600)
& 67.19 (1600)
& 60.13 (1600) \\
GSOM & GSOM + CLIP 
& 92.23 $\pm$ 1.36 (428)
& 78.84 $\pm$ 1.52 (1034)
& 65.01 $\pm$ 1.41 (1262)
& 58.25 $\pm$ 1.02 (1408) \\

\bottomrule
\end{tabular}
\end{table}

\subsection{Growth Sensitivity}
The growth in GSOM is controlled by two key hyperparameters: the number of growth iterations ($I_{\text{grow}}$) and the maximum number of units added per iteration ($G_{\text{max}}$). These parameters directly influence the trade-off between representational capacity and performance. Increasing $I_{\text{grow}}$ and $G_{\text{max}}$ allows the map to expand more aggressively, leading to larger topologies and improved accuracy due to enhanced representation of incoming data. However, excessive growth leads to disproportionately large maps with diminishing returns in performance, increasing memory and computational cost without significant accuracy gains. Conversely, restricting growth results in more compact maps with reduced computational cost but lower accuracy. We use $I_{\text{grow}} = 9$ and $G_{\text{max}} = 4$ as the baseline configuration for all primary comparisons. As shown in Table~\ref{tab:growth_sensitivity}, GSOM demonstrates consistent behavior across datasets, where increased growth flexibility improves performance at the expense of model size. The selected baseline configuration balances this trade-off, providing stable accuracy with controlled expansion.

\begin{table}[t]
\centering
\scriptsize
\caption{Growth sensitivity analysis of GSOM with global ResNet-18. Increasing growth parameters improves accuracy but increases map size. Results are reported as accuracy (\%), with the number of best-matching units (BMUs) after final training shown in parentheses.}
\label{tab:growth_sensitivity}
\begin{tabular}{lccccc}
\toprule
\textbf{$E_{\text{grow}}$} & \textbf{$G_{\text{max}}$} & \textbf{Split-MNIST} & \textbf{Split-CIFAR-10} & \textbf{Split-CIFAR-100} & \textbf{Split-TinyImageNet} \\
\midrule

6 & 3 & 95.01 (432) & 57.41 (1012) & 35.01 (1284) & 31.74 (1386) \\
7 & 3 & 95.26 (468) & 57.72 (1089) & 35.32 (1352) & 32.05 (1438) \\
9 & 4 & 96.12 (498) & 58.00 (956) & 35.67 (1161) & 32.12 (1312) \\
10 & 5 & 96.42 (548) & 58.11 (1184) & 35.74 (1426) & 32.28 (1502) \\
20 & 14 & 96.45 (712) & 58.18 (1621) & 35.81 (1895) & 32.36 (2143) \\
\bottomrule
\end{tabular}
\end{table}

\subsection{Growth Criterion}

Table ~\ref{tab:growth_ablation} compares three variants of the growth trigger: accumulated quantization error alone ($E_b>T_E$, equivalent to standard GSOM \citep{Alahakoon}), Mahalanobis surprisal alone ($S_b(z) > T_S$), and the proposed combined criterion. The results indicate that the original GSOM growth rule based solely on accumulated quantization error yields the lowest performance across all datasets. Replacing the error criterion with the proposed surprisal measure improves performance, suggesting that distribution-aware growth better captures emerging data structure. The combined criterion achieves the strongest overall results, indicating that the quantization error term acts as a stabilizing gate while the surprisal term enables more adaptive topology expansion.

\begin{table}[h]
\centering
\small
\caption{Ablation study of GSOM growth criteria across continual learning benchmarks on Split datasets.}
\label{tab:growth_ablation}
\begin{tabular}{lccc}
\hline
\textbf{Method} & \textbf{MNIST} & \textbf{CIFAR-10} & \textbf{CIFAR-100} \\
\hline
Global VAE-GSOM (Error only) 
& $89.42 \pm 1.21$ 
& $44.87 \pm 1.03$ 
& $10.92 \pm 0.88$ \\

Global VAE-GSOM (Surprisal only) 
& $91.26 \pm 1.04$ 
& $48.63 \pm 0.95$ 
& $12.97 \pm 0.96$ \\

Global VAE-GSOM (Error + Surprisal) 
& $\mathbf{92.11 \pm 0.98}$ 
& $\mathbf{50.11 \pm 0.89}$ 
& $\mathbf{13.81 \pm 1.05}$ \\
\hline
\end{tabular}
\end{table}

\subsection{Studying Bias versus No Bias Correction}
We evaluate the impact of bias correction on the running statistics used within GSOM. Without bias correction, exponential moving averages are biased toward their initial values, particularly during early training when units receive limited updates, which leads to underestimation of the true mean and poorly conditioned variance and covariance estimates, which negatively affect both generative replay and surprisal-based growth decisions.

As observed in our experiments (Table~\ref{tab:bias_vs_nobias}), the bias-corrected variant consistently outperforms the non-corrected version across all datasets. Although the improvements are modest, they are consistent, indicating that bias correction provides more reliable statistical estimates and improves overall training stability.

\begin{table}[t]
\centering
\scriptsize
\caption{Effect of bias correction across GSOM + ResNet-18 configurations. Results are reported as accuracy (\%).}
\label{tab:bias_vs_nobias}
\begin{tabular}{llcccc}
\toprule
\textbf{Setting} & \textbf{Model} 
& \textbf{Split-MNIST} 
& \textbf{Split-CIFAR10} 
& \textbf{Split-CIFAR100} 
& \textbf{Split-MiniImageNet} \\
\midrule

\multicolumn{6}{l}{\textbf{No Bias}} \\
\midrule
& Global 
& 95.84 $\pm$ 0.41 
& 57.21 $\pm$ 1.44 
& 34.72 $\pm$ 2.03 
& 31.58 $\pm$ 1.37 \\

& Global FT 
& 96.58 $\pm$ 0.52 
& 57.34 $\pm$ 1.51 
& 34.91 $\pm$ 1.39 
& 32.71 $\pm$ 1.53 \\

& BMU Specific 
& 93.61 $\pm$ 1.28 
& 56.32 $\pm$ 1.67 
& 34.02 $\pm$ 1.88 
& 31.74 $\pm$ 1.21 \\

\midrule
\multicolumn{6}{l}{\textbf{Bias}} \\
\midrule
& Global 
& 96.12 $\pm$ 0.30 
& 58.00 $\pm$ 1.31 
& \textbf{35.67 $\pm$ 1.96} 
& 32.67 $\pm$ 1.23 \\

& Global FT 
& \textbf{97.21 $\pm$ 0.45} 
& \textbf{58.10 $\pm$ 1.42} 
& 35.33 $\pm$ 1.28 
& \textbf{33.60 $\pm$ 1.45} \\

& BMU Specific 
& 94.34 $\pm$ 1.10 
& 57.11 $\pm$ 1.56 
& 34.89 $\pm$ 1.72 
& 32.50 $\pm$ 1.06 \\

\bottomrule
\end{tabular}
\end{table}

\section{Generative Capability of the Proposed Method}
\label{app:generative_model}
\begin{figure}[!t]
    \centering
    \includegraphics[width=0.45\linewidth]{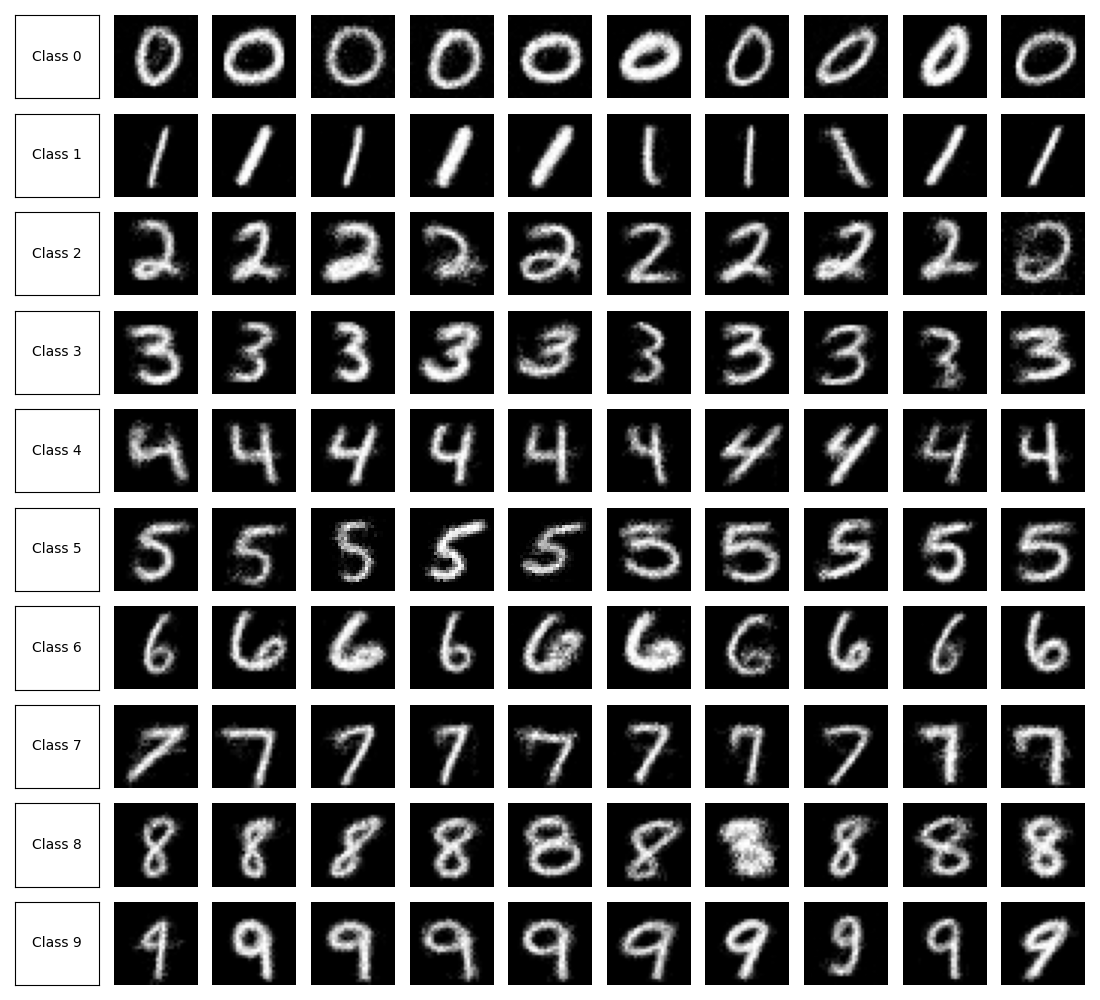}
    \caption{Synthetic samples generated using Gaussian sampling from the trained SOM. Each row corresponds to one class (0–9), and each row contains 10 synthetic samples generated from BMUs labeled with that class.}
    \label{fig:som_synthetic_samples}
\end{figure}

Our proposed model can also be used as a generative model, which is capable of synthesizing samples without storing original data. For MNIST, as we only use the SOM for training, image samples are generated using Gaussian sampling as described in the paper. After training, each SOM unit stores the specific statistics in the form of mean and variance, and they are used to generate sample images. Figure~\ref{fig:som_synthetic_samples} illustrates some of the samples generated for each class using this method.

\begin{figure}[!t]
    \centering
    \subfloat[CIFAR10]{%
        \includegraphics[width=0.45\linewidth]{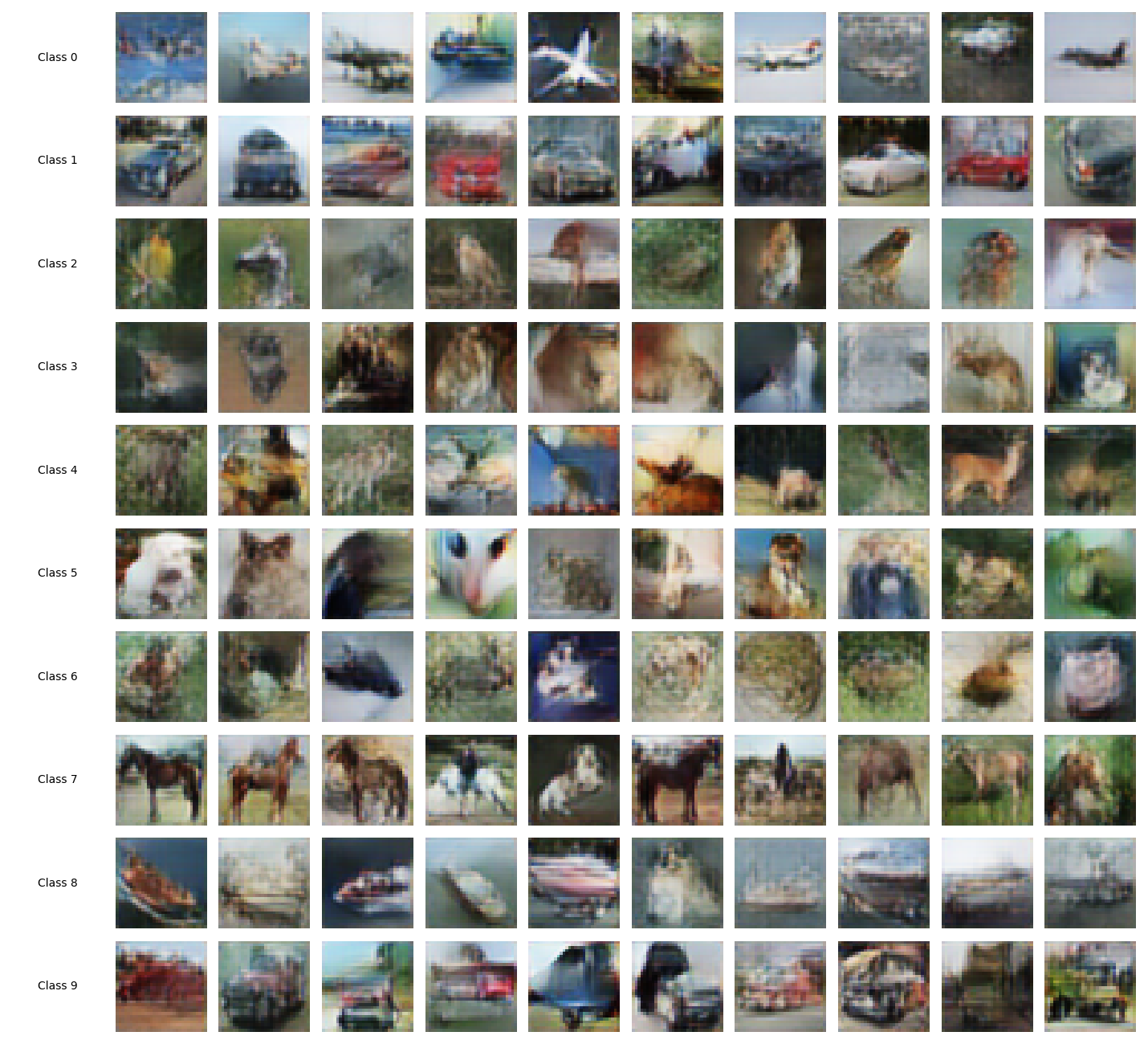}
        \label{fig:som_synthetic_mnist}
    }
    \hfill
    \subfloat[CIFAR100]{%
        \includegraphics[width=0.45\linewidth]{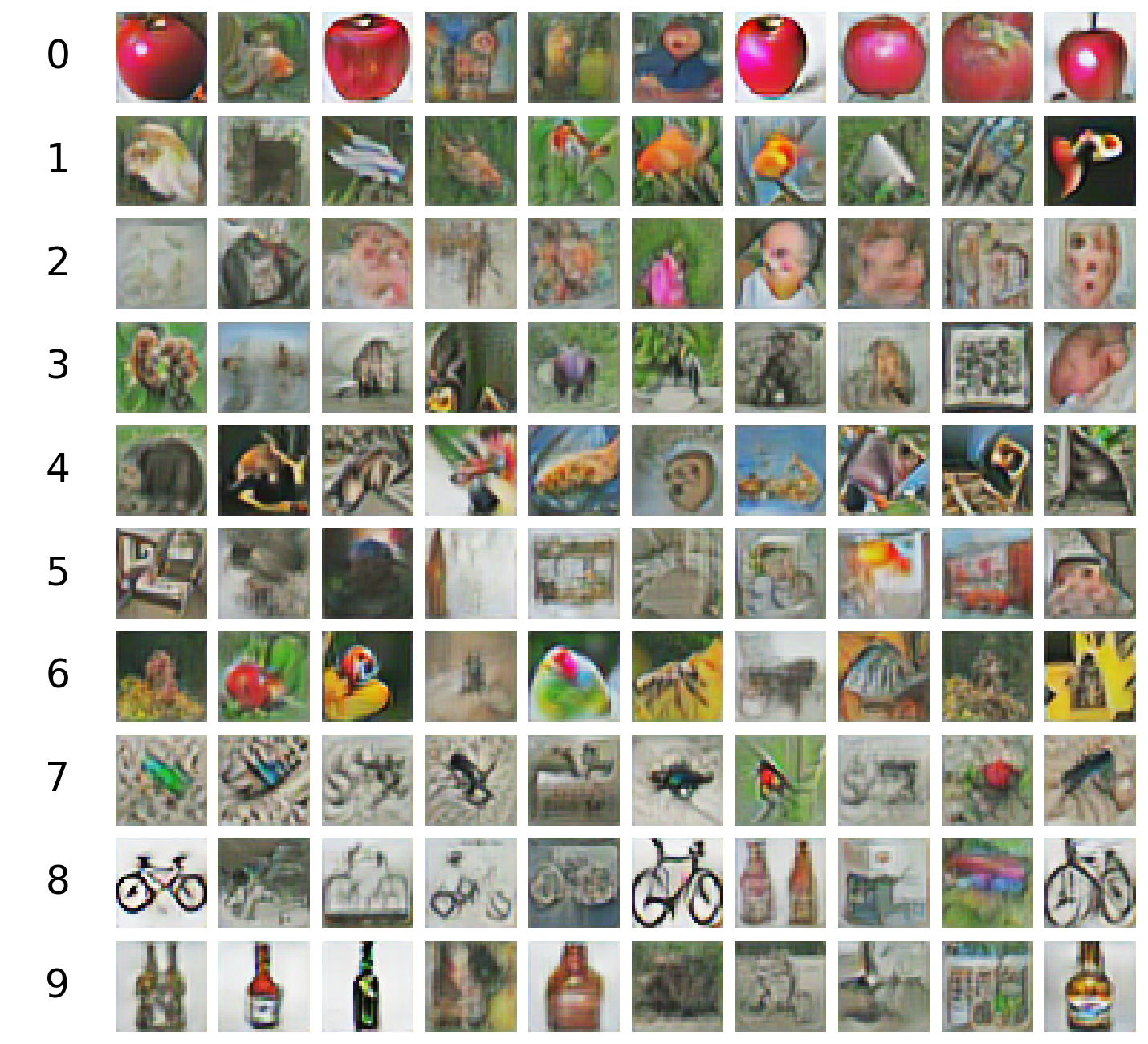}
        \label{fig:som_synthetic_fmnist}
    }
    \caption{
        Synthetic CIFAR-10 samples generated by sampling latent vectors from a Self-Organizing Map (SOM) trained on VAE latent space. Each latent vector was sampled from the full-covariance Gaussian distribution of a Best Matching Unit (BMU) labeled with the target class. The sampled latents were then decoded into images using VAE. Each row corresponds to one class (0–9).
    }
    \label{fig:som_synthetic_cifar10}
\end{figure}

For other datasets, we combine a VAE and the SOM trained on the VAE's latent space, where, after training, the SOM statistics store the mean, variance, and full covariance matrix of the latent samples. Synthetic latent samples are then generated using eigen-decomposition of the covariance, and then decoded using the VAE decoder to reconstruct the images. Figure~\ref{fig:som_synthetic_cifar10} and ~\ref{fig:som_synthetic_cifar100} illustrate some of the samples generated for each class using this method.

\section{Class Level Analysis}
\label{app:c_matrix}

\begin{figure*}[t]
\centering
\begin{subfigure}{0.44\textwidth}
    \includegraphics[width=\linewidth]{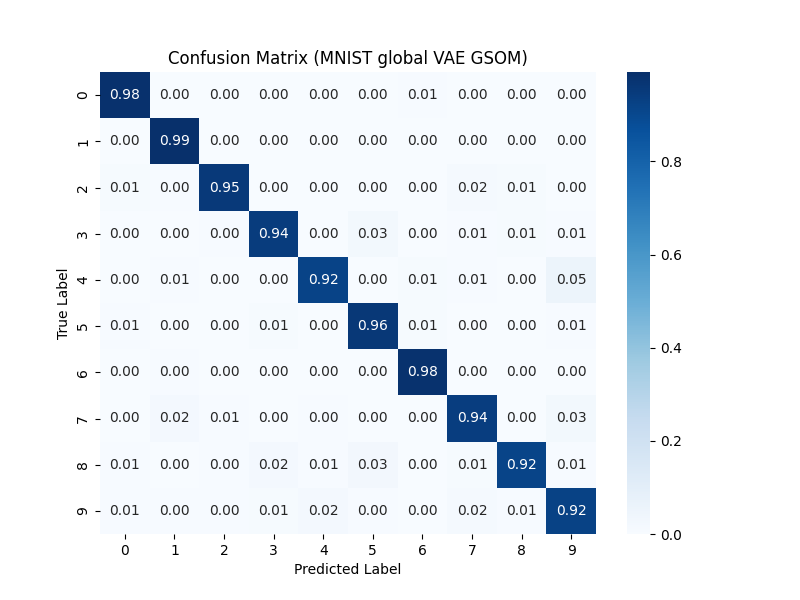}
    \caption{MNIST}
\end{subfigure}
\hfill
\begin{subfigure}{0.44\textwidth}
    \includegraphics[width=\linewidth]{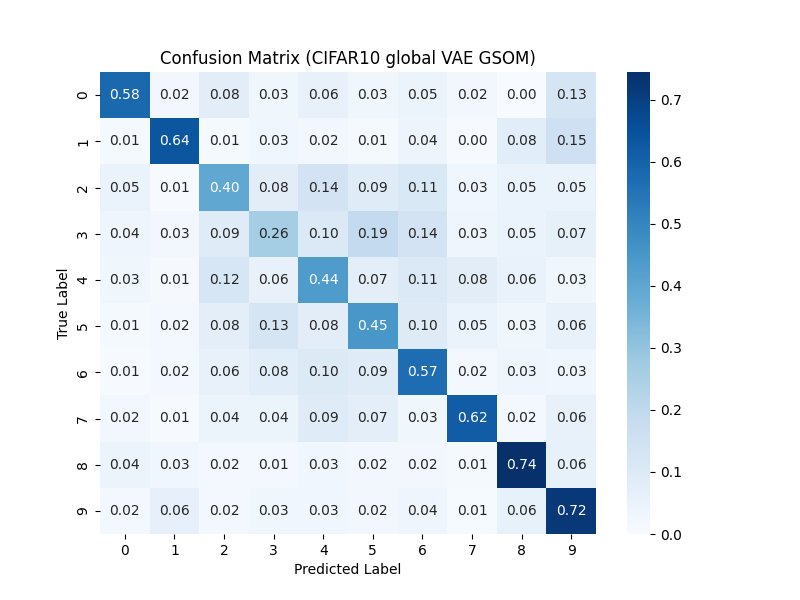}
    \caption{CIFAR-10}
\end{subfigure}
\hfill
\begin{subfigure}{0.42\textwidth}
    \includegraphics[width=\linewidth]{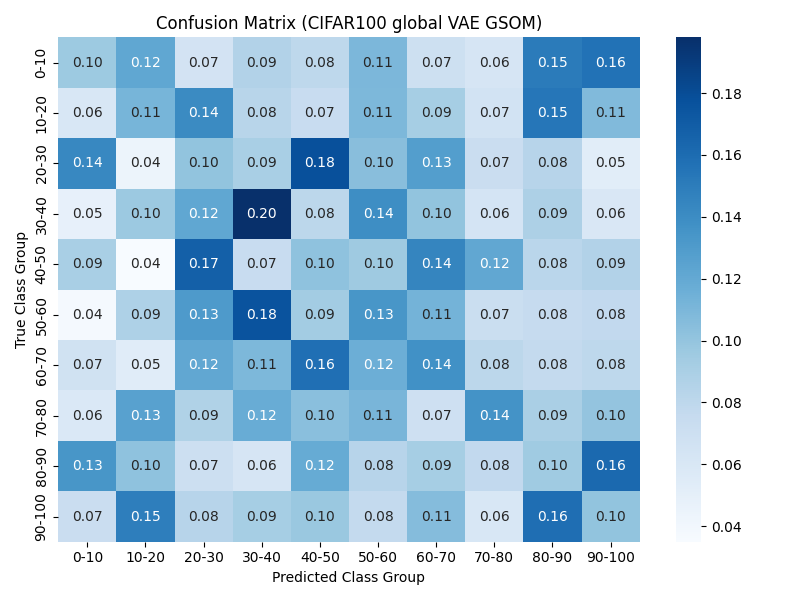}
    \caption{CIFAR-100}
\end{subfigure}
\hfill
\begin{subfigure}{0.42\textwidth}
    \includegraphics[width=\linewidth]{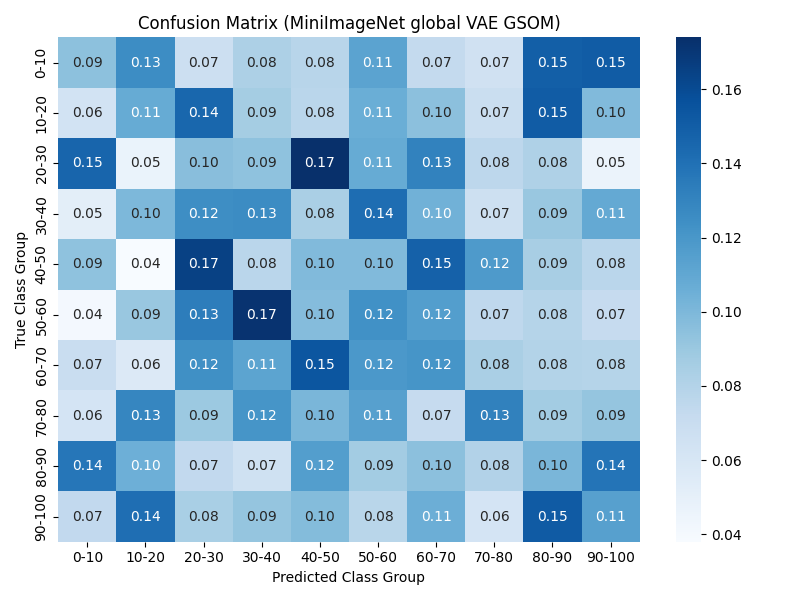}
    \caption{MiniImageNet}
\end{subfigure}
\hfill
\begin{subfigure}{0.44\textwidth}
    \includegraphics[width=\linewidth]{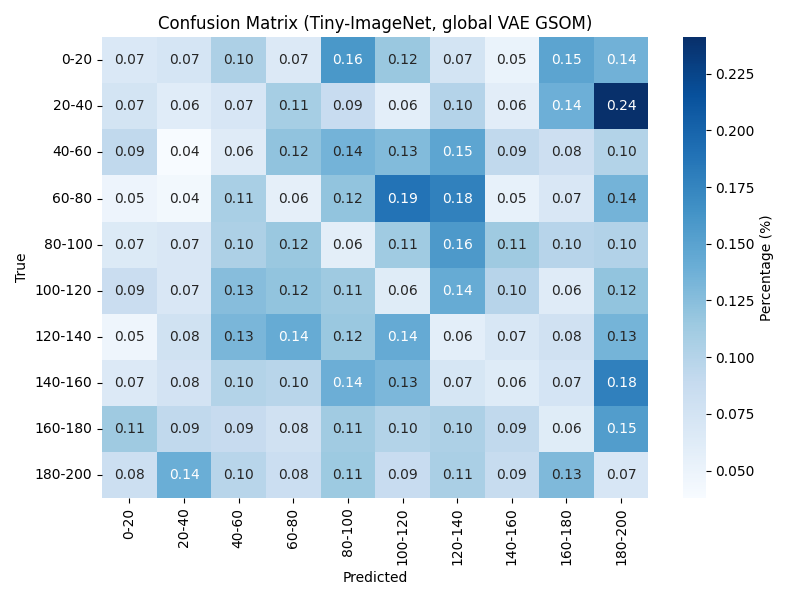}
    \caption{TinyImageNet}
\end{subfigure}

\caption{Row-normalized confusion matrices for the global VAE–GSOM model across datasets. Each row shows the prediction distribution for a given true class.}
\label{fig:confusion_all}
\vspace{-5mm}
\end{figure*}

To further analyze model behavior, we present row-normalized confusion matrices under the single-class incremental learning (SCIL) setting in Figure~\ref{fig:confusion_all}.
We analyze the global VAE–GSOM behavior using row-normalized confusion matrices, which reveal how prediction mass is distributed for each true class as dataset complexity increases. For larger datasets (CIFAR-100, MiniImageNet, and TinyImageNet), classes are grouped into coarse categories to improve interpretability.

For MNIST, predictions are almost entirely concentrated along the diagonal, indicating strong class separability and effective retention with minimal interference. On CIFAR-10, the diagonal remains dominant but becomes broader, with noticeable confusion among visually similar classes, reflecting partial overlap in learned representations. This spread increases for CIFAR-100 and MiniImageNet, where predictions are more distributed across related groups. While correct classes still receive higher mass, the diagonals are less pronounced, and structured cross-group confusion becomes more evident. On TinyImageNet, the most challenging setting, the matrix is highly diffuse, with weak diagonal dominance and significant mixing across groups. Despite this, the predictions are not entirely random, and some residual structure is preserved.

Overall, these matrices show a gradual shift from concentrated, confident predictions to more distributed outputs as task difficulty increases, highlighting both the robustness and the limitations of replay in complex, fine-grained scenarios.

\section{Continual Learning Metrics}
\label{app:cl_metrices}

Tables~\ref{tab:R-matrix-cifar10-gsom}--\ref{tab:metrics-cifar100-gsom}
report the full per-task accuracy matrices and corresponding continual learning metrics for CIFAR-10 and CIFAR-100 using the GSOM with a globally trained VAE (from scratch). We evaluate forward transfer (FWT), backward transfer (BWT), and forgetting using the standard formulation (\citep{lopezpaz2022gradientepisodicmemorycontinual}).

Across both datasets, the structure of the $R$ matrices reveals a gradual degradation of performance along the rows, indicating the presence of catastrophic forgetting. This effect is relatively mild for CIFAR-10, where earlier tasks retain moderate accuracy, but it becomes significantly more pronounced for CIFAR-100 due to increased task complexity and class diversity. Since the model operates in a single-head, fully unsupervised class-incremental setting, it has no prior capability on unseen classes, resulting in $R_{t-1,t} \approx 0$ before learning task $t$. Consequently, FWT is consistently negative and approximately equal in magnitude to the per-class baseline accuracy, confirming the absence of forward transfer. Backward transfer (BWT) is negative for early tasks, reflecting degradation in performance as new tasks are learned. For later tasks, BWT approaches zero and occasionally becomes slightly positive, indicating reduced interference and minor stabilization effects. Forgetting follows a similar trend: earlier tasks exhibit higher forgetting due to cumulative updates, while later tasks show reduced forgetting as fewer subsequent updates occur. The final task does not experience forgetting, as no further updates are applied after it is learned.

\begin{table}[h]
\centering
\small
\caption{Per-task class accuracy matrix $R$ for CIFAR-10 across 10 sequential tasks for GSOM with a global VAE (latent dimension $32\times2\times2$).}
\label{tab:R-matrix-cifar10-gsom}
\begin{tabular}{c|cccccccccc}
\hline
\textbf{t $\backslash$ j} & 0 & 1 & 2 & 3 & 4 & 5 & 6 & 7 & 8 & 9 \\
\hline
0 & 1 & -- & -- & -- & -- & -- & -- & -- & -- & -- \\
1 & 0.823 & 0.781 & -- & -- & -- & -- & -- & -- & -- & -- \\
2 & 0.692 & 0.724 & 0.703 & -- & -- & -- & -- & -- & -- & -- \\
3 & 0.641 & 0.701 & 0.553 & 0.582 & -- & -- & -- & -- & -- & -- \\
4 & 0.613 & 0.693 & 0.427 & 0.521 & 0.503 & -- & -- & -- & -- & -- \\
5 & 0.592 & 0.674 & 0.391 & 0.402 & 0.481 & 0.434 & -- & -- & -- & -- \\
6 & 0.571 & 0.651 & 0.364 & 0.352 & 0.443 & 0.423 & 0.552 & -- & -- & -- \\
7 & 0.553 & 0.632 & 0.352 & 0.341 & 0.412 & 0.401 & 0.541 & 0.603 & -- & -- \\
8 & 0.482 & 0.604 & 0.331 & 0.332 & 0.391 & 0.392 & 0.553 & 0.592 & 0.703 & -- \\
9 & 0.504 & 0.583 & 0.351 & 0.253 & 0.423 & 0.421 & 0.562 & 0.603 & 0.712 & 0.701 \\
\hline
\end{tabular}
\end{table}

\begin{table}[h]
\centering
\small
\caption{Per-task Forward Transfer (FWT), Backward Transfer (BWT), and Forgetting for CIFAR-10 using GSOM with a VAE trained from scratch.}
\label{tab:metrics-cifar100-gsom-vae}
\begin{tabular}{c|ccc}
\hline
\textbf{Task} & \textbf{FWT} & \textbf{BWT} & \textbf{Forgetting} \\
\hline
0 & --        & -0.4960 &  0.4960       \\
1 & -0.6810   & -0.1980   &  0.1980   \\
2 & -0.5130   & -0.3520   &  0.3520   \\
3 & -0.6860   & -0.3290   &  0.3290   \\
4 & -0.4900   & -0.0800   &  0.0800   \\
5 & -0.5540   & -0.0130   &  0.0130   \\
6 & -0.2680   &  0.0100   & -0.0090   \\
7 & -0.4010   &  0.0000   &  0.0000   \\
8 & -0.6150   &  0.0090   & -0.0090   \\
9 & -0.8130   &  0.0000   & --        \\
\hline
\end{tabular}
\end{table}

\begin{table}[h]
\centering
\small
\caption{Task accuracy matrix $R$ for CIFAR-100 across 10 sequential tasks for GSOM with a global VAE (latent dimension $32\times2\times2$).}
\label{tab:R-matrix-cifar100-gsom}
\begin{tabular}{c|cccccccccc}
\hline
\textbf{t $\backslash$ j} &
0 & 1 & 2 & 3 & 4 & 5 & 6 & 7 & 8 & 9 \\
\hline
0 & 0.937 & -- & -- & -- & -- & -- & -- & -- & -- & -- \\
1 & 0.781 & 0.574 & -- & -- & -- & -- & -- & -- & -- & -- \\
2 & 0.384 & 0.356 & 0.342 & -- & -- & -- & -- & -- & -- & -- \\
3 & 0.342 & 0.351 & 0.329 & 0.318 & -- & -- & -- & -- & -- & -- \\
4 & 0.311 & 0.326 & 0.304 & 0.317 & 0.292 & -- & -- & -- & -- & -- \\
5 & 0.284 & 0.302 & 0.281 & 0.297 & 0.278 & 0.265 & -- & -- & -- & -- \\
6 & 0.263 & 0.281 & 0.267 & 0.279 & 0.264 & 0.252 & 0.241 & -- & -- & -- \\
7 & 0.241 & 0.259 & 0.247 & 0.262 & 0.251 & 0.239 & 0.229 & 0.221 & -- & -- \\
8 & 0.181 & 0.198 & 0.193 & 0.204 & 0.199 & 0.191 & 0.186 & 0.179 & 0.173 & -- \\
9 & 0.106 & 0.112 & 0.101 & 0.198 & 0.102 & 0.134 & 0.137 & 0.136 & 0.095 & 0.101 \\
\hline
\end{tabular}
\end{table}

\begin{table}[h]
\centering
\small
\caption{Per-task Forward Transfer (FWT), Backward Transfer (BWT), and Forgetting for CIFAR-100 using GSOM with a global VAE.}
\label{tab:metrics-cifar100-gsom}
\begin{tabular}{c|ccc}
\hline
\textbf{Task} & \textbf{FWT} & \textbf{BWT} & \textbf{Forgetting} \\
\hline
0 & --       & -0.8310 &  0.8310 \\
1 & -0.1020 & -0.4620 &  0.4620 \\
2 & -0.0890 & -0.2410 &  0.2410 \\
3 & -0.0820 & -0.1200 &  0.1200 \\
4 & -0.0790 & -0.1900 &  0.1900 \\
5 & -0.0750 & -0.1310 &  0.1310 \\
6 & -0.0730 & -0.1040 &  0.1040 \\
7 & -0.0710 & -0.0850 &  0.0850 \\
8 & -0.0690 & -0.0780 &  0.0780 \\
9 & -0.0670 &  0.0000 & --      \\
\hline
\end{tabular}
\end{table}

\section{Encoder Decoder architectures}
\label{app:encdec_arch}
For experiments involving pretrained encoders, we adopt a unified decoding strategy to reconstruct images from latent representations. Specifically, the ResNet-18 encoder (Table~\ref{tab:resnet_decoder}) produces a 512-dimensional feature vector, which is projected to a low-resolution spatial grid and progressively upsampled through a series of residual and transposed convolution blocks to recover the image. In parallel, CLIP-based configurations (Table~\ref{tab:clip_decoder}) follow a similar decoding pipeline, but incorporate higher-capacity channels and attention modules to better capture the structure of semantic embeddings. By default, pretrained encoders are kept frozen to preserve their learned representations, with optional fine-tuning of the final layers (e.g., the last residual block in ResNet-18 or the final transformer block in CLIP) to improve adaptation in the continual learning setting. For models trained from scratch, we employ the residual VAE architecture (Table~\ref{tab:vae_architecture}), which encodes images into compact latent representations and reconstructs them symmetrically using ResUp blocks. Across all configurations, decoded samples are used for generative replay, where latent representations are mapped onto GSOM grids of varying sizes depending on dataset complexity.

\begin{table*}[!t]
\centering
\caption{Architecture of the VAE model used for CIFAR-10/100 (input size: $3 \times 32 \times 32$). ResDown and ResUp denote residual blocks with downsampling and upsampling, respectively. All activations are ELU except for the output layer, which uses Tanh.}

\label{tab:vae_architecture}
\small
\begin{tabular}{lllllll}
\toprule
\textbf{Layer} & \textbf{Kernel / Type} & \textbf{Stride / Scale} & \textbf{Activation} & \textbf{Skip} & \textbf{Output Shape} & \textbf{Params} \\
\midrule
\multicolumn{7}{c}{\textit{Encoder}} \\
\midrule
Conv2d (Input)   & $3 \times 3$ / Conv     & 1           & –     & No  & $32 \times 32 \times 32$  & 896 \\
ResDown Block 1  & $3 \times 3$ / ×2       & 2 + 1       & ELU   & Yes & $64 \times 16 \times 16$  & 46.3K \\
ResDown Block 2  & $3 \times 3$ / ×2       & 2 + 1       & ELU   & Yes & $128 \times 8 \times 8$   & 184.6K \\
ResDown Block 3  & $3 \times 3$ / ×2       & 2 + 1       & ELU   & Yes & $256 \times 4 \times 4$   & 737.9K \\
ResDown Block 4  & $3 \times 3$ / ×2       & 2 + 1       & ELU   & Yes & $512 \times 2 \times 2$   & 2.36M \\
ResBlock         & $3 \times 3$ / ×2       & 1 + 1       & ELU   & Yes & $512 \times 2 \times 2$   & 2.36M \\
Conv2d ($\mu$)   & $1 \times 1$ / Conv     & 1           & –     & No  & $32 \times 2 \times 2$    & 16.4K \\
Conv2d (log$\sigma^2$)   & $1 \times 1$ / Conv     & 1           & –     & No  & $32 \times 2 \times 2$    & 16.4K \\
\midrule
\multicolumn{7}{c}{\textit{Decoder}} \\
\midrule
Conv2d (Latent)  & $1 \times 1$ / Conv     & 1           & ELU   & No  & $512 \times 2 \times 2$   & 16.9K \\
ResBlock         & $3 \times 3$ / ×2       & 1 + 1       & ELU   & Yes & $512 \times 2 \times 2$   & 2.36M \\
ResUp Block 1    & $3 \times 3$ / ×2       & Upsample×2  & ELU   & Yes & $256 \times 4 \times 4$   & 627.4K \\
ResUp Block 2    & $3 \times 3$ / ×2       & Upsample×2  & ELU   & Yes & $128 \times 8 \times 8$   & 295.2K \\
ResUp Block 3    & $3 \times 3$ / ×2       & Upsample×2  & ELU   & Yes & $64 \times 16 \times 16$  & 184.5K \\
ResUp Block 4    & $3 \times 3$ / ×2       & Upsample×2  & ELU   & Yes & $32 \times 32 \times 32$  & 82.9K \\
Conv2d (Output)  & $5 \times 5$ / Conv     & 1           & Tanh  & No  & $3 \times 32 \times 32$   & 2.4K \\
\midrule
\textbf{Total Parameters} & & & & & & \textbf{9.3M} \\
\bottomrule
\end{tabular}
\end{table*}

\begin{table}[t]
\centering
\small
\caption{Decoder architecture used for ResNet-18 + GSOM experiments. The encoder is an ImageNet-pretrained ResNet-18 that outputs a 512-dimensional feature vector. The decoder maps this embedding back to the dataset-native RGB image resolution ($32\times32$ for CIFAR-10/100, $64\times64$ for TinyImageNet, $84\times84$ for MiniImageNet). GN denotes GroupNorm.}
\label{tab:resnet_decoder}
\begin{tabular}{@{}p{2.3cm}p{4.4cm}ccc@{}}
\toprule
\textbf{Layer} & \textbf{Operation / Details} & \textbf{Activation} & \textbf{Norm} & \textbf{Output Shape} \\
\midrule

Linear projection 
& FC$(512 \rightarrow 512 \cdot 4 \cdot 4)$, reshape 
& SiLU & GN & $B \times 512 \times 4 \times 4$ \\

ResBlock 
& Conv $3{\times}3$ + Conv $3{\times}3$ 
& SiLU & GN & $B \times 512 \times 4 \times 4$ \\

Upsample Block 1 
& ConvTranspose2d $4{\times}4$, stride 2 
& SiLU & GN & $B \times 256 \times 8 \times 8$ \\

ResBlock 
& Conv $3{\times}3$ + Conv $3{\times}3$ 
& SiLU & GN & $B \times 256 \times 8 \times 8$ \\

Upsample Block 2 
& ConvTranspose2d $4{\times}4$, stride 2 
& SiLU & GN & $B \times 128 \times 16 \times 16$ \\

ResBlock 
& Conv $3{\times}3$ + Conv $3{\times}3$ 
& SiLU & GN & $B \times 128 \times 16 \times 16$ \\

Upsample Block 3 
& ConvTranspose2d $4{\times}4$, stride 2 
& SiLU & GN & $B \times 64 \times 32 \times 32$ \\

ResBlock 
& Conv $3{\times}3$ + Conv $3{\times}3$ 
& SiLU & GN & $B \times 64 \times 32 \times 32$ \\

Optional Upsample Block 
& ConvTranspose2d $4{\times}4$, stride 2; used only for TinyImageNet/MiniImageNet 
& SiLU & GN & $B \times 64 \times 64 \times 64$ \\

Output Head 
& Conv $3{\times}3 \rightarrow 3$ channels 
& Tanh & -- & $B \times 3 \times H \times W$ \\
Resize (MiniImageNet only) & Bilinear interpolation $64\!\times\!64 \to 84\!\times\!84$ & -- & -- & $B \times 3 \times 84 \times 84$ \\

\bottomrule
\end{tabular}
\end{table}

\begin{table}[t]
\centering
\small
\caption{Decoder Architecture used for CLIP+SOM experiments. The encoder is the CLIP ViT-B/32 visual tower (pretrained) and is \emph{frozen} by default; when \texttt{finetune\_last} is enabled, only the last transformer block of ViT-B/32 is unfrozen. The decoder upsamples a single CLIP embedding to the target RGB image in $[-1,1]$. GN denotes GroupNorm(32).}
\label{tab:clip_decoder}
\begin{tabular}{@{}p{2cm}p{3.6cm}cccc@{}}
\toprule
\textbf{Layer} & \textbf{Operation / Details} & \textbf{Blocks} & \textbf{Activation} & \textbf{GN} & \textbf{Output Shape} \\
\midrule
Linear (embed$\rightarrow$4$\times$4) & FC($D\!\rightarrow\!$base\_ch$\cdot4\cdot4$) & -- & --   & Yes & $B\times$1024$\times4\times4$ \\
ResBlock                               & Conv $3{\times}3$ (GN+SiLU)                 & $1{+}1$ & SiLU & Yes & $B\times$1024$\times4\times4$ \\
Upsample                               & ConvTranspose2d $4{\times}4$                 & $\times2$ & -- & Yes & $B\times$1024$\times8\times8$ \\
ResBlock (ch$\downarrow$)              & Conv $3{\times}3$ (GN+SiLU)                 & $1{+}1$ & SiLU & Yes & $B\times$512$\times8\times8$ \\
Upsample                               & ConvTranspose2d $4{\times}4$                 & $\times2$ & -- & Yes & $B\times$512$\times16\times16$ \\
ResBlock (+Attn)                       & Conv $3{\times}3$ + SelfAttn2d               & $1{+}1$ & SiLU & Yes & $B\times$512$\times16\times16$ \\
ResBlock (ch$\downarrow$)              & Conv $3{\times}3$ (GN+SiLU)                 & $1{+}1$ & SiLU & Yes & $B\times$256$\times16\times16$ \\
Upsample                               & ConvTranspose2d $4{\times}4$                 & $\times2$ & -- & Yes & $B\times$256$\times32\times32$ \\
ResBlock (+Attn)                       & Conv $3{\times}3$ + SelfAttn2d               & $1{+}1$ & SiLU & Yes & $B\times$256$\times32\times32$ \\
ResBlock (ch$\downarrow$)              & Conv $3{\times}3$ (GN+SiLU)                 & $1{+}1$ & SiLU & Yes & $B\times$128$\times32\times32$ \\
Upsample                               & ConvTranspose2d $4{\times}4$                 & $\times2$ & -- & Yes & $B\times$128$\times64\times64$ \\
ResBlock                               & Conv $3{\times}3$ (GN+SiLU)                 & $1{+}1$ & SiLU & Yes & $B\times$128$\times\text{target}\times\text{target}$ \\
Output head                            & GN $\rightarrow$ SiLU $\rightarrow$ $3{\times}3$ $\rightarrow$ Tanh & $1$ & Tanh & Yes & $B\times3\times\text{target}\times\text{target}$ \\
Resize (MiniImageNet only) & Bilinear interpolation $64\!\times\!64 \to 84\!\times\!84$ & -- & -- & -- & $B \times 3 \times 84 \times 84$ \\

\midrule
\multicolumn{6}{p{0.95\linewidth}}{\textbf{Notes:} Encoder output dim $D{=}512$ for ViT-B/32. 
SelfAttention2d uses $4$ heads with GN(32). 
By default, only the decoder is trainable; enabling \texttt{finetune\_last} unfreezes the last ViT block. 
Decoder output target size is set to match dataset resolution ($32\times32$ for CIFAR-10/100, $64\times64$ for TinyImageNet, $84\times84$ for MiniImageNet).}\\
\bottomrule
\end{tabular}
\end{table}

\section{Visualization of Learned Representations}
\label{app:som_representations}
\begin{figure*}[!t]
    \centering
    \includegraphics[width=0.32\textwidth]{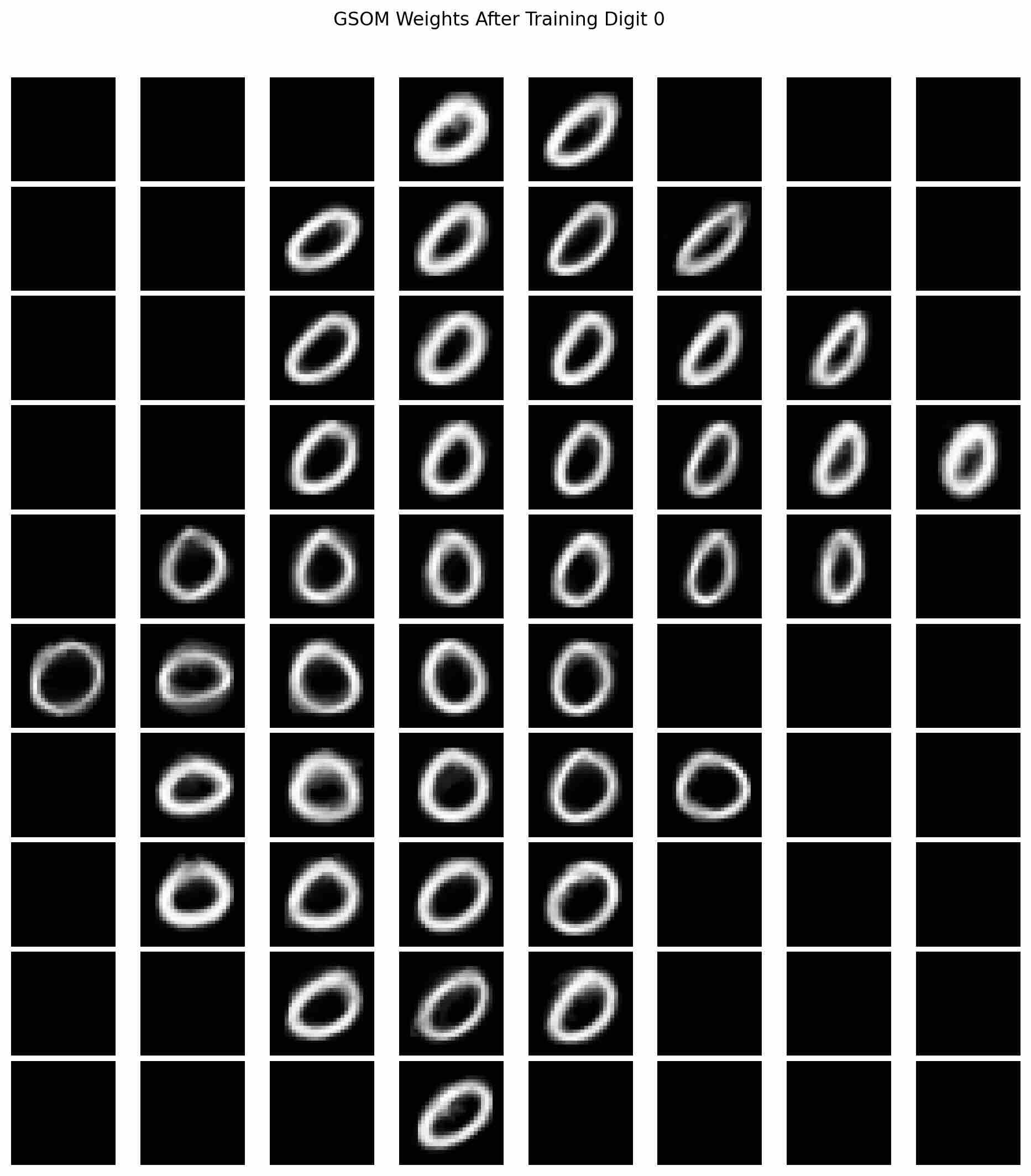}
    \includegraphics[width=0.32\textwidth]{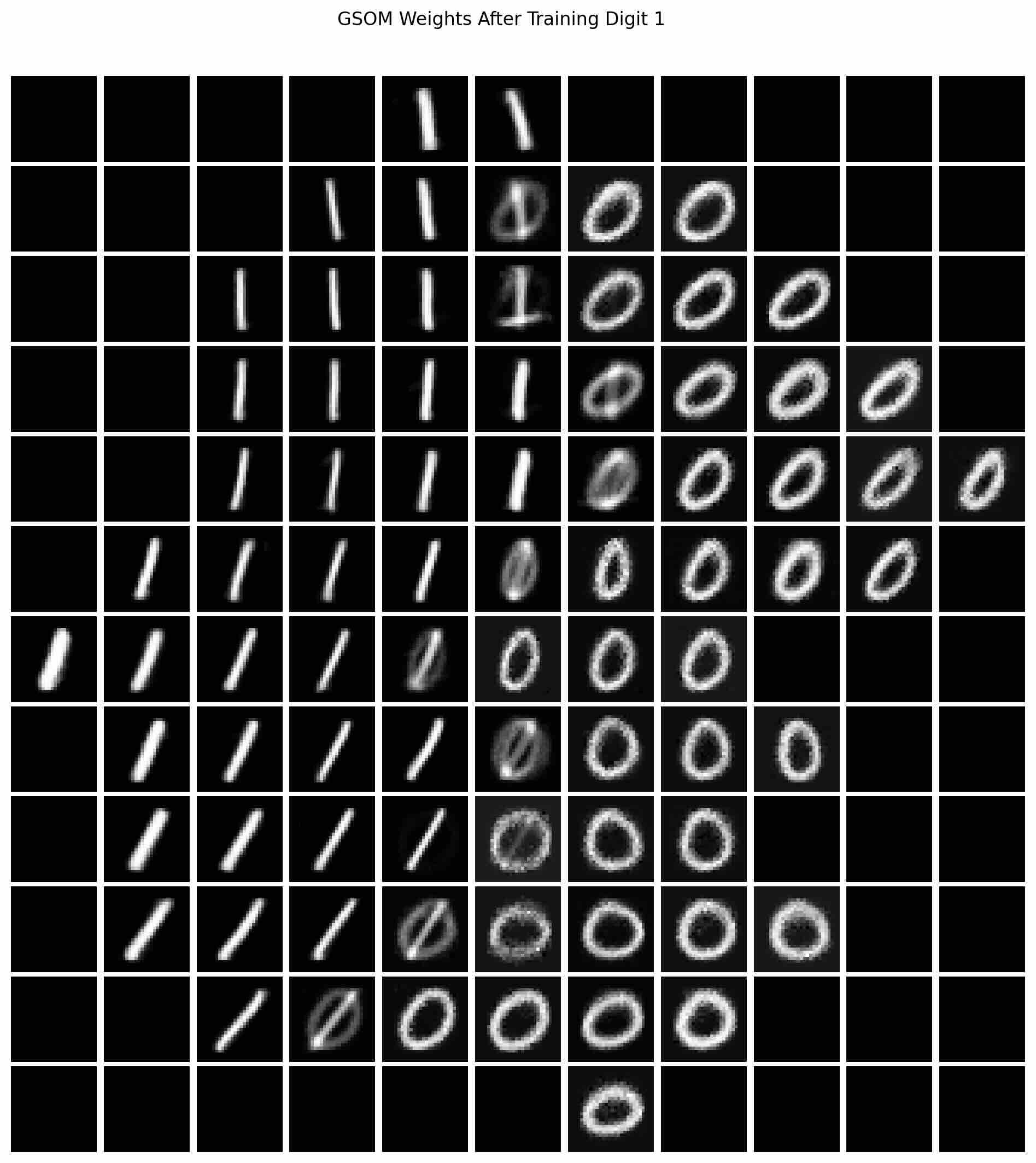}
    \includegraphics[width=0.32\textwidth]{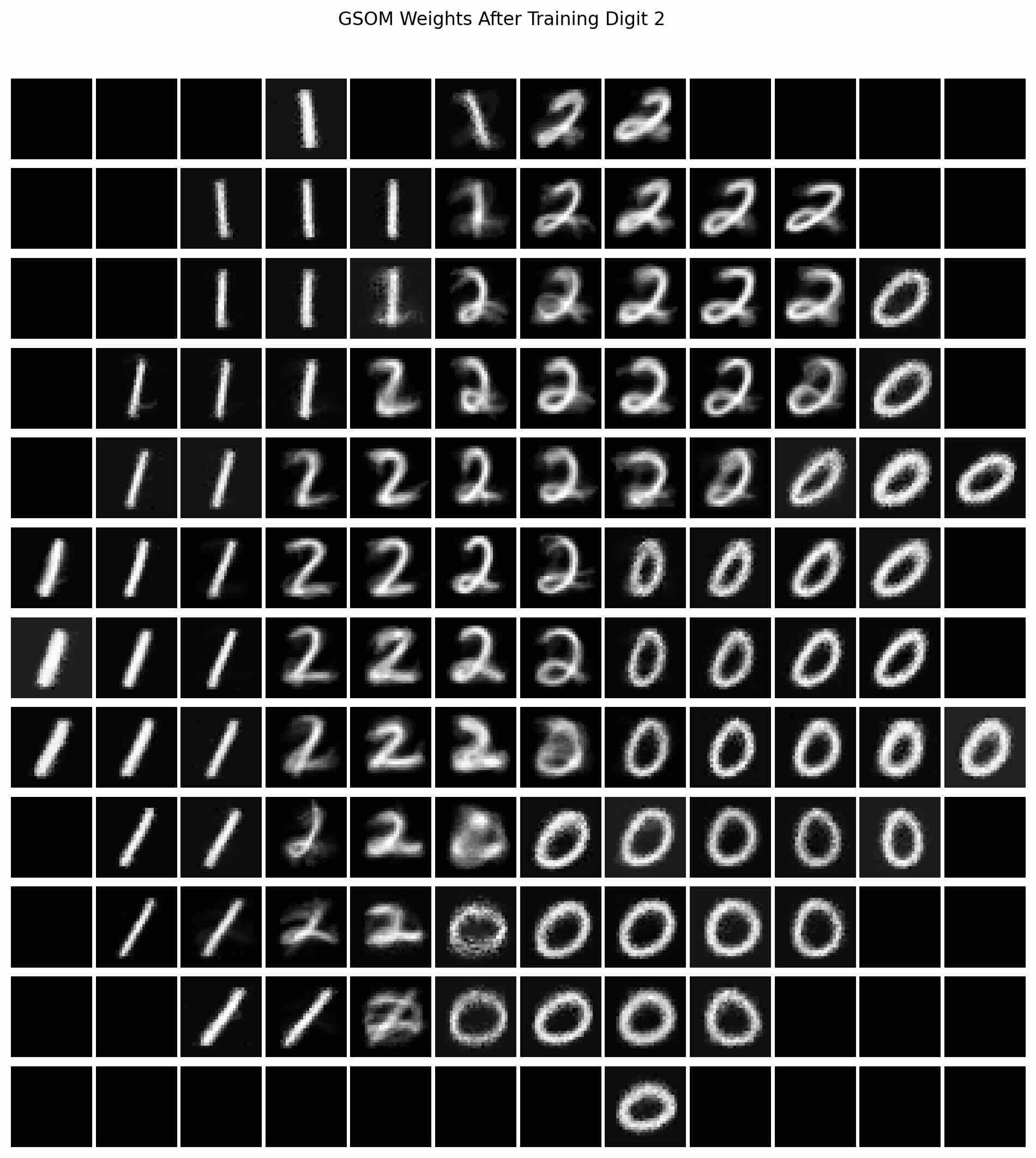} \\[1mm]

    \includegraphics[width=0.32\textwidth]{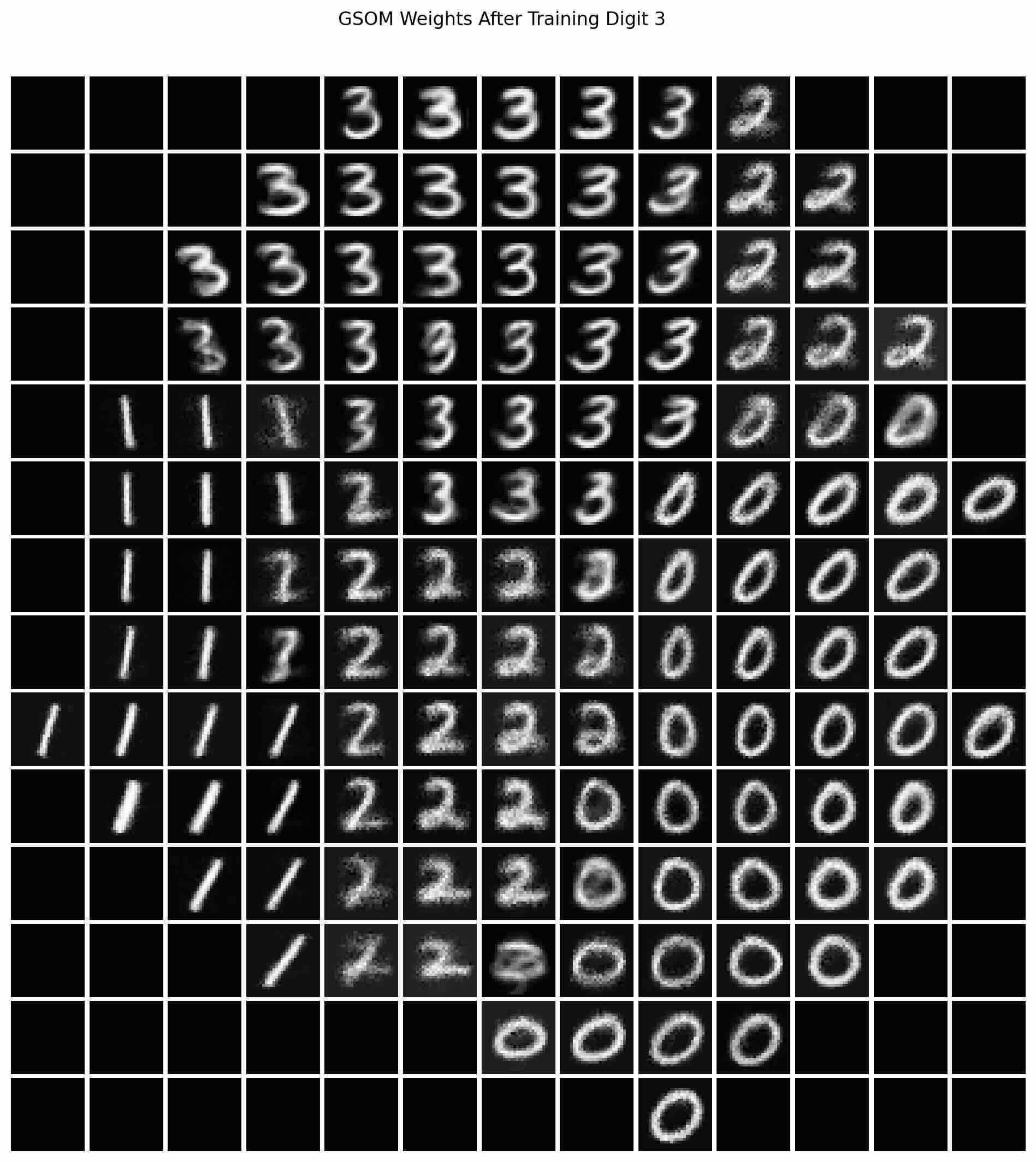}
    \includegraphics[width=0.32\textwidth]{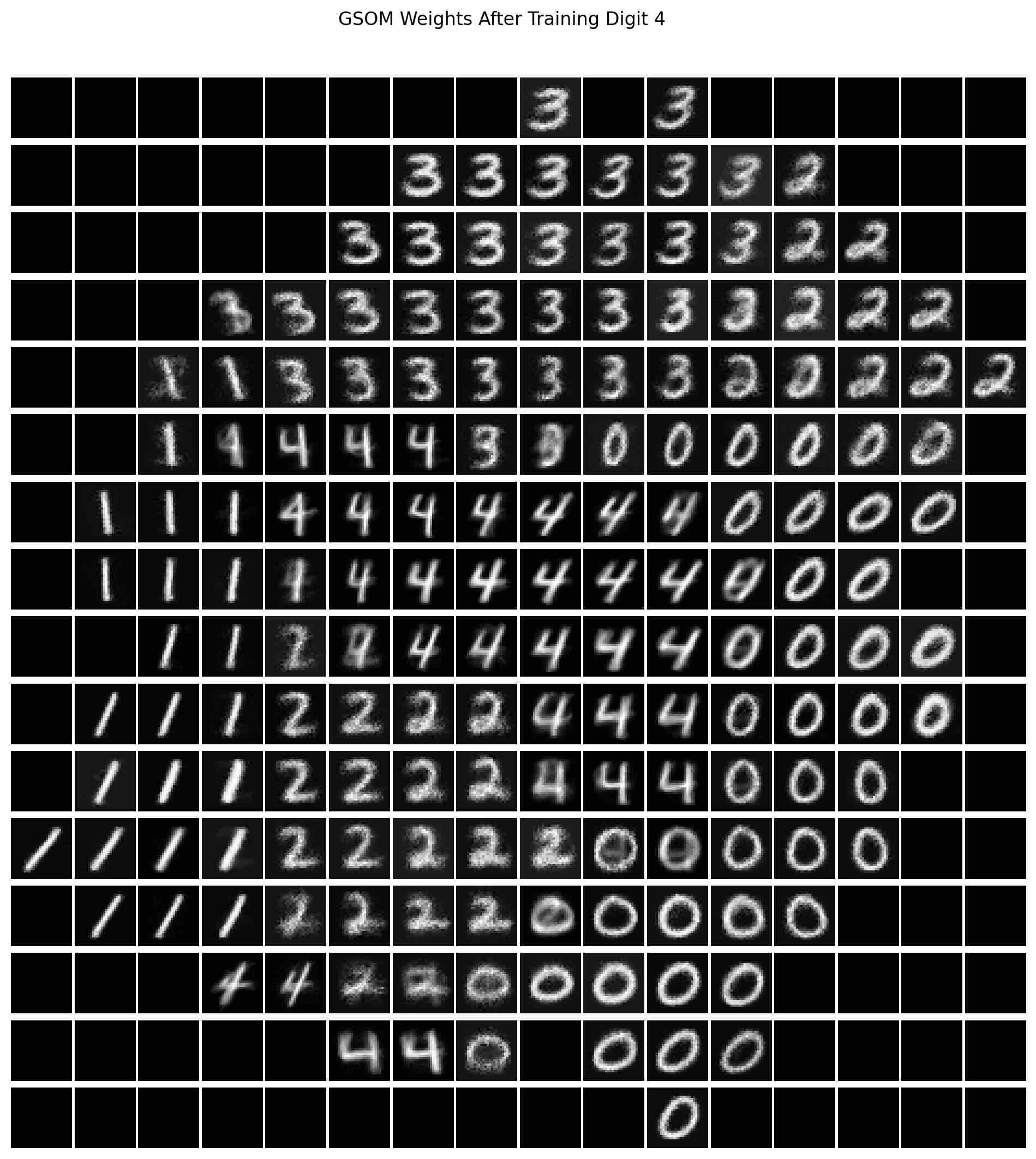}
    \includegraphics[width=0.32\textwidth]{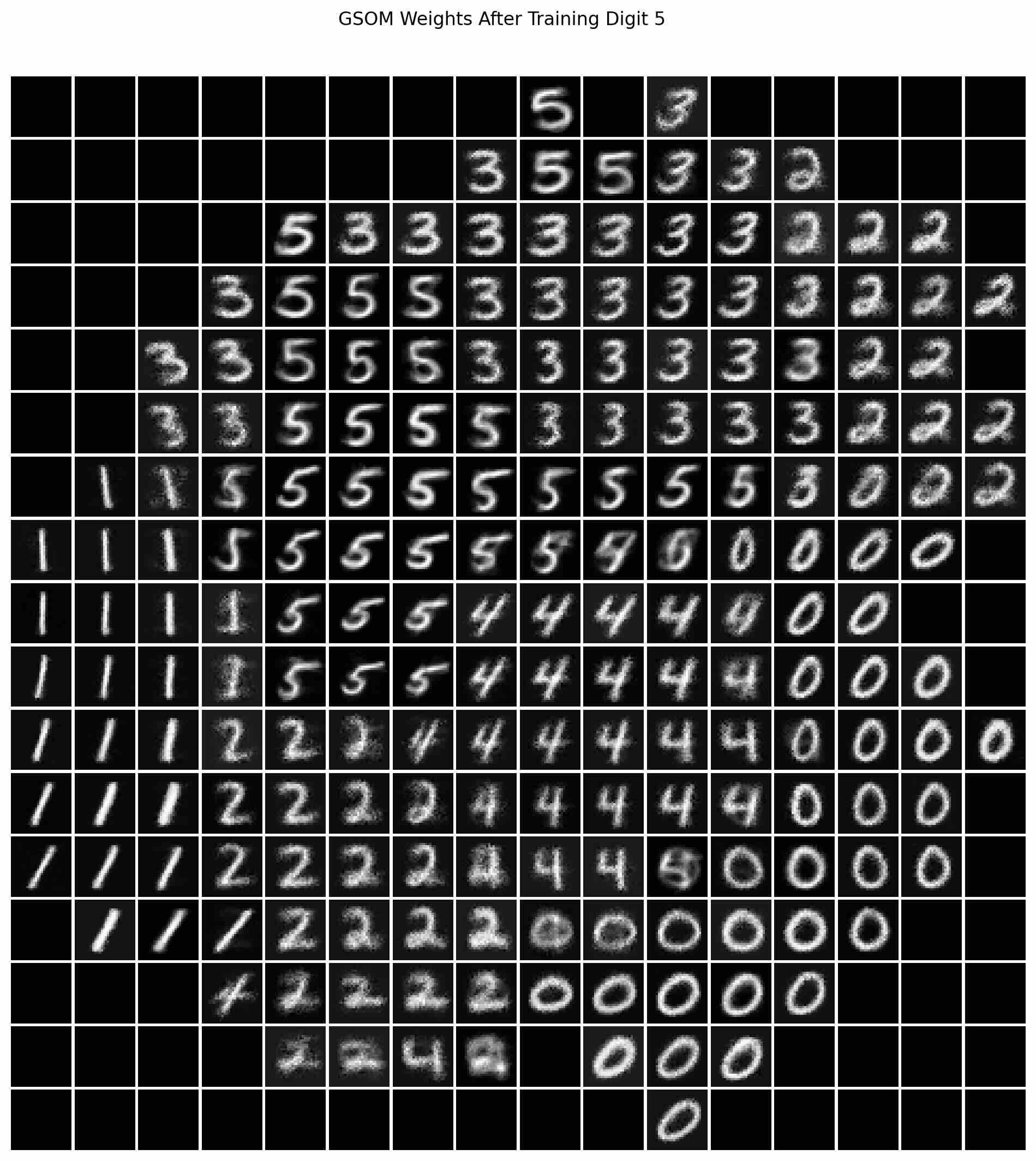} \\[1mm]

    \includegraphics[width=0.32\textwidth]{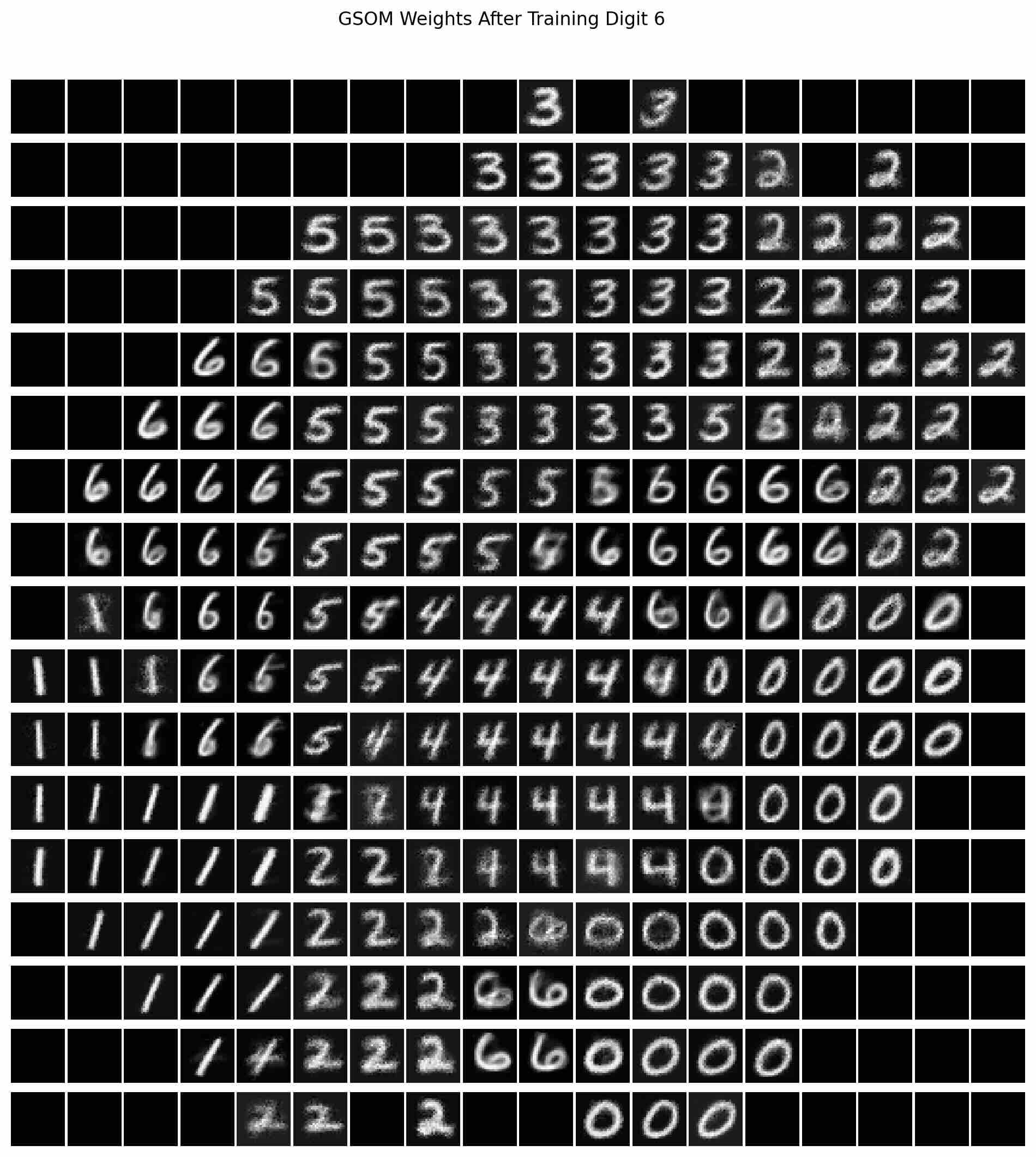}
    \includegraphics[width=0.32\textwidth]{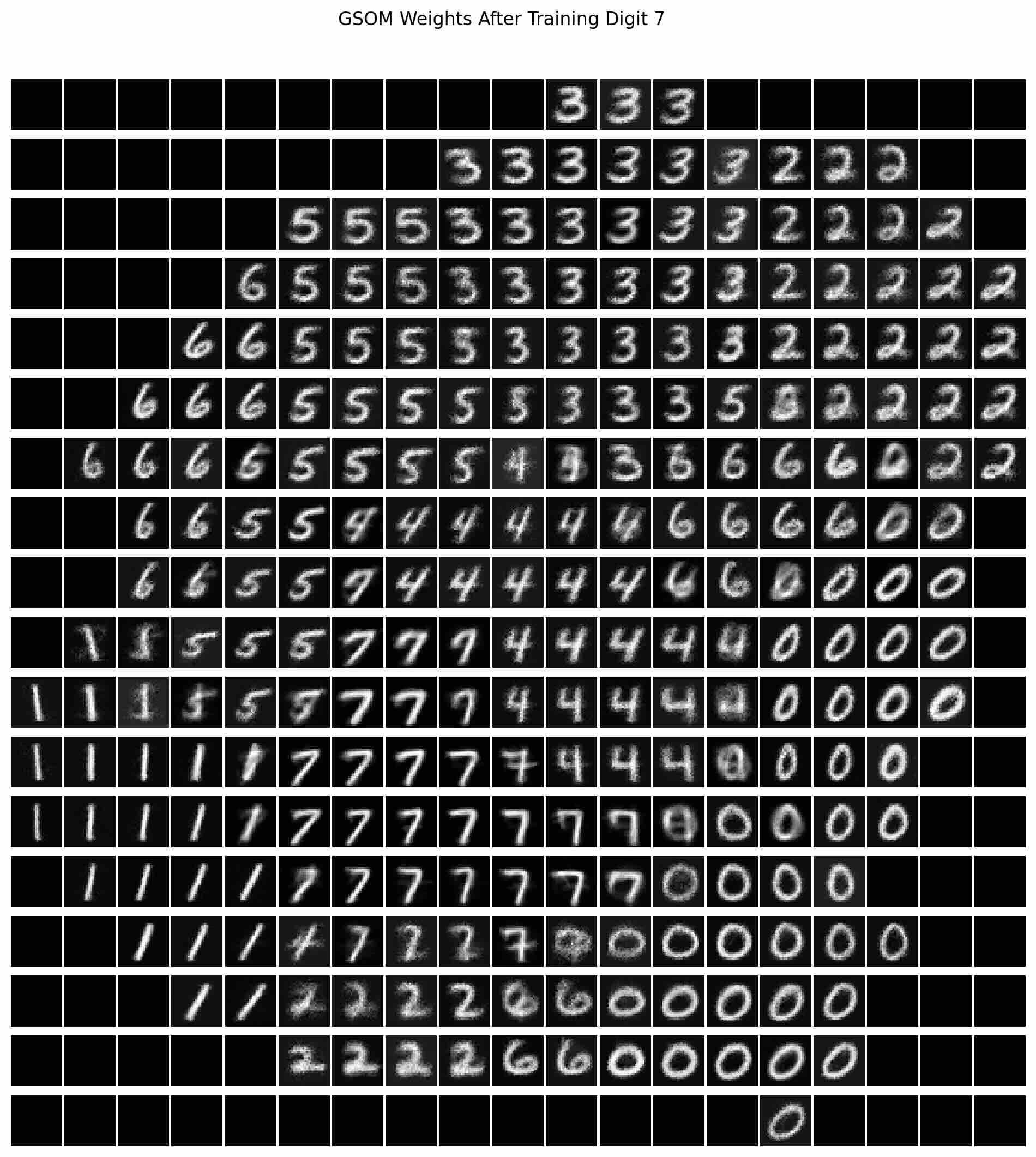}
    \includegraphics[width=0.32\textwidth]{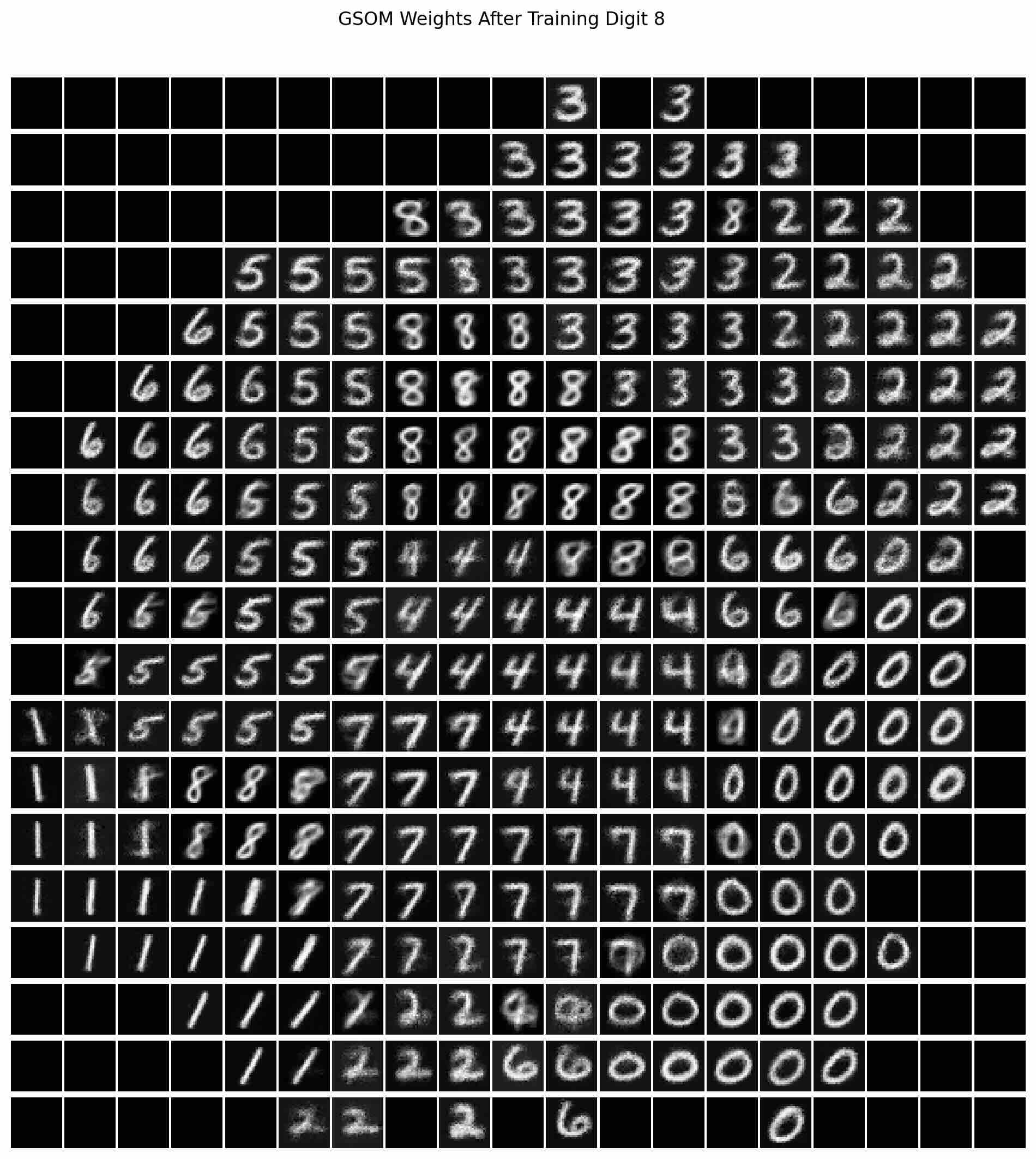} \\[1mm]

    \includegraphics[width=0.32\textwidth]{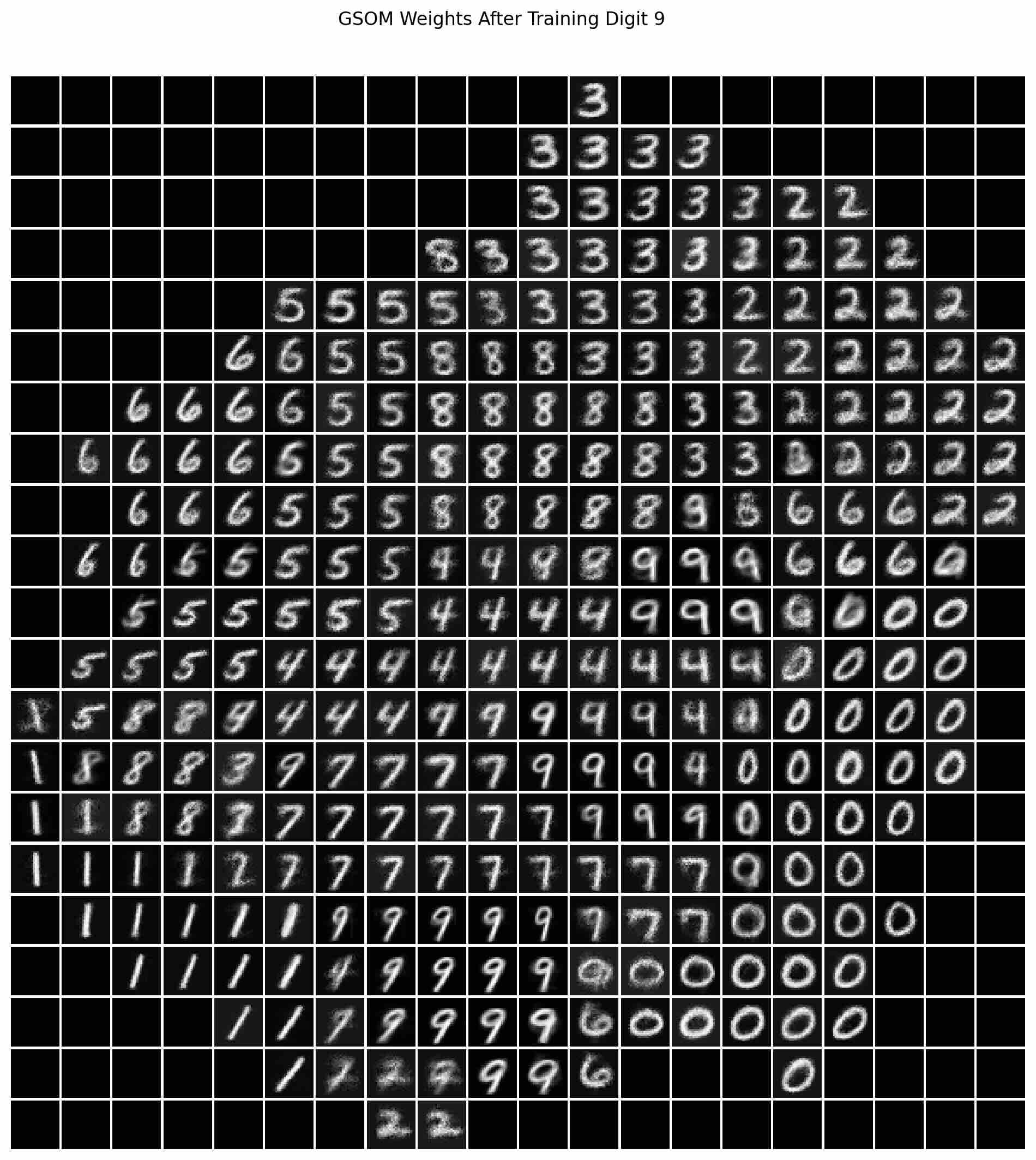}

    \caption{Visualization of the GSOM unit vectors after each task(every single task) for MNIST.}
    \label{fig:appendix_mnist_gen}
\end{figure*}

\begin{figure*}[!t]
    \centering
    \includegraphics[width=0.31\textwidth]{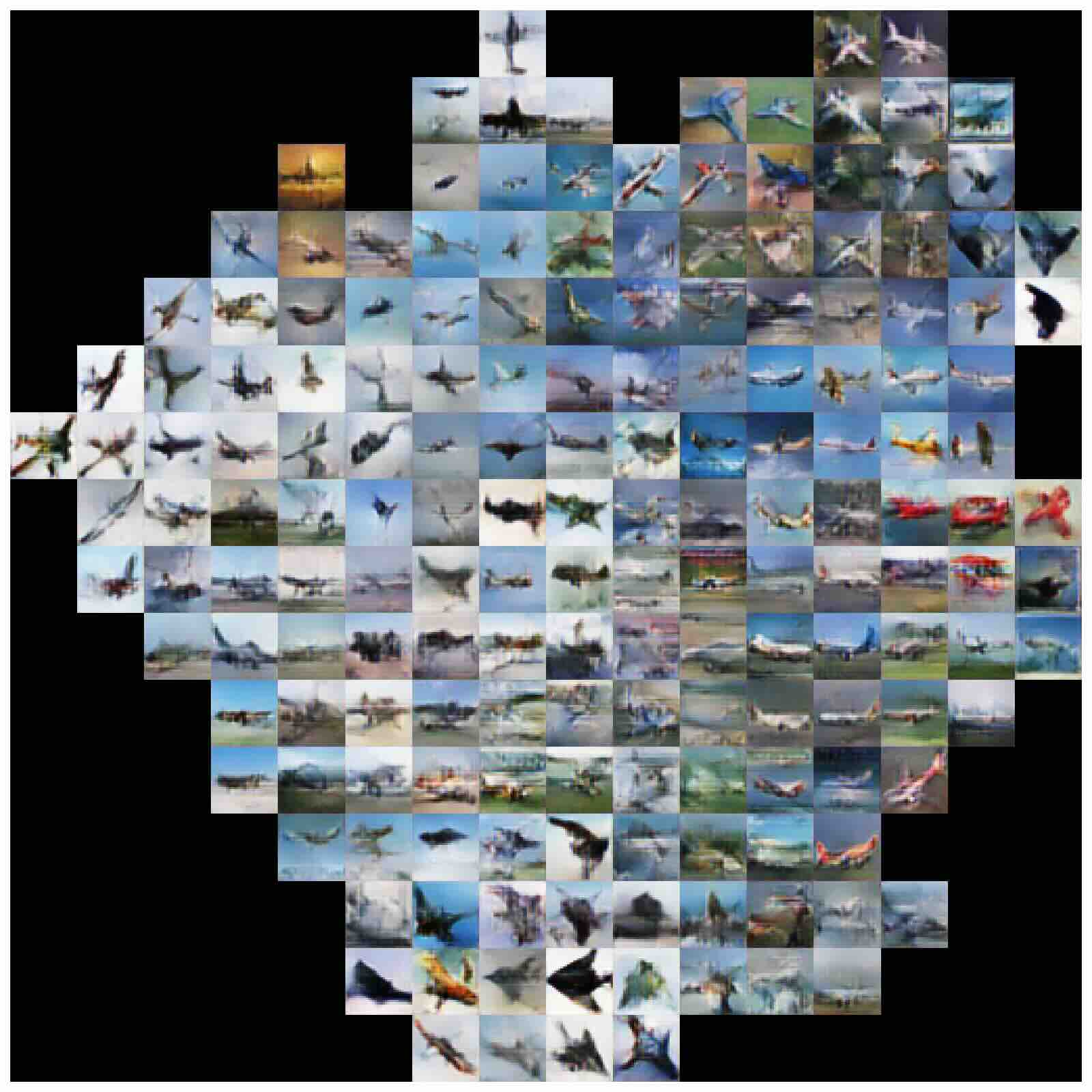}
    \includegraphics[width=0.31\textwidth]{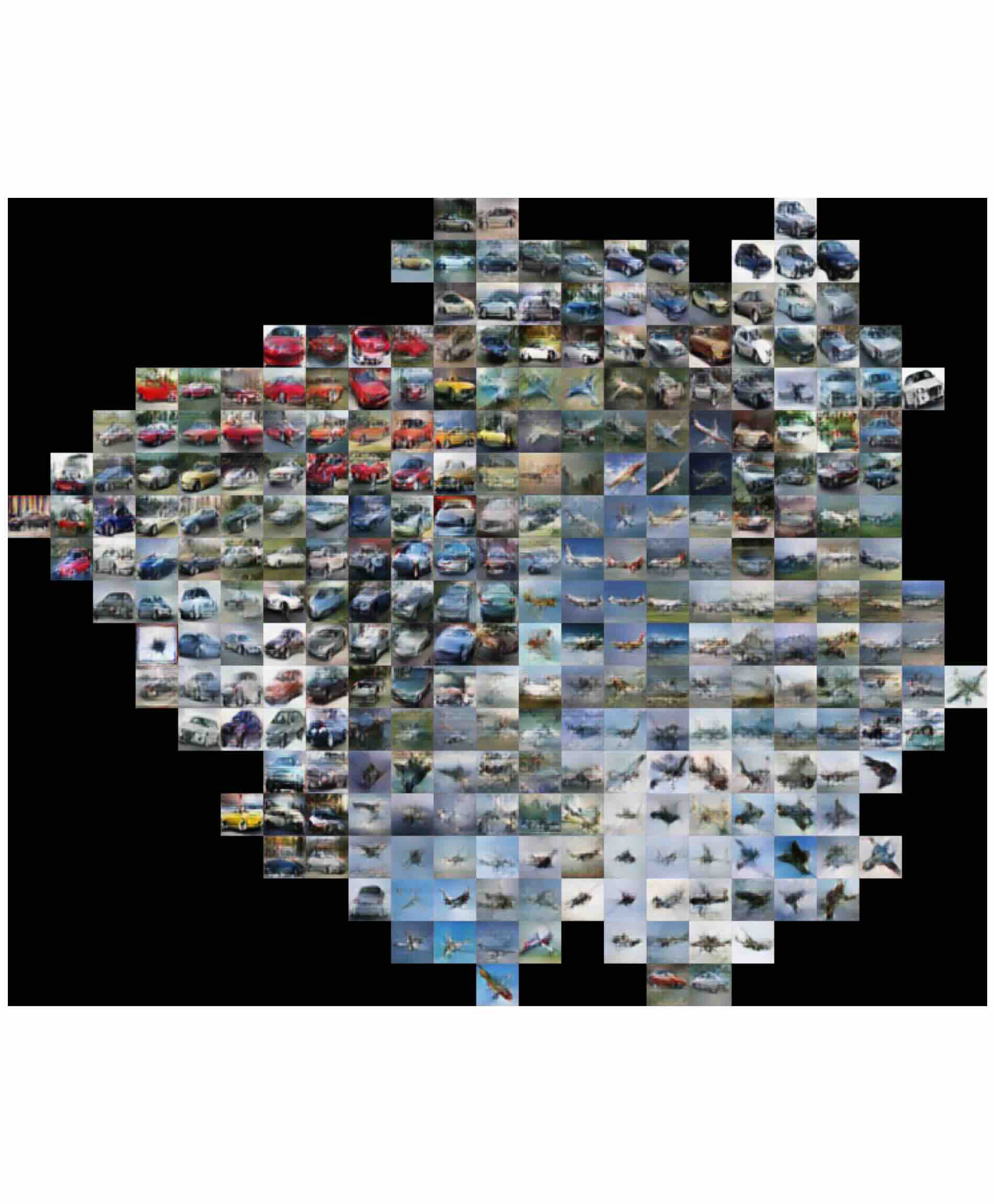}
    \includegraphics[width=0.31\textwidth]{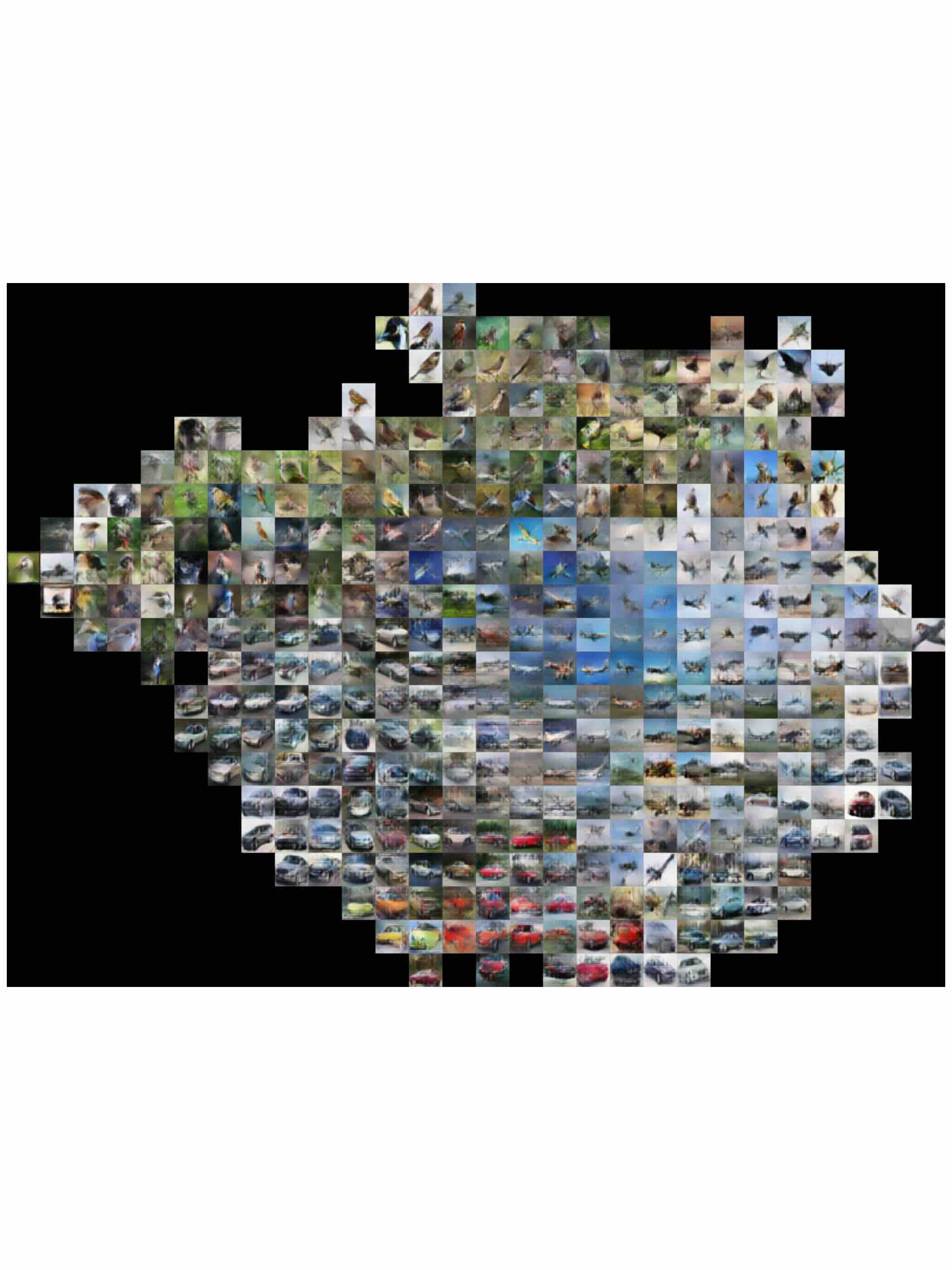} \\[1mm]

    \includegraphics[width=0.32\textwidth]{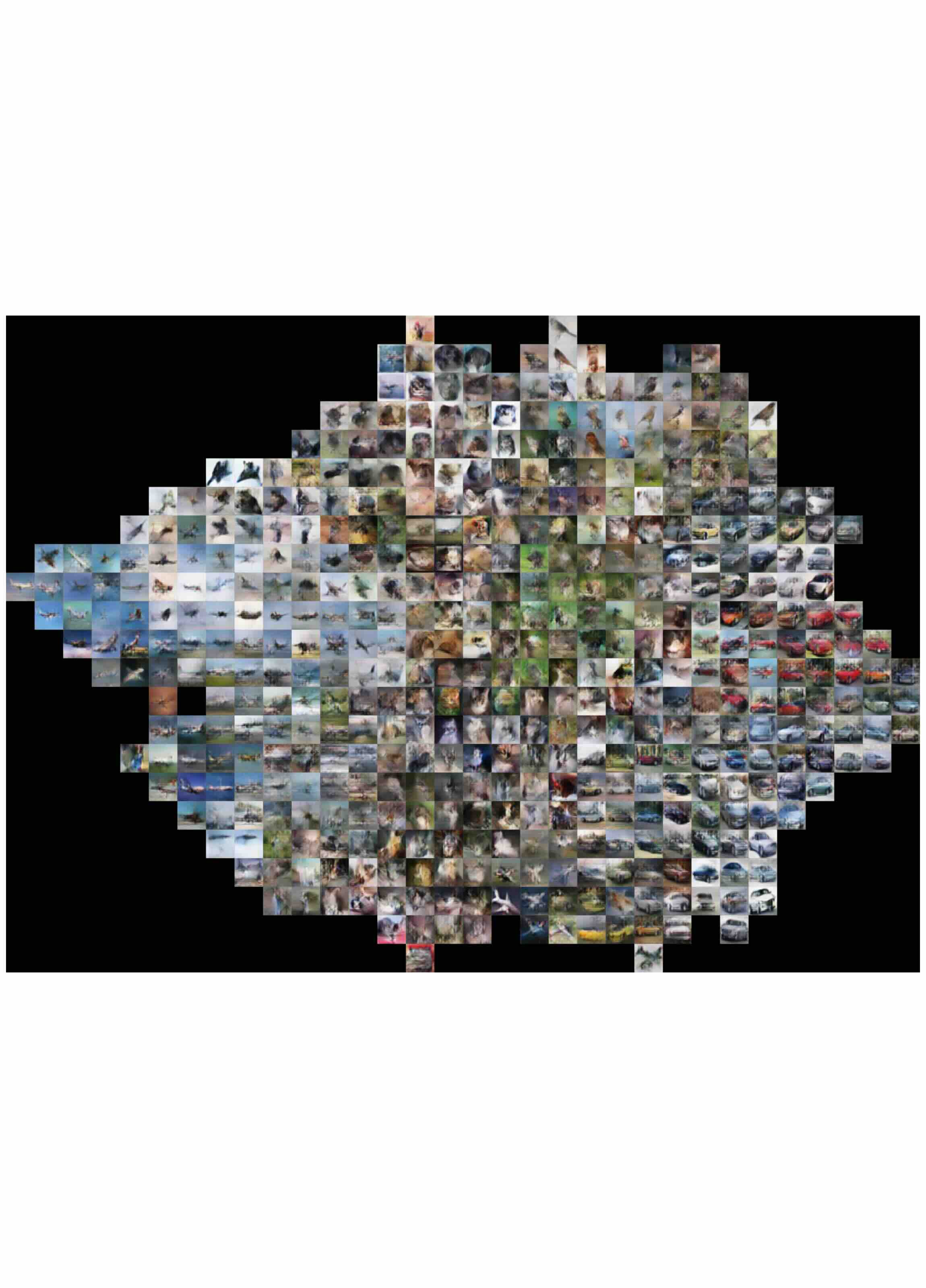}
    \includegraphics[width=0.32\textwidth]{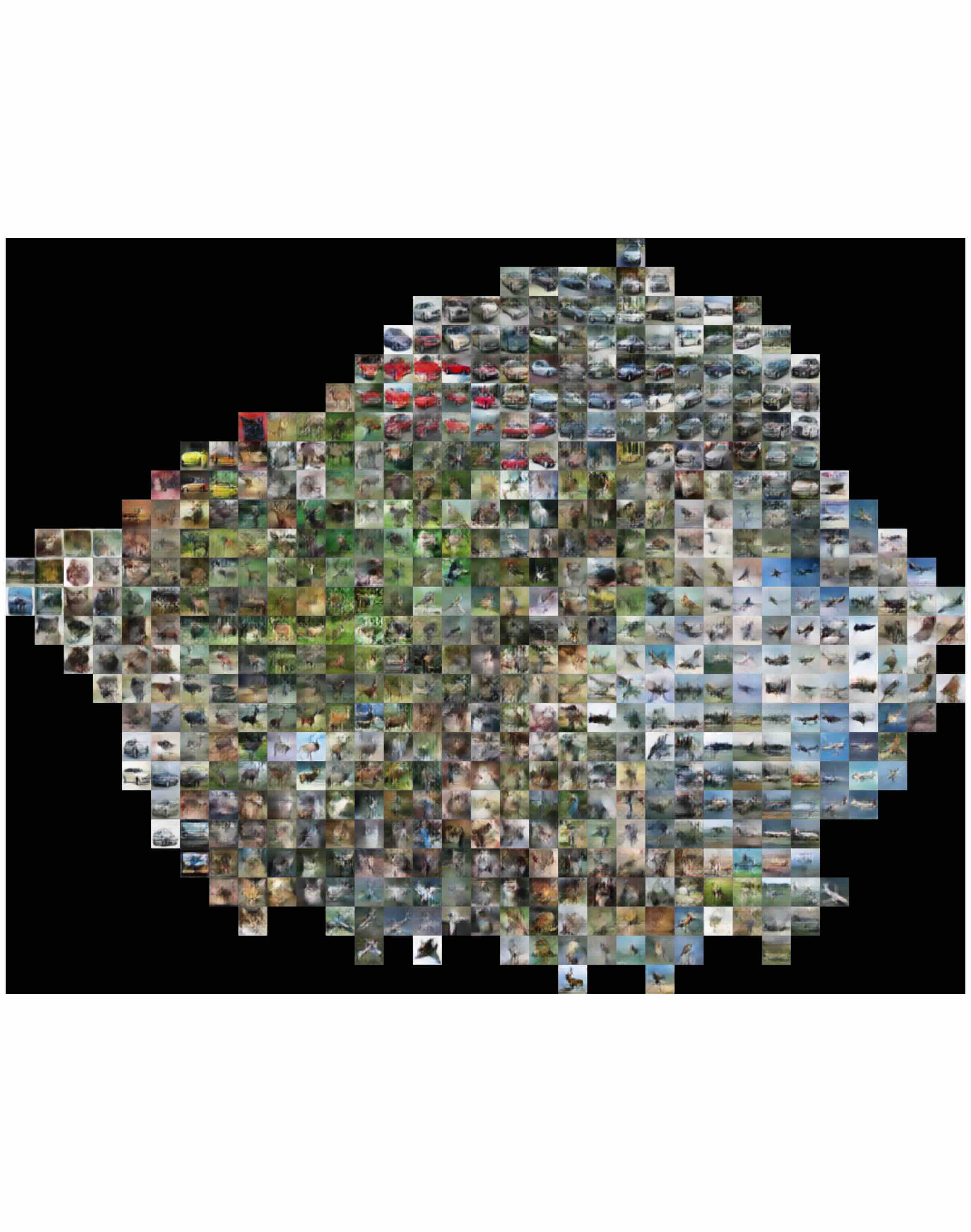}
    \includegraphics[width=0.32\textwidth]{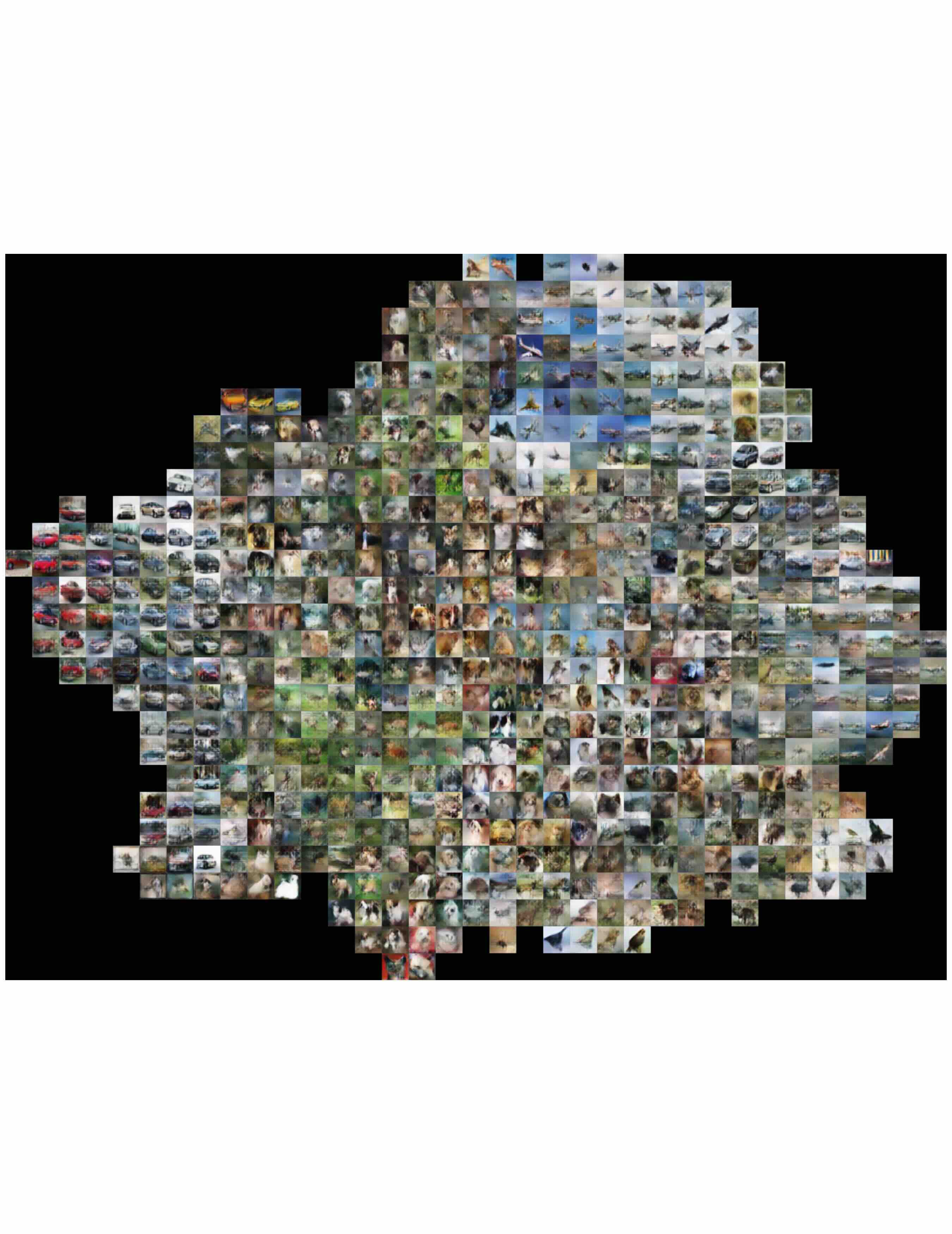} \\[1mm]

    \includegraphics[width=0.32\textwidth]{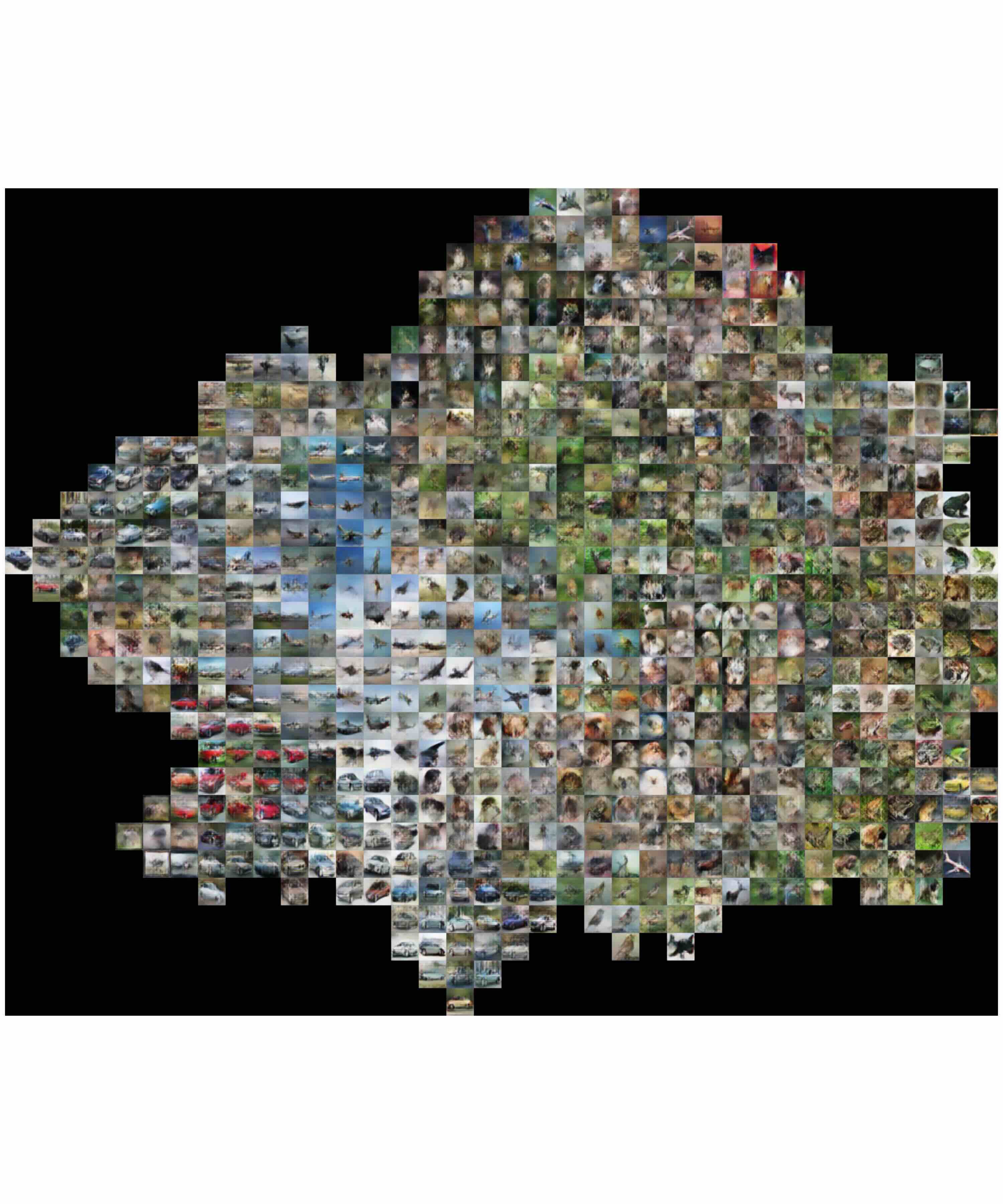}
    \includegraphics[width=0.32\textwidth]{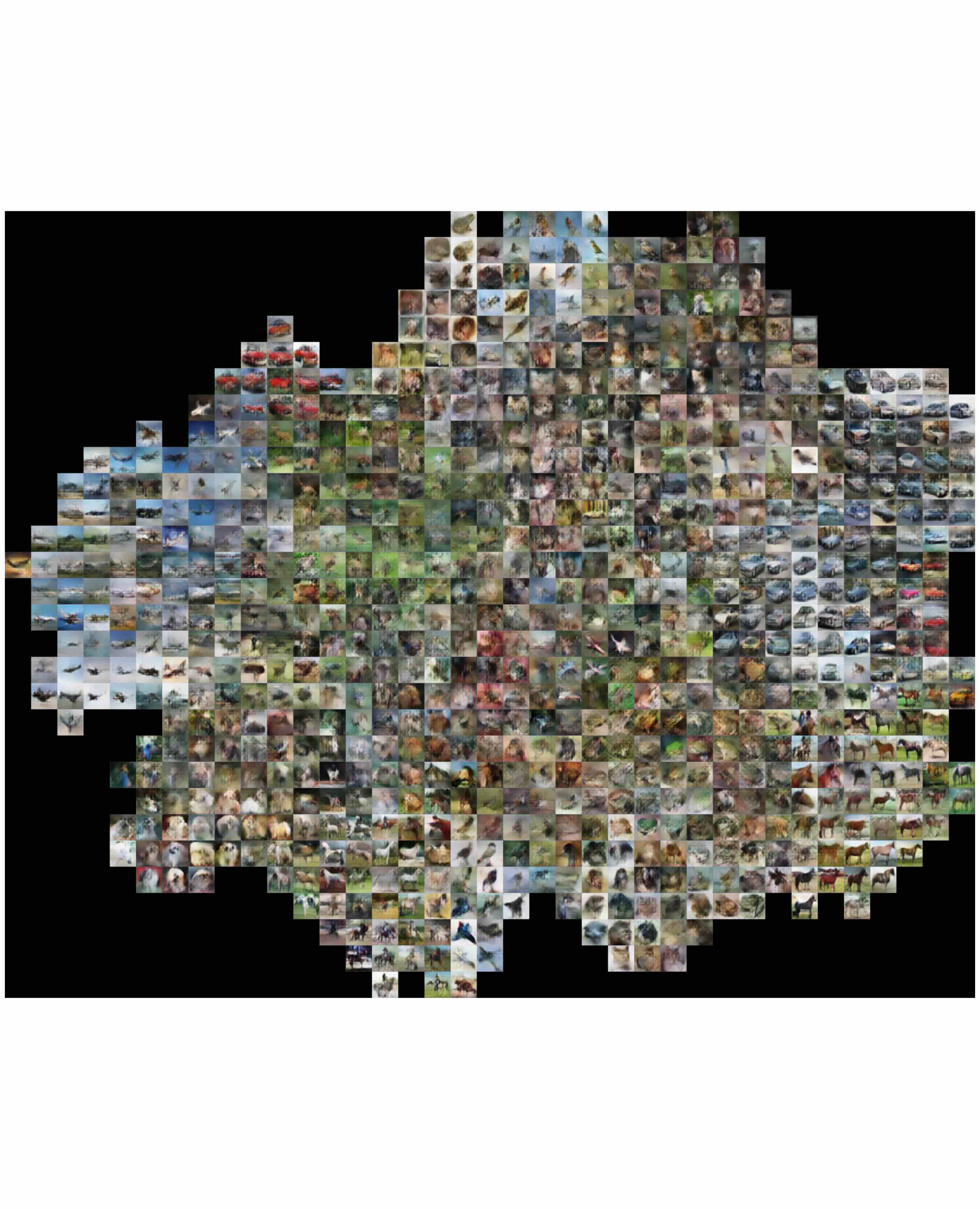}
    \includegraphics[width=0.32\textwidth]{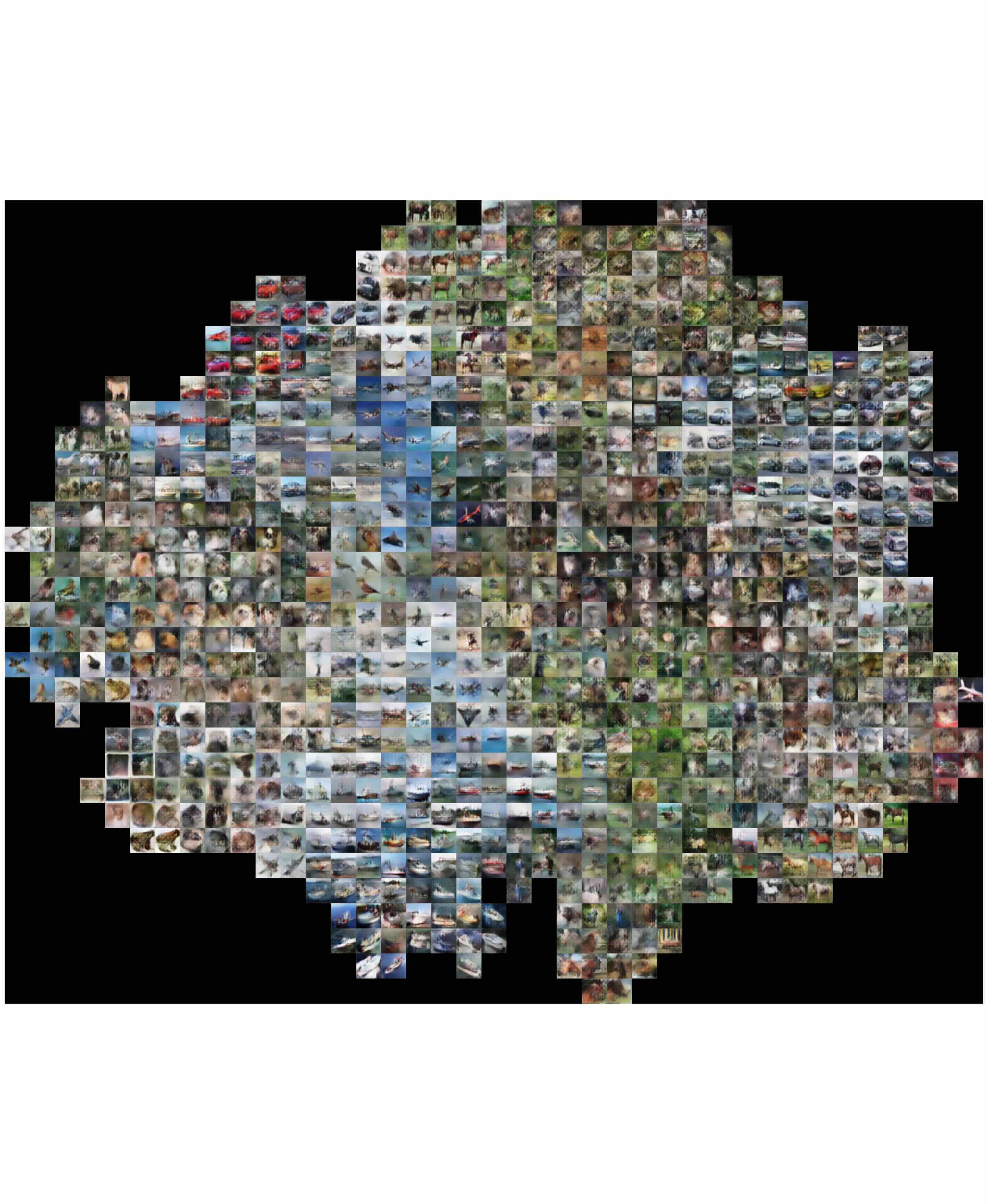} \\[1mm]

    \includegraphics[width=0.33\textwidth]{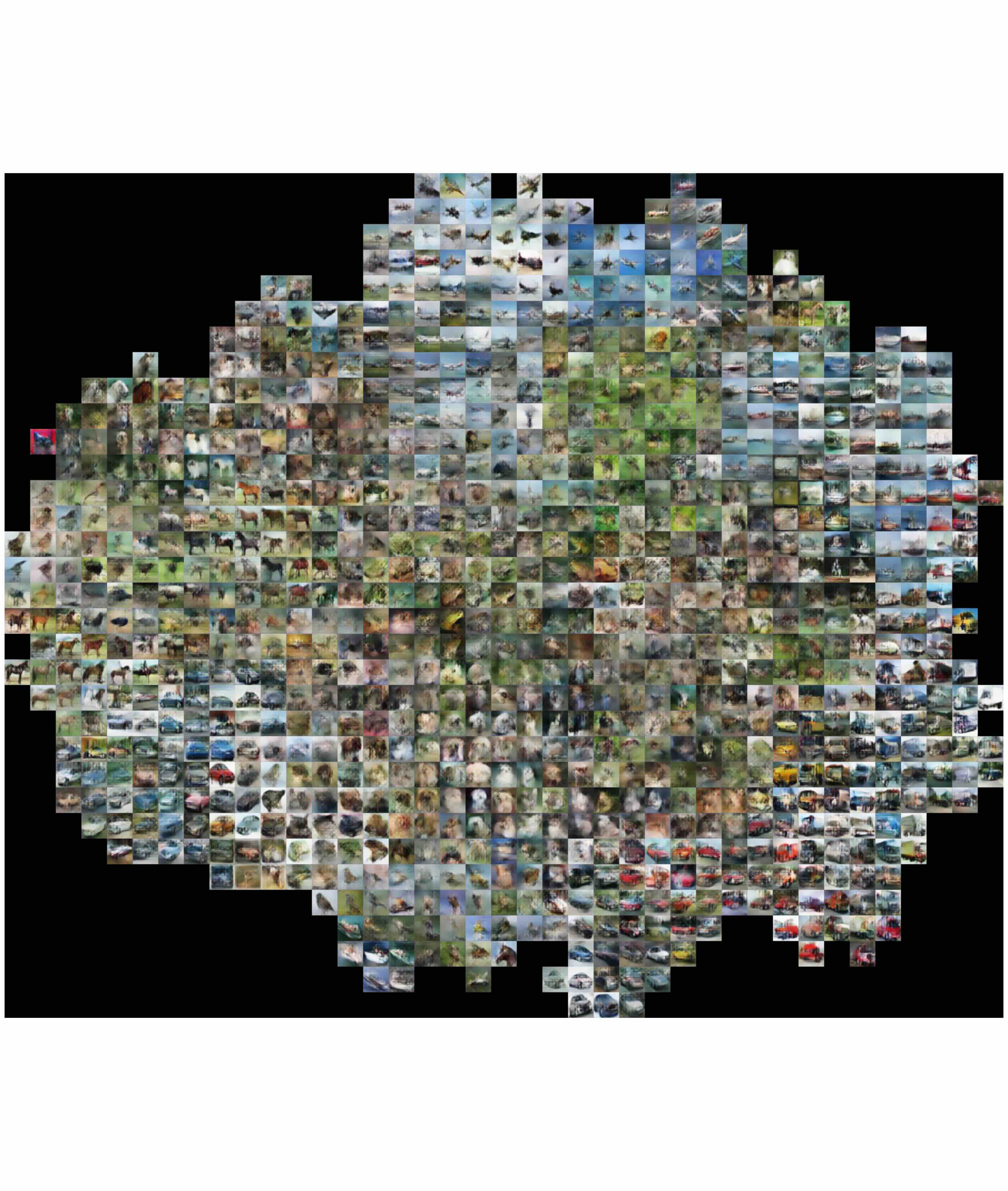}

    \caption{Visualization of decoded GSOM unit vectors after each task(every class) on CIFAR-10.}
    \label{fig:appendix_cifar_10_gen}
\end{figure*}

\begin{figure*}[!t]
    \centering
    \includegraphics[width=0.32\textwidth]{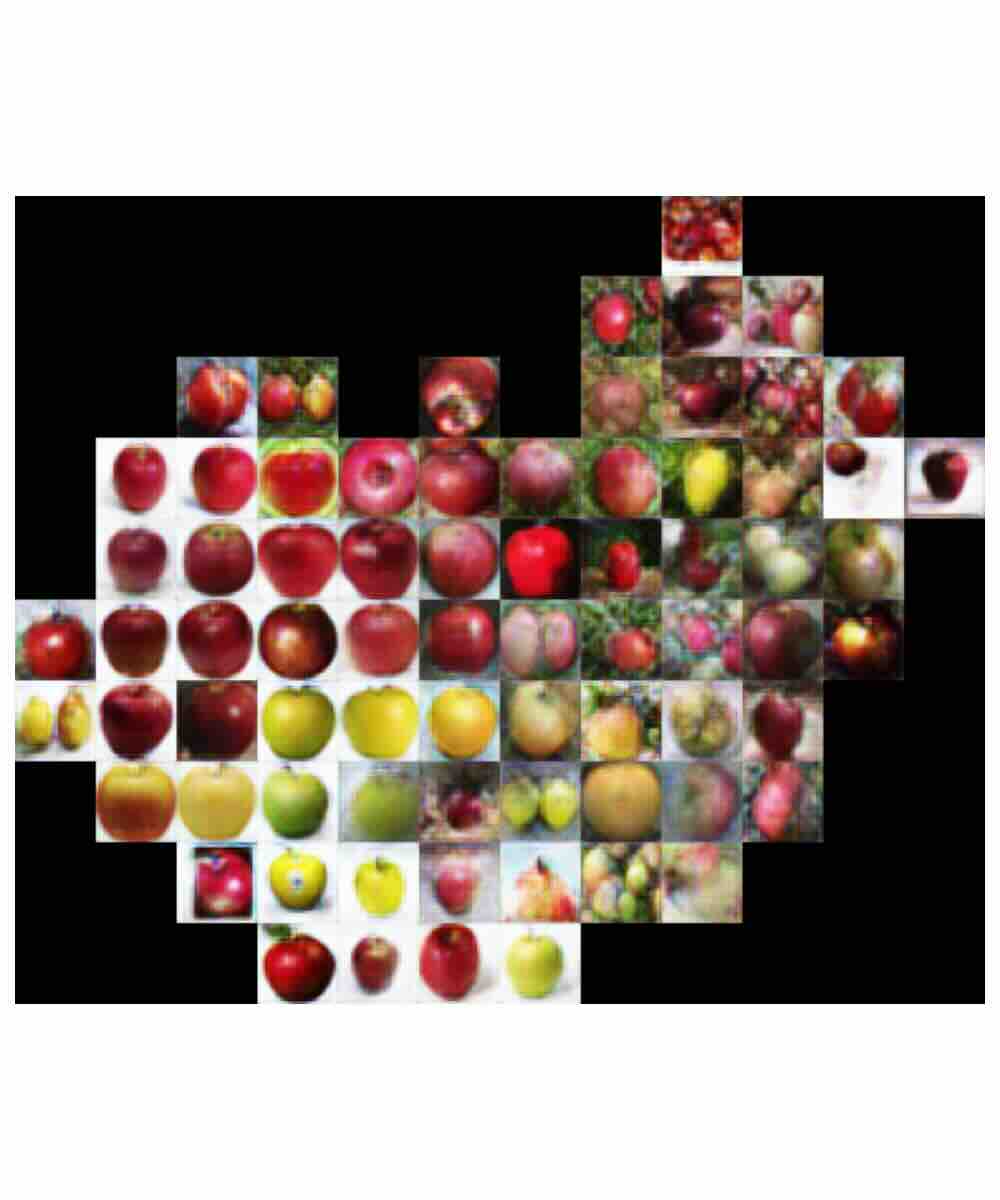}
    \includegraphics[width=0.32\textwidth]{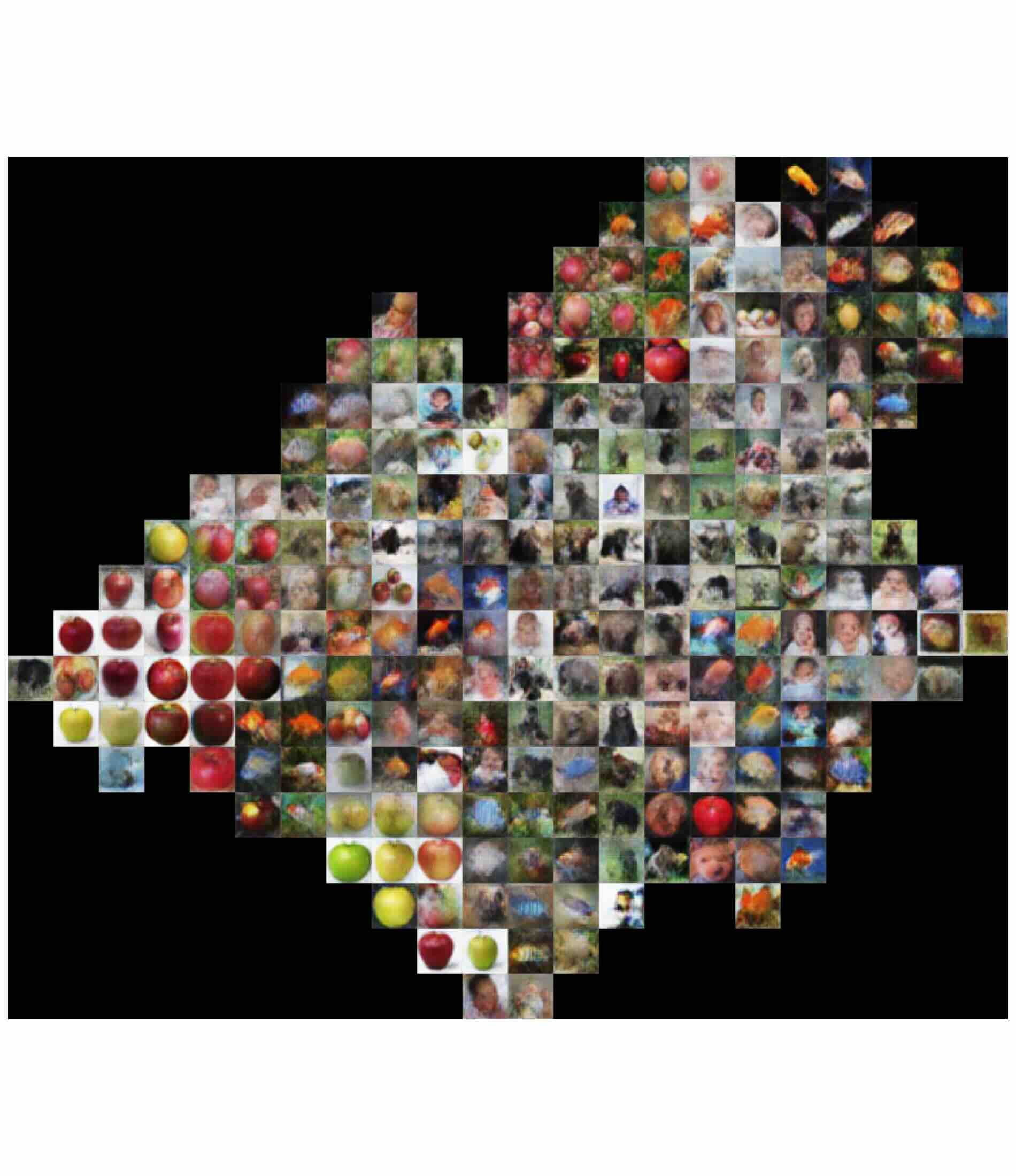}
    \includegraphics[width=0.32\textwidth]{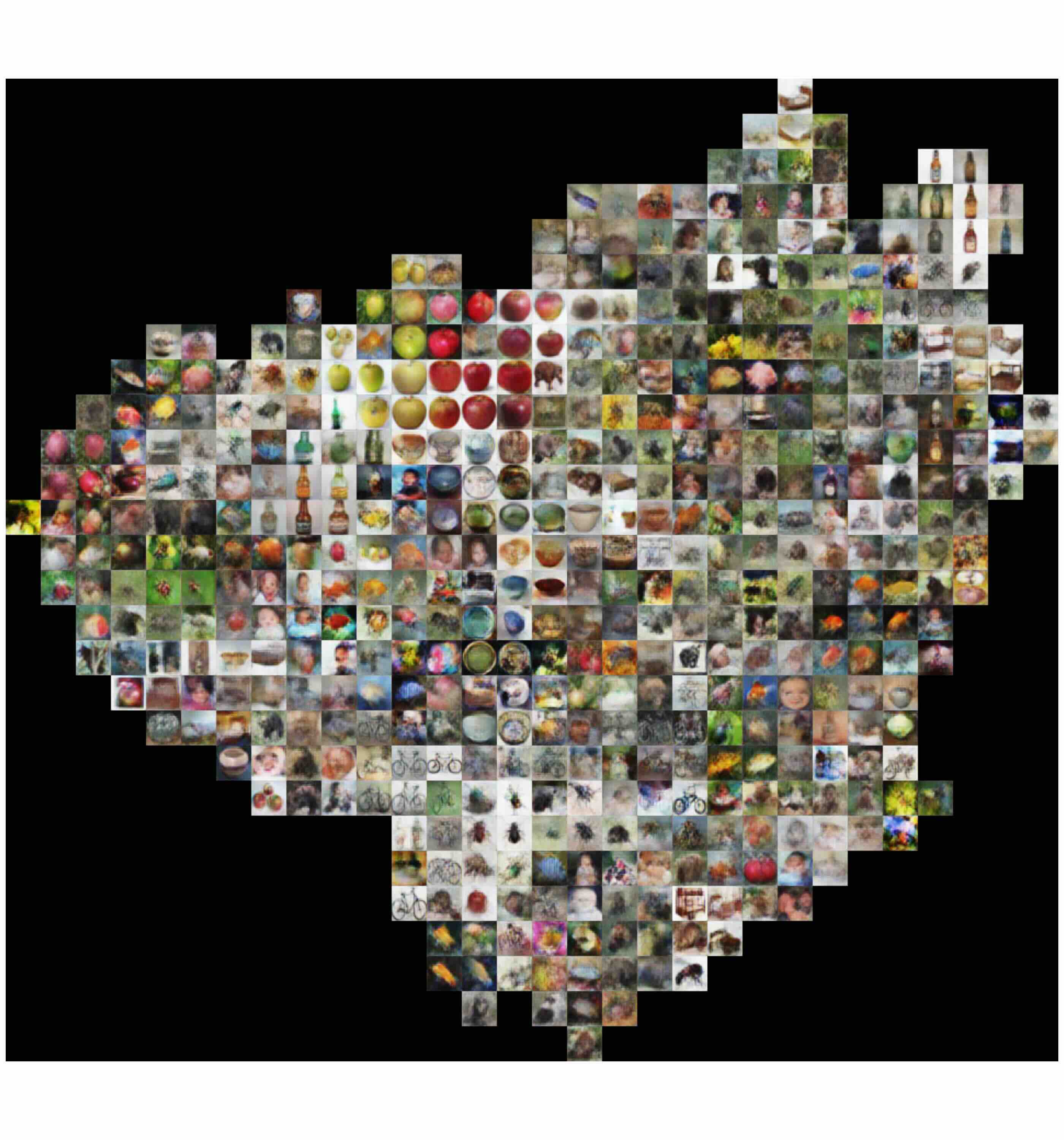} \\[1mm]

    \includegraphics[width=0.32\textwidth]{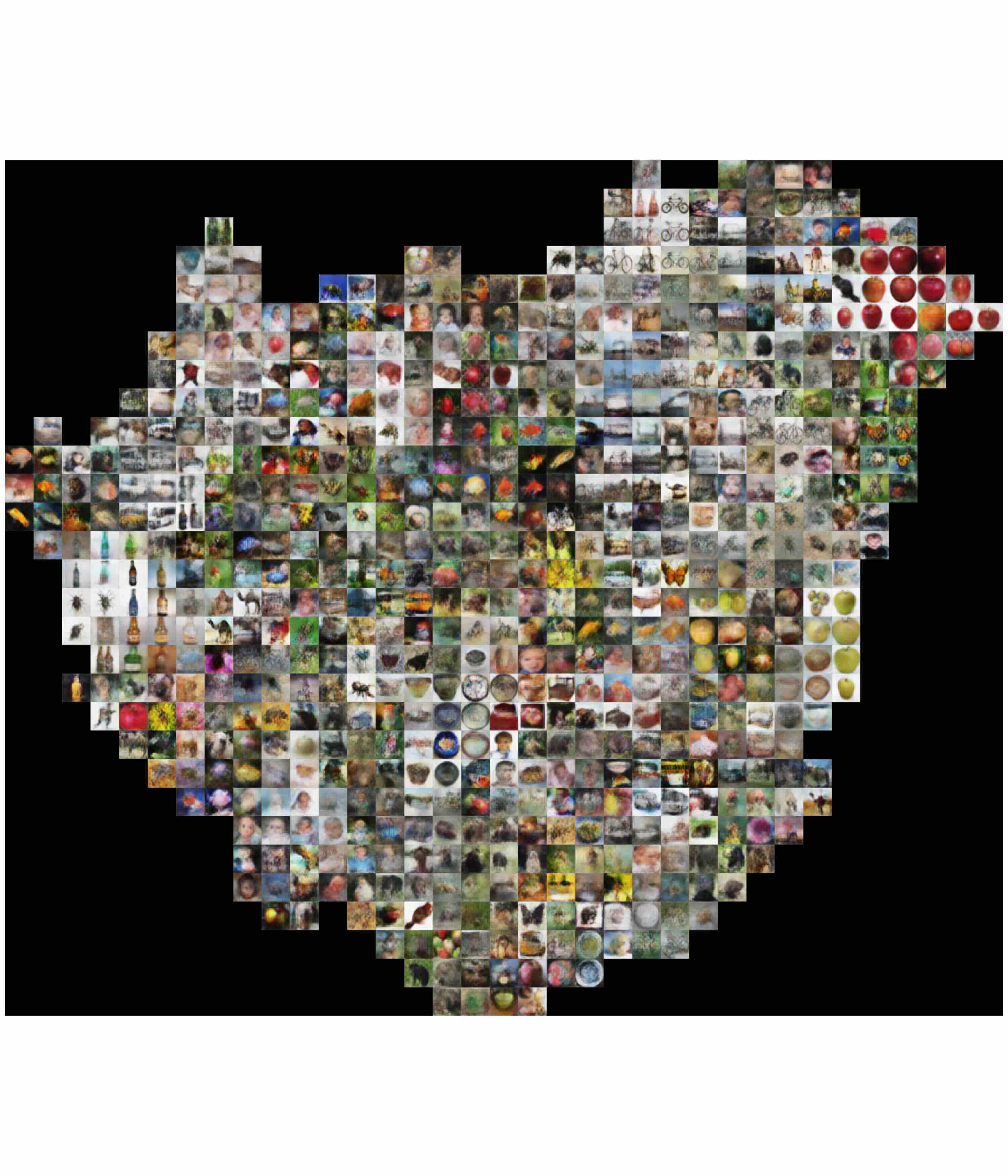}
    \includegraphics[width=0.31\textwidth]{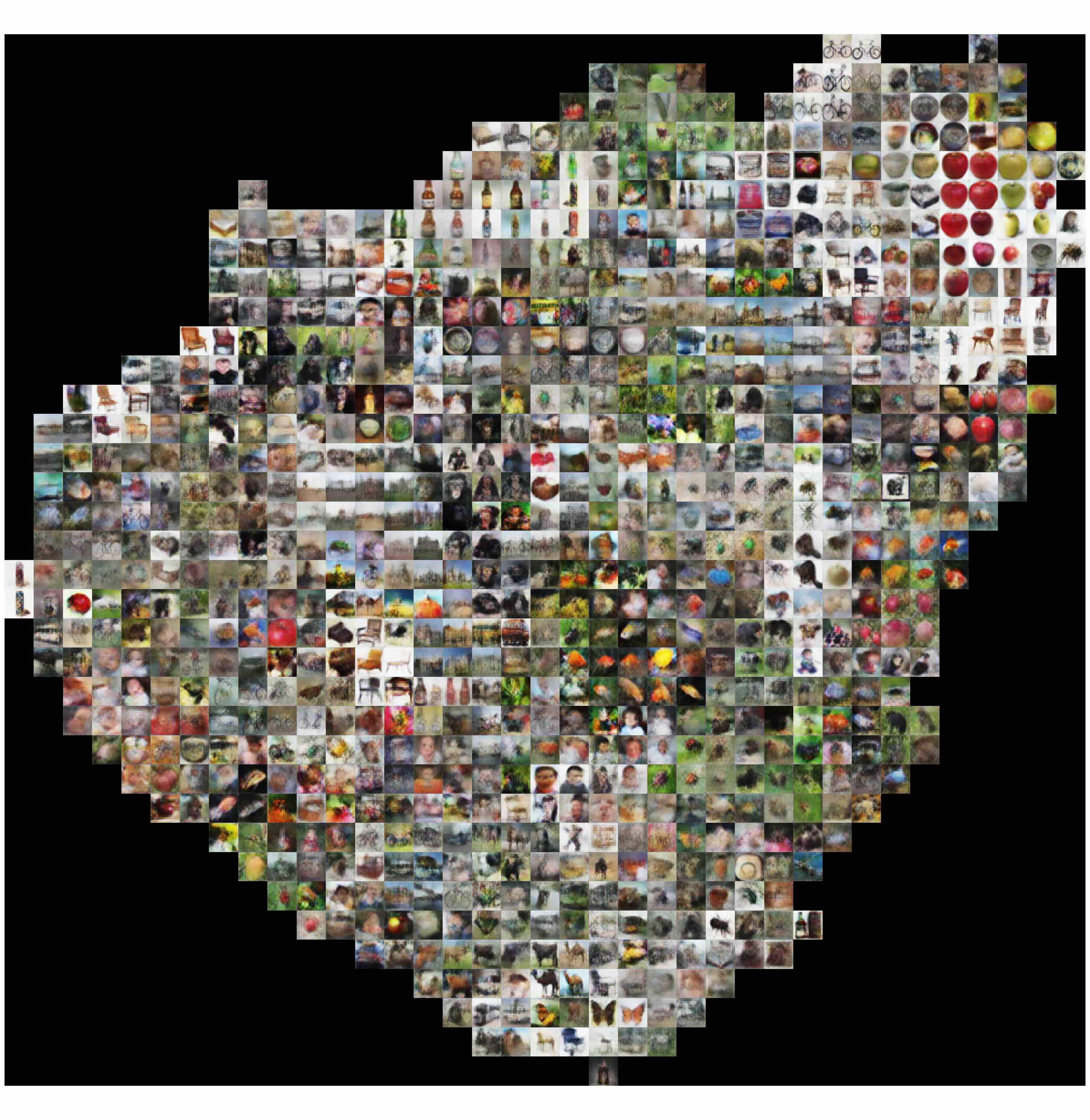}
    \includegraphics[width=0.33\textwidth]{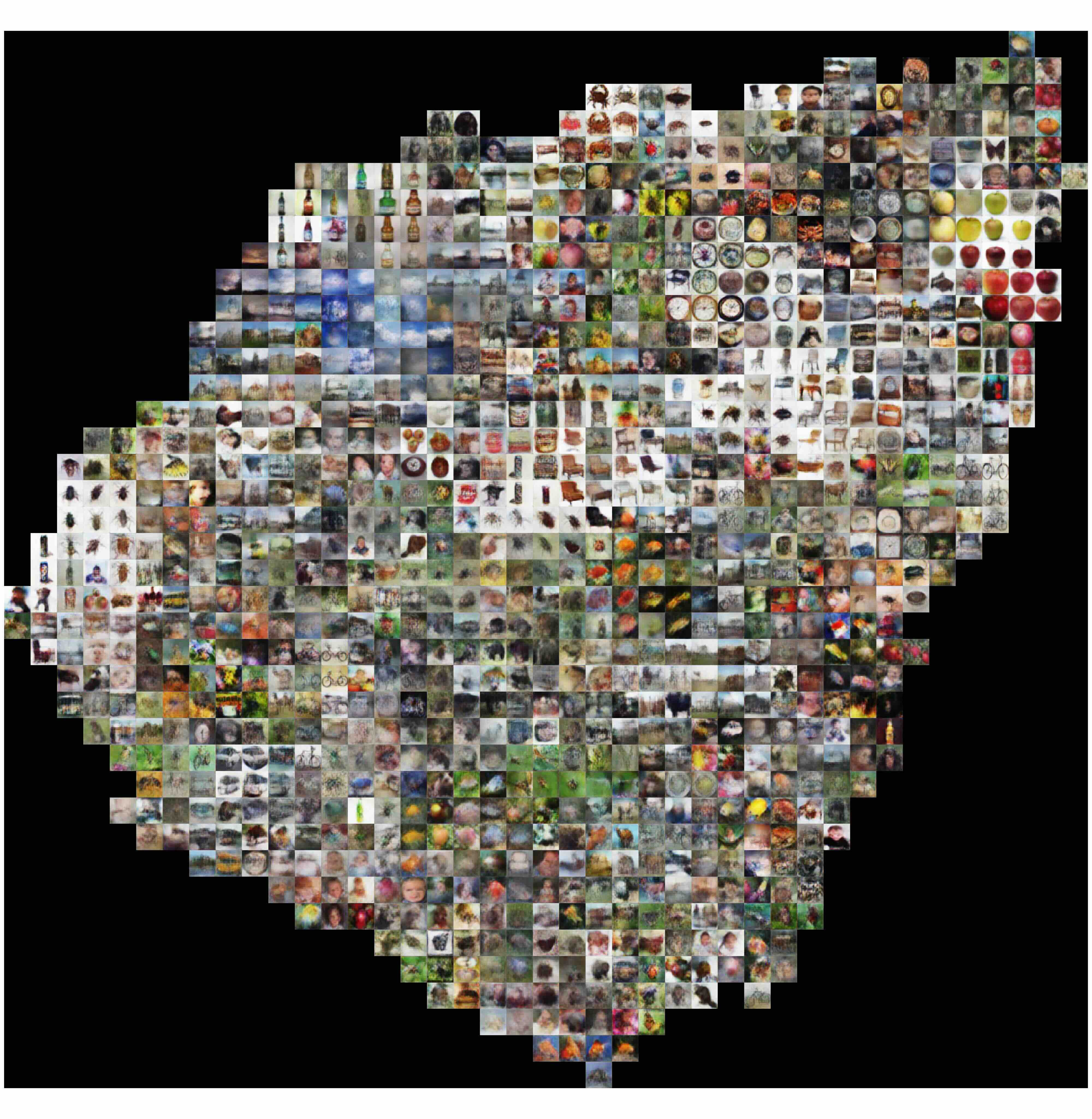} \\[1mm]

    \includegraphics[width=0.32\textwidth]{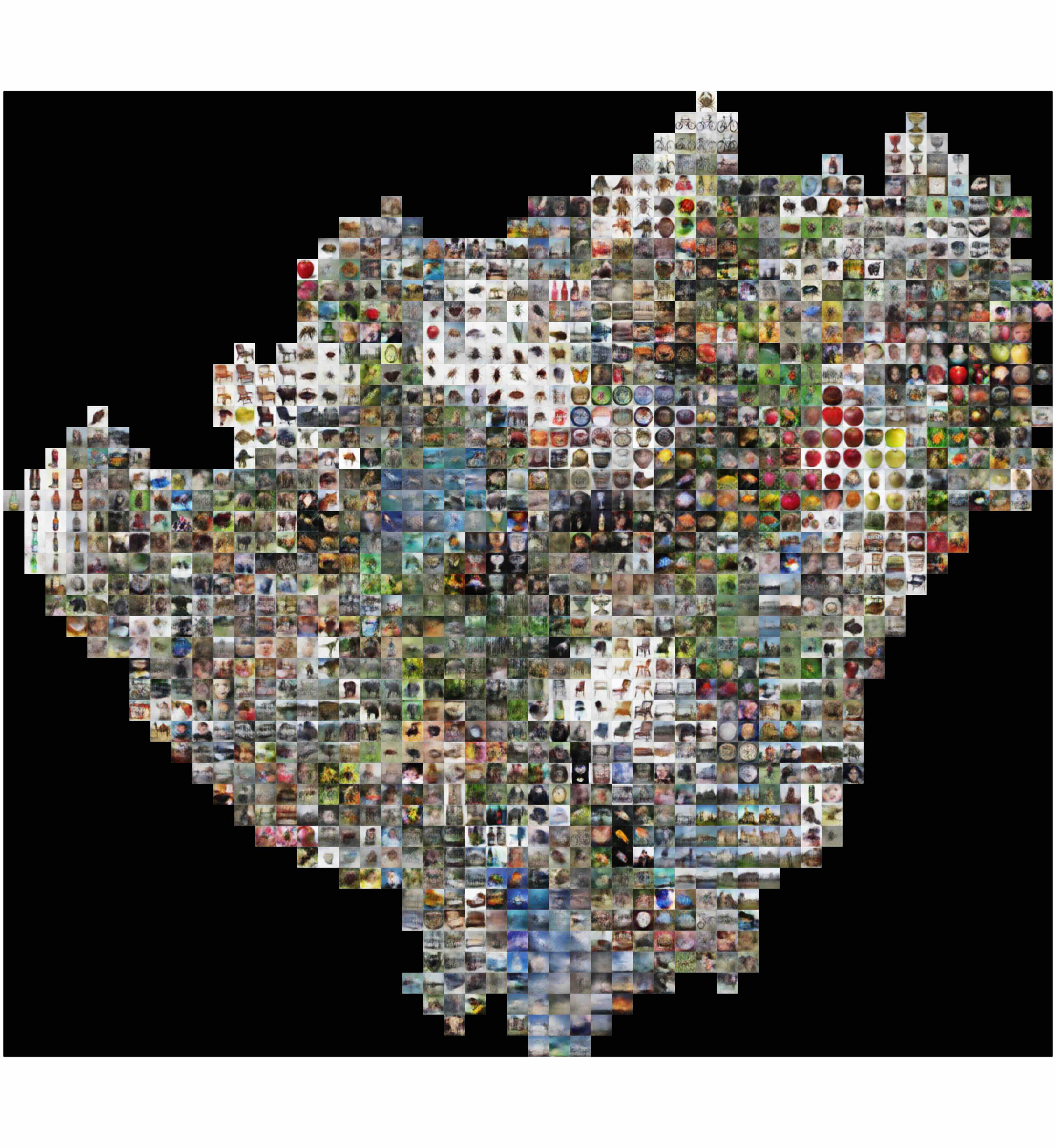}
    \includegraphics[width=0.32\textwidth]{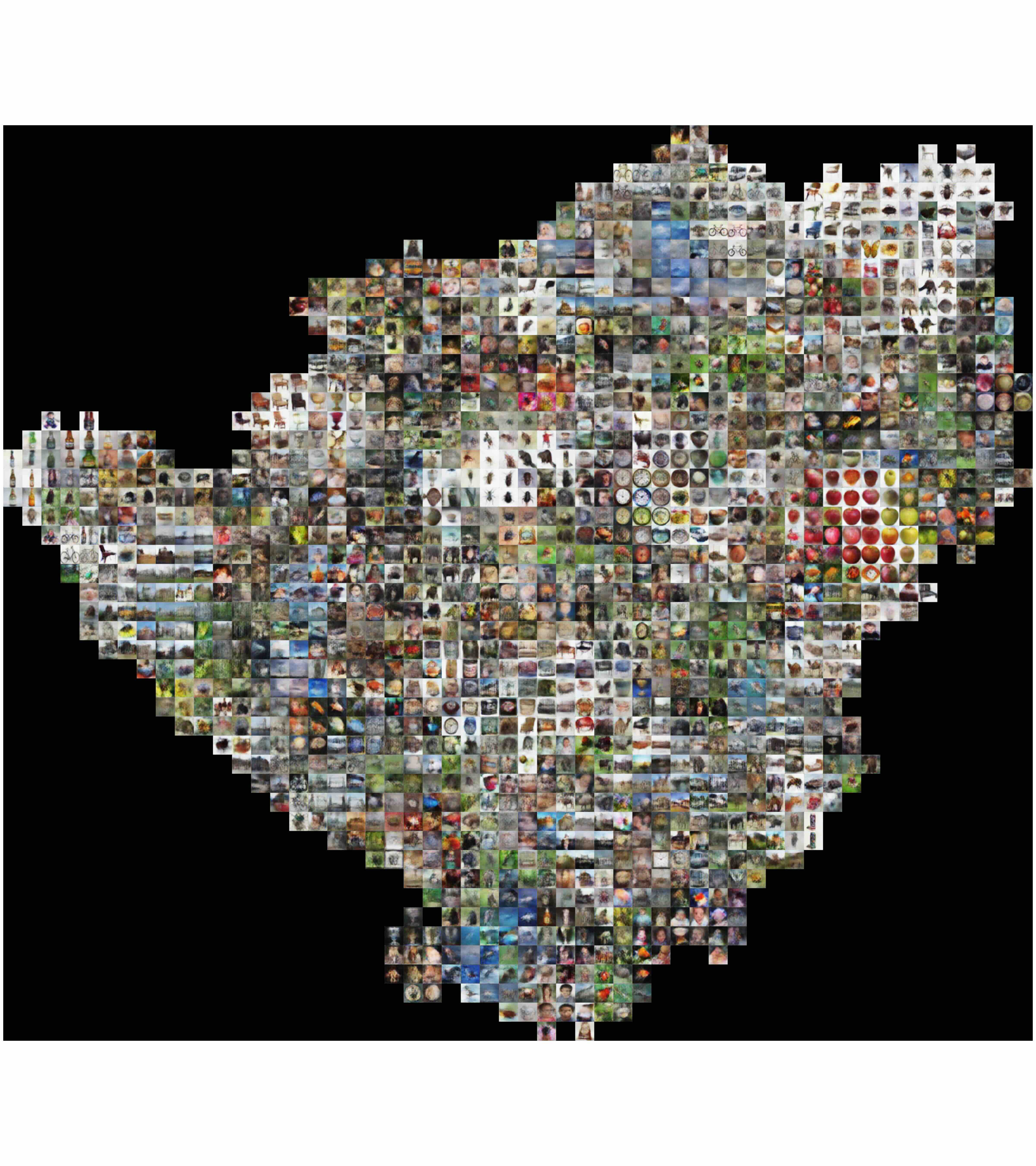}
    \includegraphics[width=0.32\textwidth]{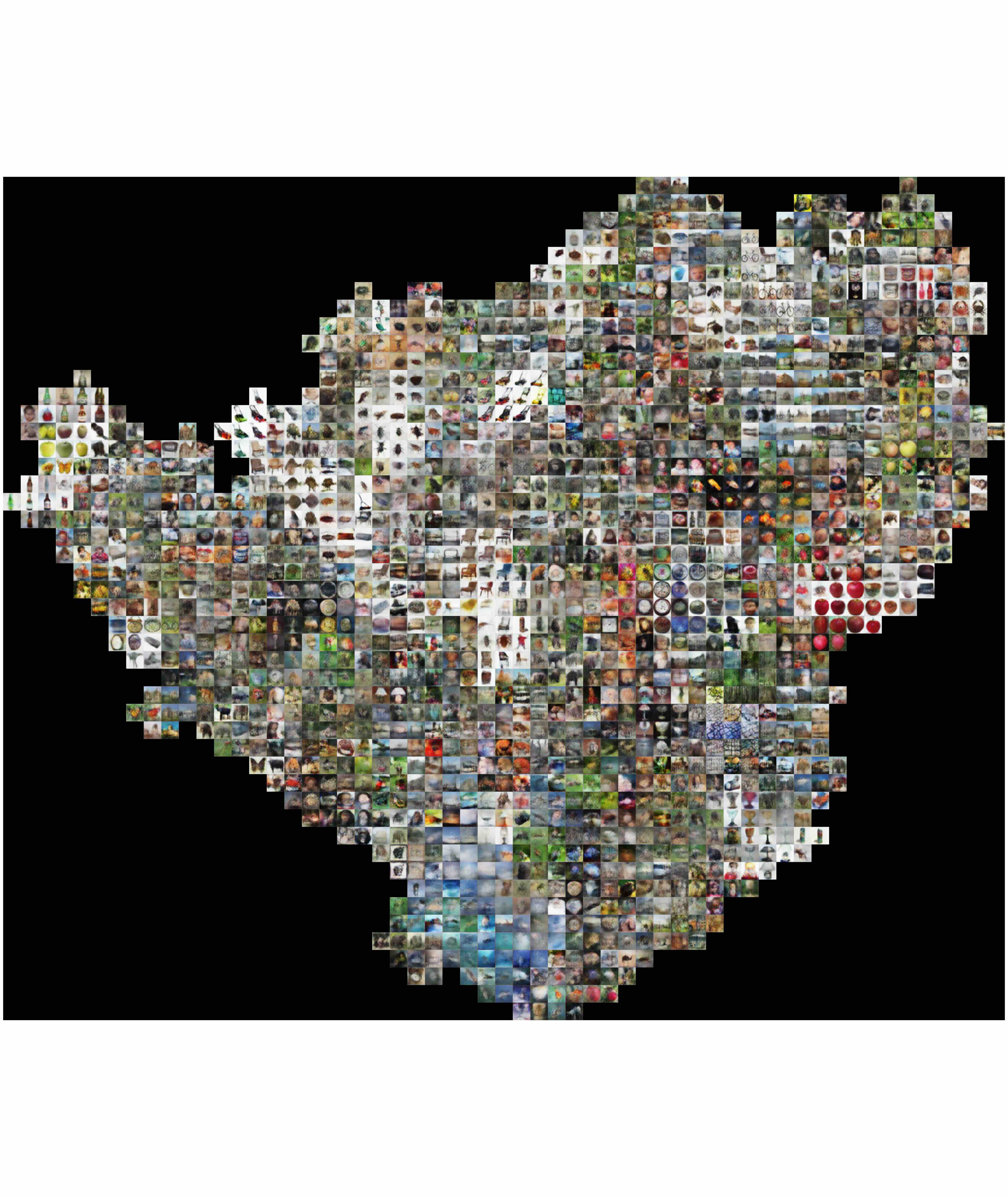} \\[1mm]

    \includegraphics[width=0.33\textwidth]{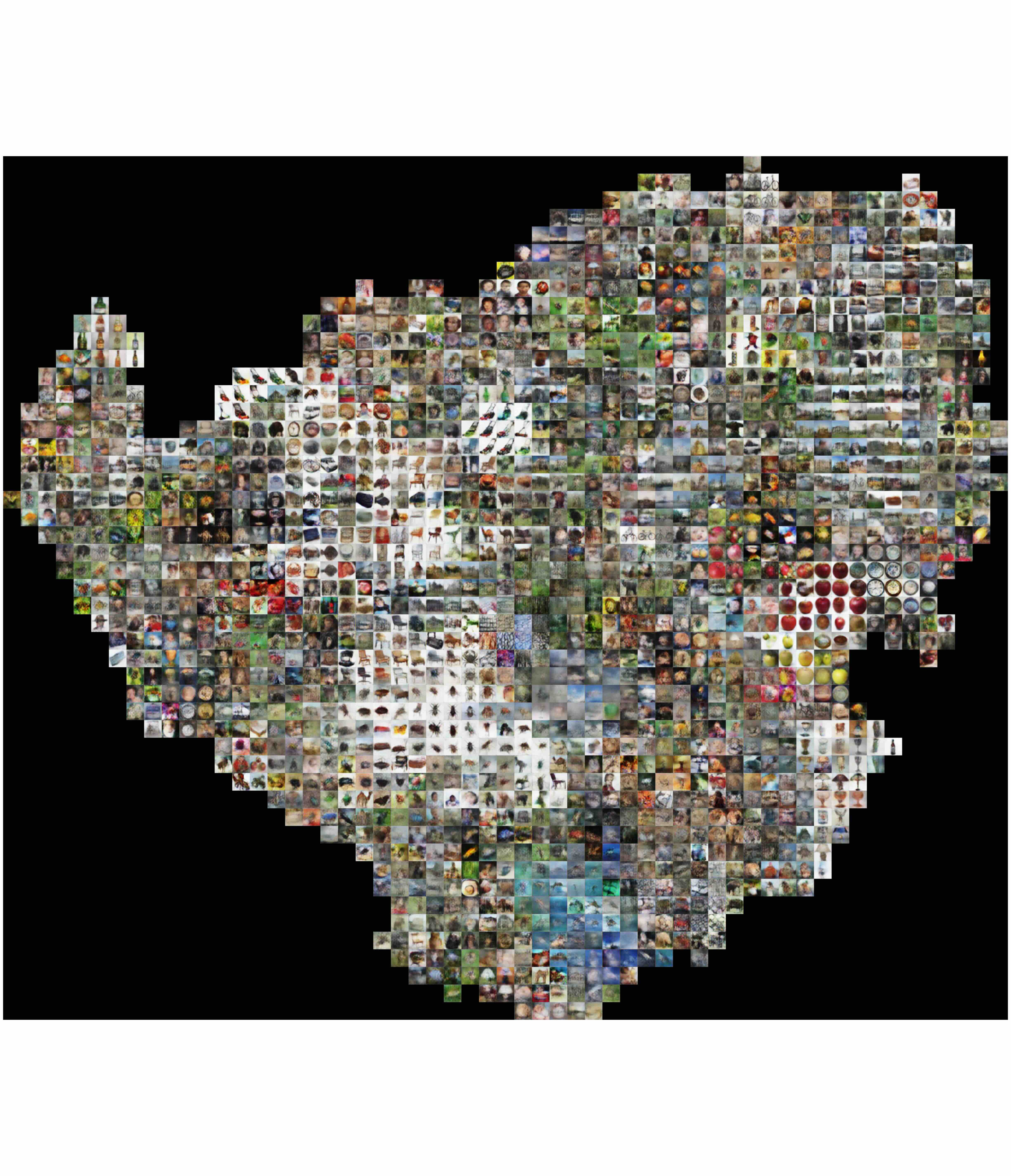}

    \caption{Visualization of decoded GSOM unit vectors after each task(every 10 classes) on CIFAR-100.}
    \label{fig:appendix_cifar_100_gen}
\end{figure*}

Figure~\ref{fig:appendix_mnist_gen}--\ref{fig:appendix_cifar_100_gen} present the evolution of GSOM unit representations across tasks for MNIST, CIFAR10, and CIFAR100 datasets. For MNIST, the learned representations quickly organize into well-structured and visually coherent prototypes, with clear class-specific patterns emerging after each task and remaining relatively stable over time. In CIFAR-10, the decoded representations are more diverse but still exhibit recognizable semantic structure, indicating that the global VAE is able to capture meaningful features despite increased complexity. For CIFAR-100, the representations become significantly more fragmented and less visually distinct, reflecting the higher intra-class variability and task difficulty. Overall, these visualizations highlight the ability of GSOM to progressively organize latent space representations, while also illustrating the challenges in maintaining distinct and stable prototypes under more complex data distributions.

\section{Societal Impacts}
\label{app:societal_impacts}
The proposed method promotes positive societal impacts through scalable and memory-efficient continual learning systems that can support long-term adaptive AI without storing large raw datasets, which may help reduce privacy concerns associated with exemplar replay. However, continually adapting models may also inherit biases from evolving data streams or exhibit unreliable behavior under distribution shifts. In safety-critical or decision-making applications, such failures could lead to unintended consequences if models are deployed without appropriate oversight and evaluation. Although the proposed approach is developed for research purposes, ensuring safe deployment and misuse prevention remains future work.

\end{document}